\def\shortonly#1{}
\def\longonly#1{#1}
\def\nonblind#1{}
\def\vshortonly#1{}

\documentclass{article}
\usepackage{ijcai22}

\usepackage{times}
\usepackage{soul}
\usepackage{url}
\usepackage[hidelinks]{hyperref}
\usepackage[utf8]{inputenc}
\usepackage[small]{caption}
\usepackage{graphicx}
\usepackage{amsmath}
\usepackage{amsthm}
\usepackage{booktabs}
\usepackage{algorithm}
\usepackage{algorithmic}
\usepackage{amsfonts}
\usepackage{amsmath}
\usepackage{multirow}
\usepackage{subcaption}

\def\asinh{\mathrm{asinh}}
\def\asin{\mathrm{asin}}

\def\dist{\delta}

\def\subfig#1#2{
\begin{minipage}[b]{#1} \noindent
  \includegraphics[width=\textwidth]{#2}
  \subcaption{}
\end{minipage}
}

\def\dist{\delta}
\def\bbH{\mathbb{H}}
\def\bbR{\mathbb{R}}
\def\bbS{\mathbb{S}}
\def\bbD{\mathbb{D}}

\def\bbT{\mathbb{T}}
\def\ra{\rightarrow}

\title{Non-Euclidean Self-Organizing Maps}
\author{
    Dorota Celińska-Kopczyńska \and Eryk Kopczyński \\
    \affiliations
    { Institute of Informatics, University of Warsaw}
    \emails
    \{dot, erykk\}@mimuw.edu.pl
    \
}

\begin{document}

\maketitle

\begin{abstract}
Self-Organizing Maps (SOMs, Kohonen networks) belong to neural network models of the unsupervised class. In this paper, we present the generalized setup for non-Euclidean SOMs. Most data analysts take it for granted to use some subregions of a flat space as their data model; however, by the assumption that the underlying geometry is non-Euclidean we obtain a new degree of freedom for the techniques that translate the similarities into spatial neighborhood relationships. We improve the traditional SOM algorithm by introducing topology-related extensions. Our proposition can be successfully applied to dimension reduction, clustering or finding similarities in big data (both hierarchical and non-hierarchical). 
\end{abstract}

\section{Introduction}

Self-Organizing Maps (SOMs, also known as Kohonen networks) belong to neural network models of the unsupervised class allowing for dimension reduction in data without a significant loss of information. SOMs preserve the underlying topology of high-dimensional input and transform the information into one or two-dimensional layer of neurons.
The projection is nonlinear, and in the display, the clustering of the data space and the metric-topological relations of the data items are visible \cite{kohonen}.  
In comparison to other techniques of reducing dimensionality, SOMs have many advantages. They do not impose any assumptions regarding the distributions of the variables and do not require independence among variables. They allow for solving non-linear problems; their applications are numerous, e.g., in pattern recognition (see, e.g., \cite{pattern}), brain studies \cite{brain_review,brain,brain_3} or biological modeling \cite{bio,tomatoes}. At the same time, they are relatively easy to implement and modify \cite{kohonen,soms}.

A typical setup for SOM assumes usage of a region of Euclidean plane. On the other hand, non-Euclidean geometries are steadily gaining attention of the data scientists \cite{tda,tda_chazal}. In particular, hyperbolic geometry has been proven useful in data visualization \cite{munzner}
and the modeling of scale-free networks \cite{hypgeo,papa}. Such a usefulness comes from the exponential growth property of hyperbolic geometry, which makes it much more appropriate than Euclidean for modeling and visualizing hierarchical data. Since the idea of SOM is rooted in geometry, we can expect to gain new insights from non-Euclidean SOM setups. Surprisingly, there are nearly no attempts to do so. Even if there have been propositions to use hyperbolic geometry in SOMs \cite{ritter99,ontrup}, other possibilites of inclusion of non-Euclidean geometries and different topologies (e.g., spherical geometry, quotient spaces) have been neglected. There is also no research on characteristics of data that affect the quality of Self-Organizing Maps.

Against this background, our contributions in this paper can be summarized as follows:
\begin{itemize}
\item We are the first to present the generalized setup for non-Euclidean SOMs. Our proposition allows for usage of (so far neglected or absent) quotient spaces. In consequence, we get a more regular and visually appealing results that the previous setups.  
\item By using Goldberg-Coxeter construction, our proposition allows for easy scalability of the templates. It also makes spheres a worthy counterpart for analysis -- we are no longer restricted to usage of platonic solids.
\item To our best knowledge, we are the first to extend SOM setup by non-Euclidean aspects other than the shape of the template. We introduce geometry-related adjustments in the dispersion function. Moreover, we show that our proposition improves the results in comparison to traditional Gaussian dispersion.
\item Our quantitative analysis proves that the shape of data matters for Self-Organizing Maps. We use measures of topology preservation from the literature\longonly{, as well as our own measures}.
\item The results of non-Euclidean SOMs have interpretation. Usage of different geometries allows us to find and highlights various aspects in the data sets. E.g., spherical geometry allows for an easy examination of polarization, and hyperbolic geometry due to the exponential growth fosters finding similarities. This makes non-Euclidean SOMs suitable both as stand-alone technique, but also as an auxiliary one to include in other models. 
\end{itemize} 

\section{Prerequisities}
\subsection{Non-Euclidean geometries}
Most data analysts take it for granted to use some subregions of a flat space as their data model, which means utilizing constructs which follow the principles of the Euclidean geometry.%
\longonly{However, the fifth axiom of this geometry is a problematic one and raises some questions about the nature of parallelness. Take a line $L$ and a~point $A$. According to Euclidean geometry principles, there is exactly one line going through $A$ which does not cross $L$. However, can there be more? Or less? \\}%
\longonly{We can find the answers to those questions in non-Euclidean geometries. The first, and probably the most famous one, is the hyperbolic geometry, discovered by Gauss, Lobachevsky, and Bolyai. In this case, there are infinitely many lines going through $A$ which do not cross $L$. One of the properties of this geometry is that the amount
of the area in the distance $d$ from a given point is exponential in $d$; intuitively, 
the metric structure of the hyperbolic plane is similar to that of an infinite binary
tree, except that each vertex is additionally connected to two adjacent vertices on the same
level. \\
While hyperbolic geometry is not common in our world (typical examples include coral reef or lettuce), the second kind of non-Euclidean geometry is more common---that is, the geometry of the sphere. When we consider
great circles on the sphere (such as the equator, or the lines of constant longitude) to be straight lines, no lines go through $A$ which do not cross $L$.} %
\shortonly{However, by}\longonly{By}
the assumption that the underlying geometry is non-Euclidean, we obtain a~new degree of freedom for the techniques of analysis which translate the similarities into spatial neighborhood relationships \cite{ritter99}. Recall that formally, the Euclidean plane is the set of points $\{(x,y); x, y \in \mathbb{R}^2\}$, with the metric $d(a,b) = {||a-b||}$, where $||(x,y)||=\sqrt{x^2+y^2}$. The sphere $\mathbb{S}^2$ is the set of points $\{(x,y,z); x, y, z \in \mathbb{R}^3, x^2+y^2+z^2=1\}$, with the metric $d(a,b) = 2\asin(||a-b||/2)$, where $||(x,y,z)||=\sqrt{x^2+y^2+z^2}$.
The Minkowski hyperboloid $\mathbb{H}^2$ is the set of points $\{(x,y,z); x, y, z \in \mathbb{R}^3, x^2+y^2+1=z^2, z>0\}$, with the metric $d(a,b) = 2\asinh(||a-b||/2)$, where we use the Minkowski metric $||(x,y,z)||=\sqrt{x^2+y^2-z^2}$. 

\begin{figure}
\centering  
\subfig{0.24\linewidth}{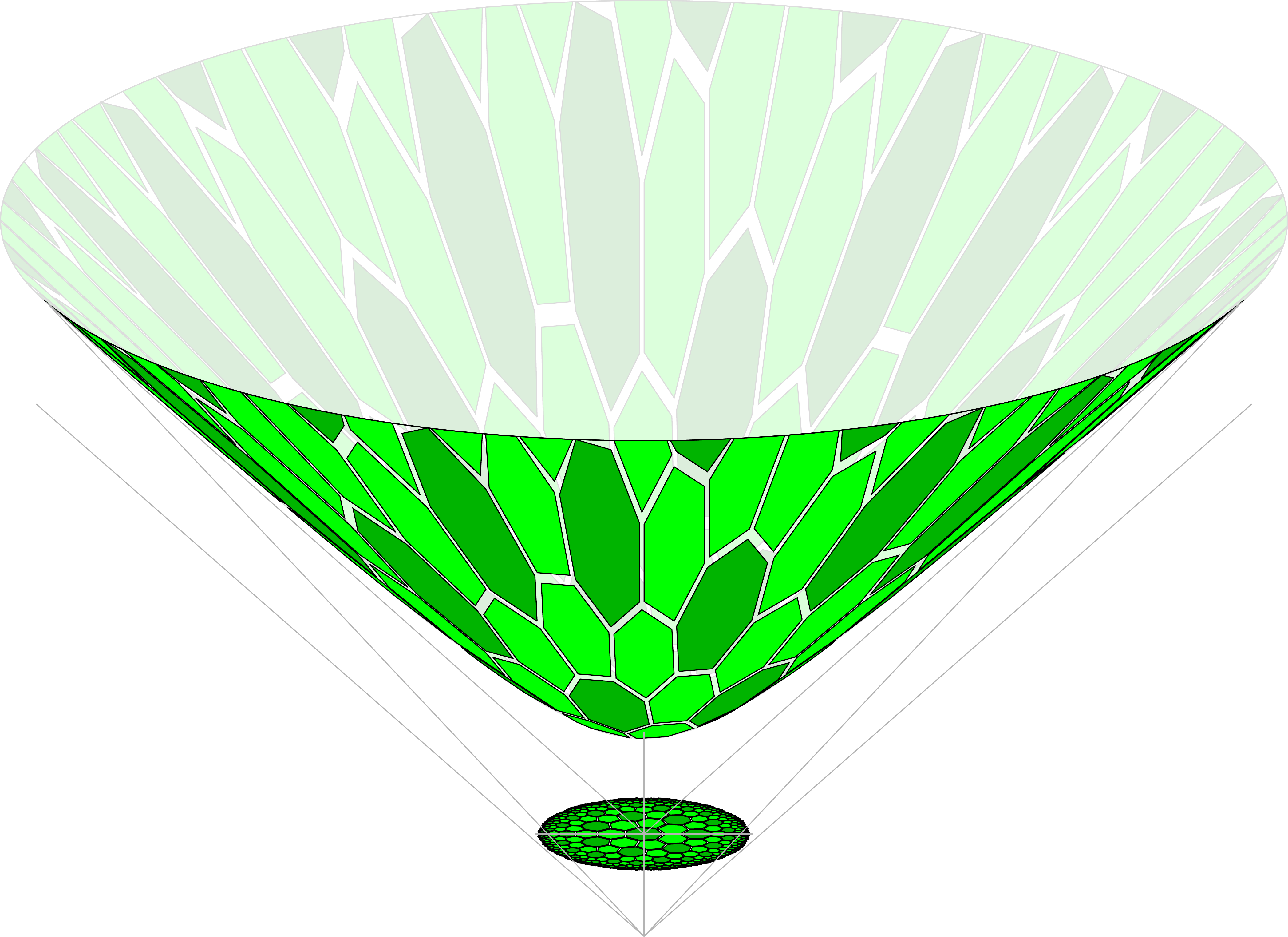}
\subfig{0.22\linewidth}{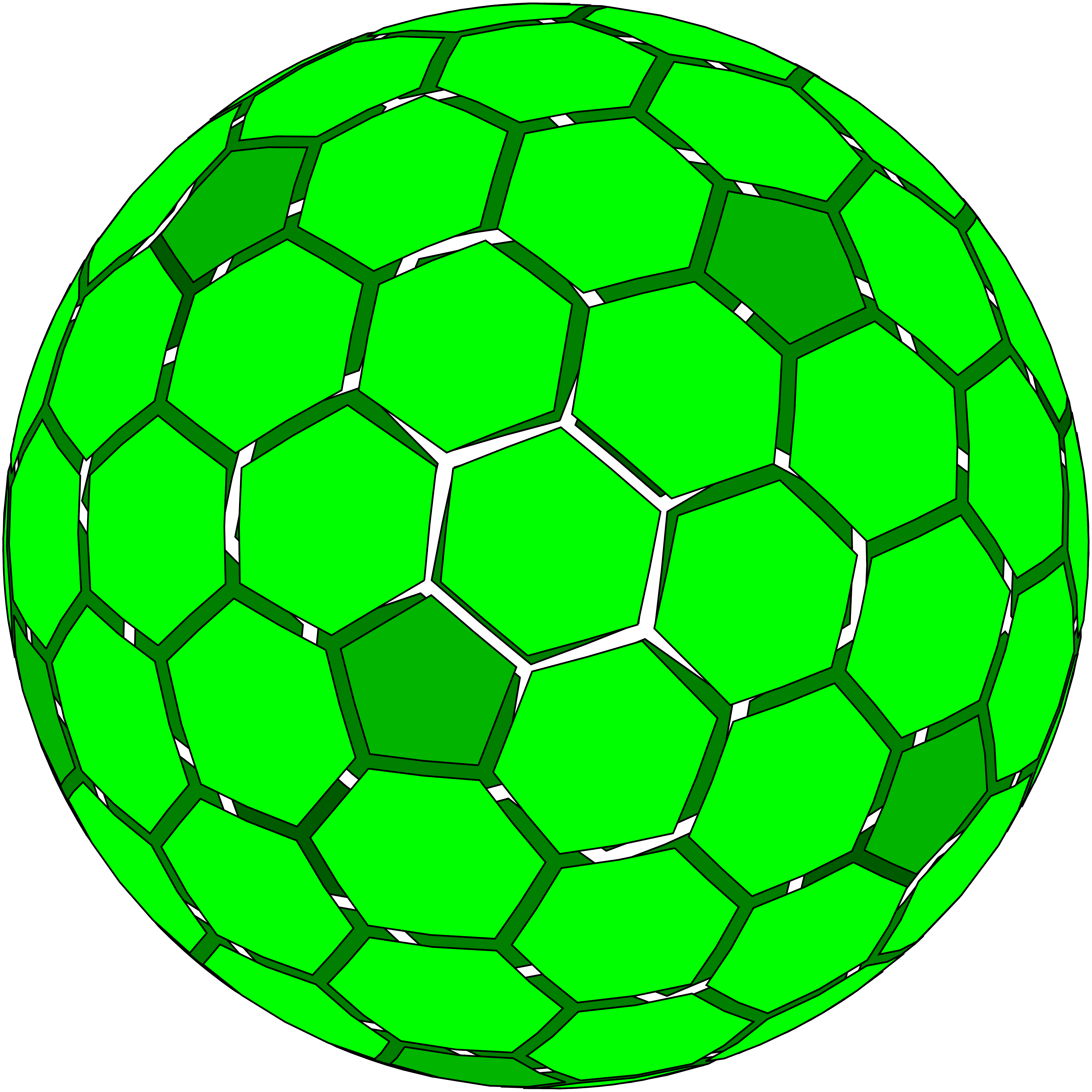}
\subfig{0.22\linewidth}{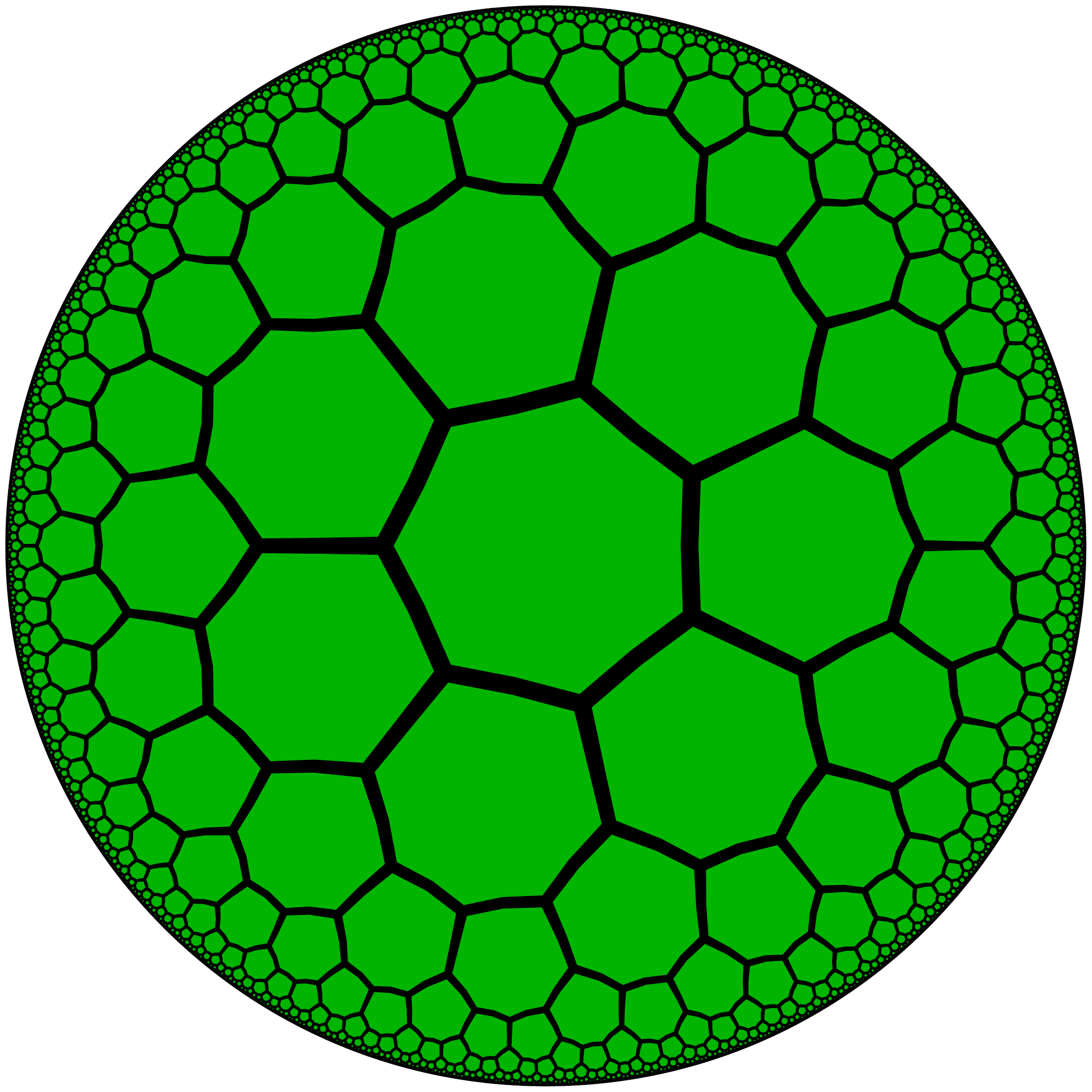} \hskip -1mm
\subfig{0.22\linewidth}{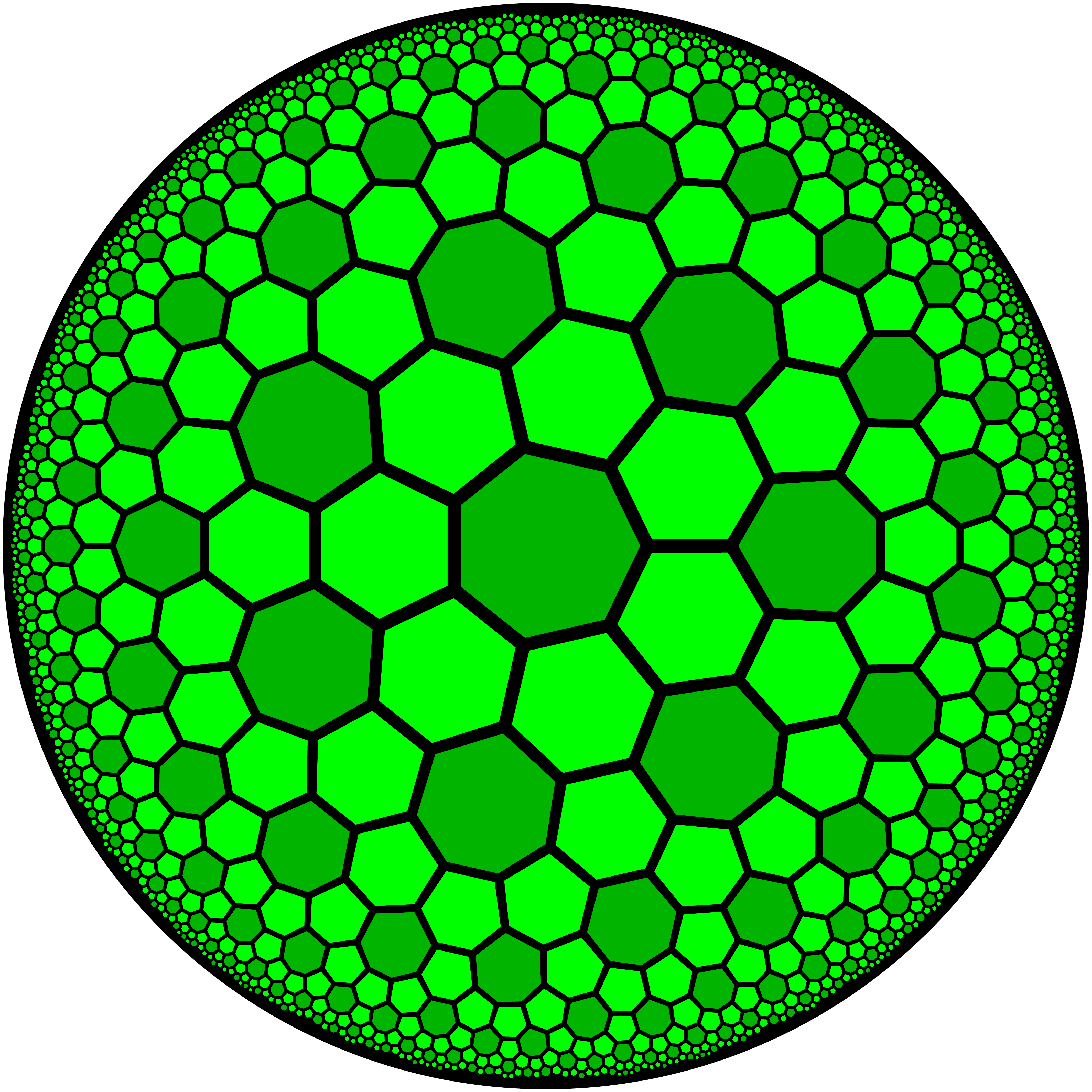} \hskip -1mm


\caption{\label{big}
Representations of non-Euclidean geometries: (a) Minkowski hyperboloid; (b) Sphere.
Hyperbolic tessellations in Poincar\'e disk model: (c) order-3 heptagonal tiling, (d) bitruncated order-3 heptagonal tiling.
}
\end{figure}

\longonly{If we perceive the surface of the sphere in $\mathbb{R}^3$ as the ``true form'' of spherical geometry (Figure \ref{big} (b)), then the Minkowski hyperboloid should be a ``true form'' of hyperbolic geometry. However, this model may be unintuitive. Minkowski hyperboloid lives in the Minkowski space, defined by Minkowski metric. This means that if the points on the hyperbolic space are in the distance $d$ to each other, they will be in the distance $d$ to each other on the Minkowski hyperboloid, but only according to the Minkowski metric. According to the usual metric, they can be very distant even if $d$ is small! Figure \ref{big} (a)  depicts such a situation; the heptagons which appear to be oblong are regular.}

All the geometries mentioned are characterized by constant curvature $K$: $K=0$ in the case of Euclidean plane, while in \longonly{the hyperbolic geometry}\shortonly{$\bbH^2$} we have $K<0$, and in \longonly{spherical geometry}\shortonly{$\bbS^2$} we have $K>0$.


\longonly{From the practical point of view, we live on a spherical sector (similar to the flat plane), which makes us more comfortable with imagining things in Euclidean rather than spherical
or hyperbolic way. The limitations of our senses make solving the problem of visualization with tangible results quite a challenge. However, the technology allows us to use computer simulations to 
picture being inside a non-Euclidean space. There are numerous projections of non-Euclidean surfaces; here we will present popular examples.}

\longonly{\paragraph{Orthographic projection.} The surface of the sphere (Figure \ref{big} (b)) is an isometric 3D model of spherical geometry. To represent it in two dimensions, we need a~projection. In orthographic projection we project $(x,y,z)$ to $(x,y)$. The shapes and areas are distorted, particularly near the edges. For hyperbolic geometry, Gans model is orthographic.}

\longonly{\paragraph{Stereographic projection.}  Stereographic projection projects the point $a$ of the unit sphere to the point $b$ on the plane $z=1$ such that $a$, $b$, and $(0,0,-1)$ are colinear.
This projection is conformal, i.e., it preserves angles at which curves meet. One of the widely used models of hyperbolic geometry, the Poincar\'e disk model, is the hyperbolic counterpart of the stereographic projection.
We can obtain Poincar{\'e} model from the Minkowski
hyperboloid model by viewing the Minkowski hyperboloid from $(0,0,-1)$ (Figure \ref{big}).}

Figure \ref{big} shows two tilings of \longonly{the hyperbolic plane}\shortonly{$\bbH^2$}, the order-3 heptagonal
tiling and its bitruncated variant, in the Poincar\'e disk model.
In the Poincar\'e disk model, points of \longonly{the hyperbolic plane}\shortonly{\shortonly{$\bbH^2$}} are represented by points inside a disk. 
\longonly{We can view the Poincar\'e model as a planar map of the hyperbolic
plane -- however, the scale of the map is not a constant: if a point $A$ of the
hyperbolic plane is represented as a point $A'$ such that the distance of $A'$ from
the boundary circle of the model is $d$ then the scale is roughly proportional to $d$.}
In the hyperbolic metric, all the triangles, heptagons and hexagons in each of the tessellations in Figure \ref{big} are actually of the same size,
and the points on the boundary of the disk are infinitely far from the center.

\subsection{Tessellations of non-Euclidean spaces}\label{tones}

Tessellations from Figure \ref{big} can be 
\longonly{naturally }interpreted as metric spaces, where the points are the tiles,
and the distance $\dist(v,w)$ is the number of edges \longonly{we have to traverse}\shortonly{crossed} to reach $w$ from $v$.
Such metric spaces have properties similar to the underlying surface.

\def\sch#1#2{\{#1,#2\}}
\def\scht#1{\sch{3}{#1}}
\def\schq#1{\sch{4}{#1}}
\def\gp#1#2{GC_{#1,#2}}

\paragraph{Schl\"afli symbol} In a regular tessellation every face is a regular $p$-gon, and every degree has degree $q$ (we assume $p,q\geq 3$). We say that such a tessellation has a Schl\"afli symbol $\sch{p}{q}$. Such a tessellation exists on the sphere if and only if $(p-2)(q-2)<4$, plane if and only if $(p-2)(q-2)=4$, and hyperbolic plane if and only if $(p-2)(q-2) > 4$.


Contrary to the Euclidean tessellations, we cannot scale hyperbolic or spherical tessellations. On a hyperbolic plane of curvature -1, every face in a $\sch{q}{p}$ tessellation 
will have area $\pi(q\frac{p-2}{p}-2)$. Thus, among hyperbolic tessellations of form $\sch{q}{3}$, $\sch{7}{3}$
is the finest, and they get coarser and coarser as $q$ increases. Regular spherical tessellations correspond to the platonic solids.

\subsection{Self-Organizing Maps: general idea}\label{somdefault}

SOM network consists of two layers: the input layer containing the variables in the input data, and the output layer of the resulting clustering.

We describe every element in the input data $D$ using $k$ variables: $D \subseteq \bbR^k$. The elements of $x \in D$ are the values of the $k$ variables which serve as the basis for clustering.
\longonly{Similarly to other dimension-reduction techniques, if there are large differences in the values of variances of the variables in the dataset, standardization of the data is required in order to avoid the dominance of a~particular variable or the subset of variables.} 

Neurons are traditionally arranged in a lattice.
For each neuron $i$ in the set of neurons we initialize the weight vector $w_i \in \bbR^k$. Weights are links that connect the input layer to the output layer. The final results may depend on the distribution of the initial weights \cite{soms}.
\longonly{The weights can be random, determined arbitrarily or obtained during a~preliminary training phase.}
\shortonly{We assign the initial weights randomly.}
The neurons need to be exposed to a~sufficient number of different inputs to ensure the quality of learning processes.
In a usual setup, the formation of the SOM is controlled by three parameters: the learning rate $\eta$, the number of iterations $t_{max}$, and the initial neighborhood radius $\sigma(t_{max})$.
Every iteration involves two stages\longonly{: competition and adaptation}.

\paragraph{Competition stage.} We pick $x_t \in D$. 
The neurons compete to become activated. Only the node that is the most similar to the input data $x_t$ will be activated and later adjust the values of weights in their neighborhood. 
\longonly{The Euclidean distance is a~generally accepted measure of distance, but other methods, e.g., Mahalanobis distance are also available.} For each neuron $i$ in the set of neurons we compute the value of the scoring function $d(w_i, x_t) = \|w - x\|$. The neuron for which the value of the scoring function is the lowest becomes the winning neuron.

\paragraph{Adaptation.} For a~given input, the winning neuron and its neighbors adapt their weights. The adjustments enhance the responses to the same or to a~similar input that occurs subsequently. This way the group of neurons specializes in attracting given pattern in input data.
The input data $x_t$ affects every other neuron $j$ with the factor of $d_{\sigma(t)}(r) = \eta\exp(-r^2/{2\sigma(t)^2})$, where $r$ is the distance between
the neuron $j$ and the winning neuron $i$, and $\sigma(t)$ is the neighborhood radius in the iteration $t$; we take
$\sigma(t) = \sigma(t_{max})(1-t/t_{max})$ \cite{kohonen,soms,ritter99,ontrup}.

This dispersion has a~natural interpretation in the Euclidean geometry. Imagine the information as particles spreading between
neurons according to the random walk model: each particle starts in neuron $i$, and in each of time steps, the information can randomly spread
(with probability $p$) to one of the adjacent neurons. From the Central Limit Theorem we know that the distribution of particles after $tC$ time steps approximates the normal distribution with variance proportional to $t$, which motivates using the function $d_{\sigma(t)}(r)$. Heat conduction is a~well-known physical process which works according to \longonly{very }similar rules, but where time and space are \longonly{considered }continuous.


\section{Our contribution}

The core idea of the SOM algorithm is using a~deformable template to translate data similarities into spatial relationships. 
The overwhelming majority of SOM applications use subregions of Euclidean space. 
Instead, we use non-Euclidean geometries to take advantage of their properties, such as the exponential growth of hyperbolic space.
While the basic idea has appeared in \cite{ritter99,ontrup}, we improve on it in the following ways.

\subsection{Choice of the tessellation}
Continuous computations can be costly and prone to precision errors. Continuity is also not always essential. Usually, SOMs are based on the regular grids.
Ritter \cite{ritter99} argues that spherical tessellations are not useful in data analysis, because there are only five regular tessellations, namely platonic solids. Those solids are rather coarse and provide limited possibilities for manipulations of neighborhoods, even in comparison with the Euclidean surfaces. 
Similarly, regular hyperbolic tessellations such as $\sch{7}{3}$ suffer because the exponential growth is too fast.

We combat these issues while losing only a bit of regularity by using the Goldberg-Coxeter construction. 
This construction adds additional hexagonal tiles. 
Consider the hexagonal grid $\sch{6}{3}$ on the plane, and take an equilateral triangle $X$ with one vertex in the point $(0,0)$ and another vertex in the point obtained by moving $a$ steps in a
straight line, turning 60 degrees right, and moving $b$ steps more. The tessellation $\gp{a}{b} \sch{p}{3}$ is obtained from the triangulation $\sch{3}{p}$ by replacing each of regular triangles with a copy
of $X$. In Figure \ref{goldberg}, brown lines depict the underlying regular triangulation. Regular tessellations are a special case where $a=1, b=0$.
Figure~\ref{big}b shows the result of applying the Goldberg-Coxeter construction to the sphere.

\longonly{
\begin{figure}
\centering  
\subfig{0.48\linewidth}{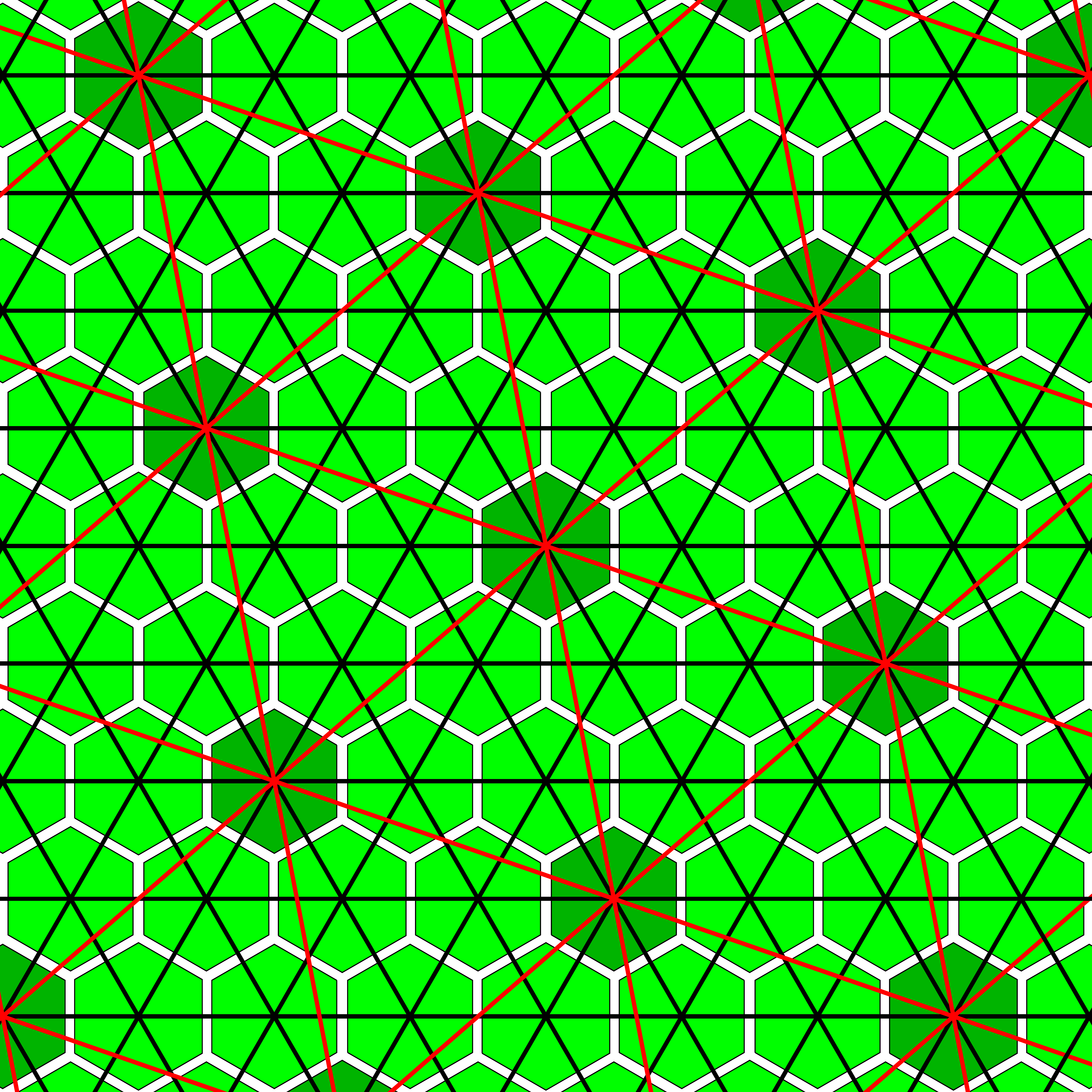}
\subfig{0.48\linewidth}{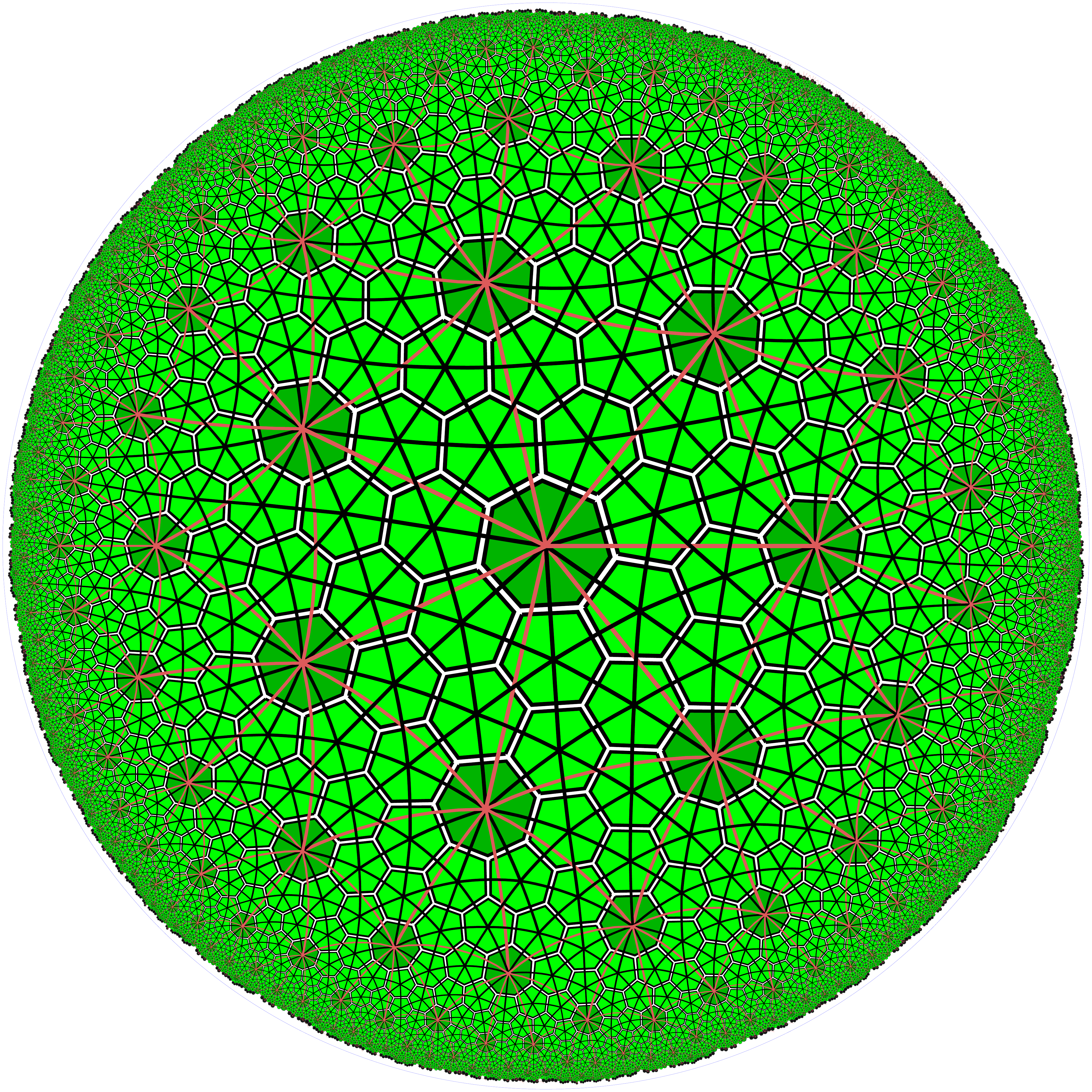}
\caption{Goldberg-Coxeter construction: (a) Euclidean plane; (b) $\gp{2}{1} \sch{7}{3}$ \label{goldberg}}
\end{figure}
}
\shortonly{
\begin{figure}
\centering  
\subfig{0.23\linewidth}{img/goldberg-eu.pdf}
\subfig{0.23\linewidth}{img/goldberg.pdf}
\subfig{0.23\linewidth}{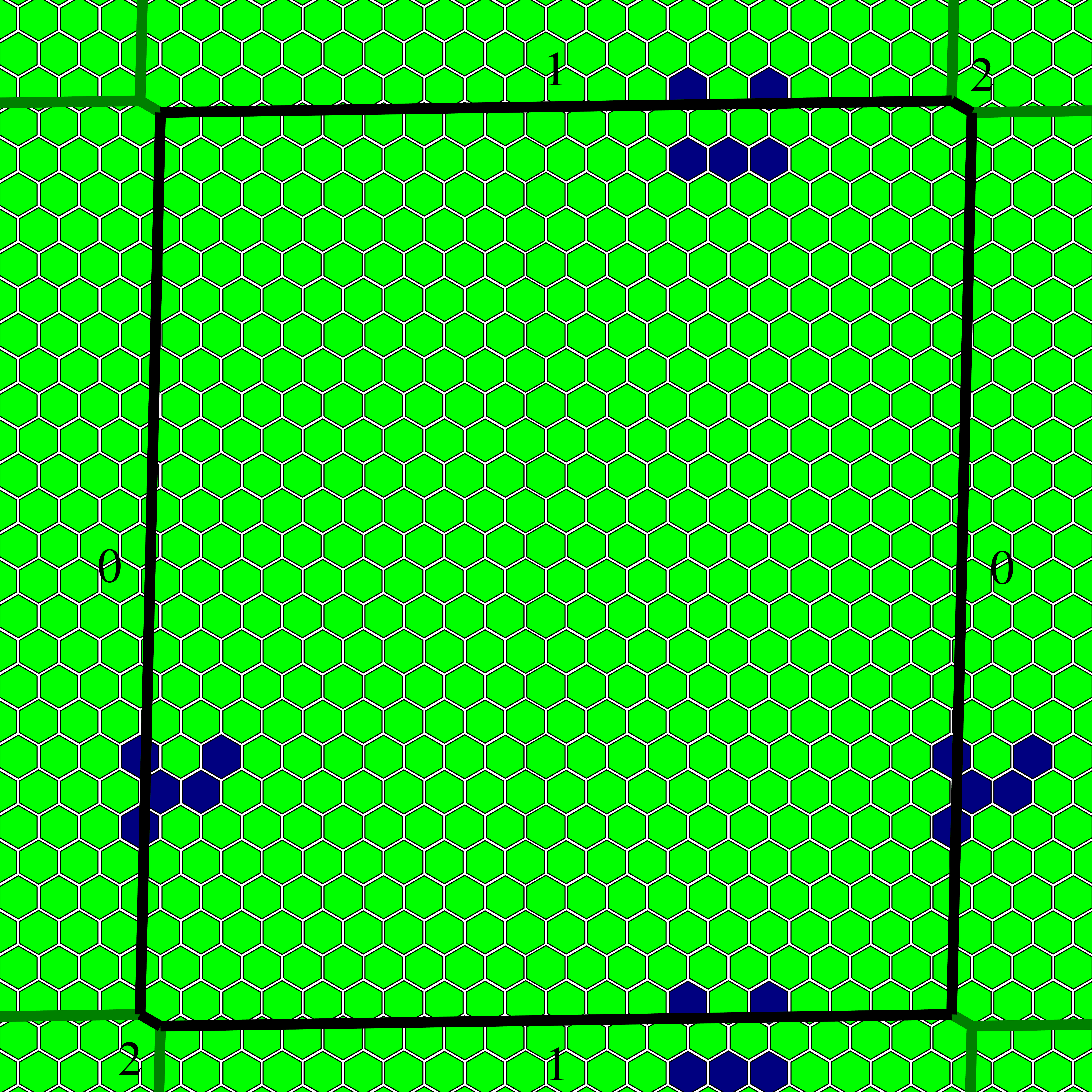} \hskip -1mm
\subfig{0.23\linewidth}{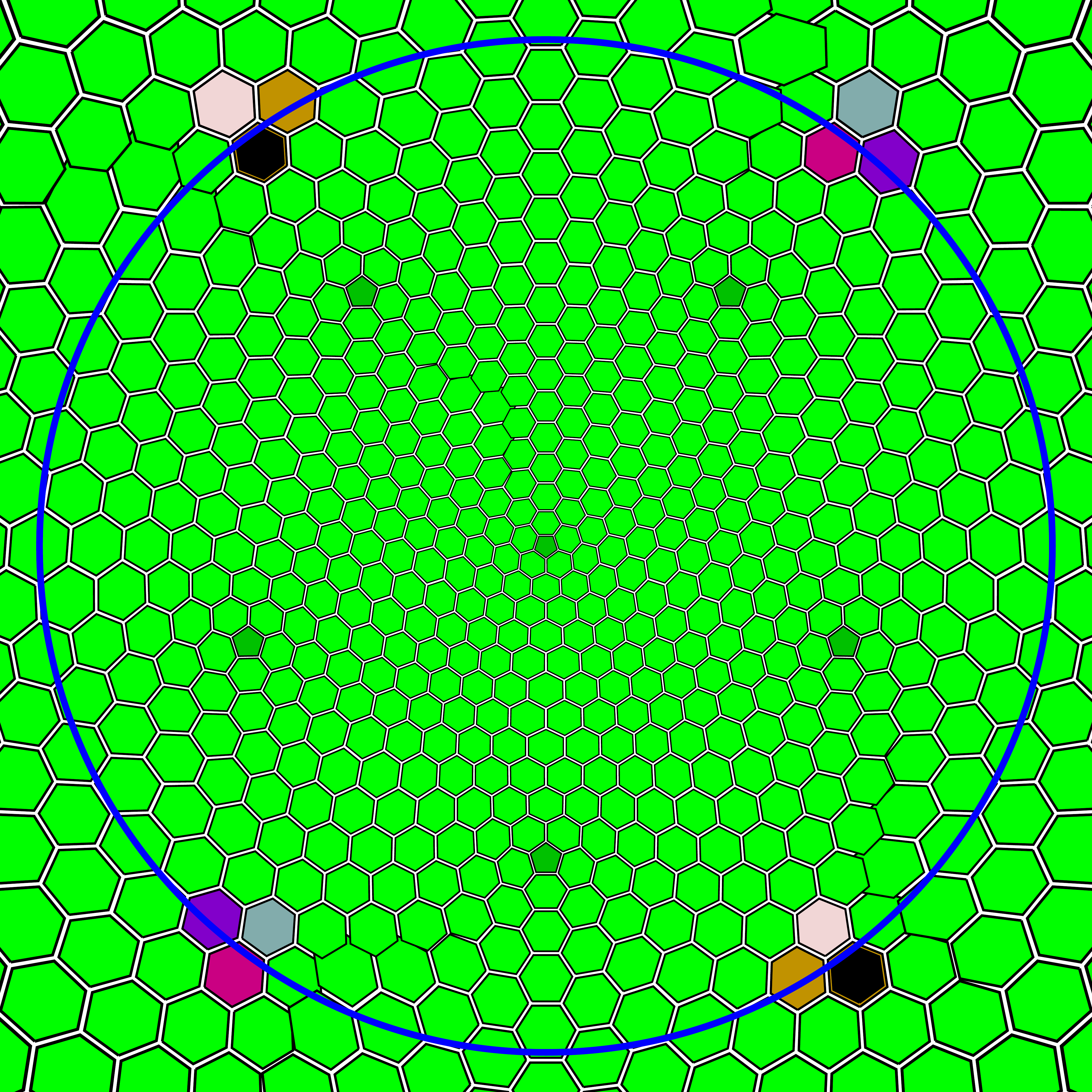}  \hskip -1mm
\caption{Goldberg-Coxeter construction: (a) Euclidean plane; (b) $\gp{2}{1} \sch{7}{3}$. \label{goldberg} \label{goldfund}
Fundamental domains: (c) torus, (d) elliptic plane.}
\end{figure}
}

\subsection{Using closed manifolds}
The effects caused by the neurons on the boundary
having less neighbors may make the maps less regular and less visually attractive.
This problem does not appear on the sphere, which is a closed manifold. On the other hand, it is magnified
in hyperbolic geometry, where the perimeter of a region is proportional to its area, causing a large fraction of the neurons
to be affected by the boundary effects.

We combat these issues by using quotient spaces. A quotient space is obtained by identifying points in the manifold.
For example, a square torus, a quotient space of the Euclidean plane, is obtained by identifying points $(x,y)$ and $(x',y')$ such that $x-x'$ and $y-y'$ are both integers. 
We call the original Euclidean plane the {\bf covering space} of the torus.
Intuitively, a quotient space is created by cutting a fragment of a surface and gluing its edges.

While the torus is usually presented in introductory topology books in its wrapped, donut-like shape, we present our quotient spaces in the covering space presentation, such as in 
\longonly{Fig.~\ref{quoths} (a)}\shortonly{Fig.~\ref{goldfund} (c))}.
We show the covering space of our manifold; our quotient space corresponds to the periodically repeating part of the picture. Such a presentation lets us present the whole manifold on a
single picture, and is much more clean, especially for hyperbolic or non-orientable quotient spaces. Intuitively, the covering space presentation simulates how the manifold is perceived by the native beings
(or neurons).

In the spherical geometry, we can identify each point with its antipodal point, obtaining the elliptic plane \longonly{(Figure \ref{quoths} (b))}\shortonly{(Figure \ref{goldfund} (d))}. The elliptic plane is non-orientable: a right-handed neuron would see a left-handed version of themselves on the other pole. Figure \longonly{\ref{quoths} (b)}\shortonly{\ref{goldfund} (d)} depicts 
the stereographic projection of the elliptic plane; the blue circle is the equator. \longonly{The tiles of the same color are the same objects -- we may see that the pink tile is symmetrical to its counterpart.}

The sphere is a surface of genus 0, while the torus is a surface of genus 1; the genus of an orientable surface is, intuitively, the number of ``holes'' in it.
Orientable quotient spaces of the hyperbolic plane have genus greater than 1, or equivalently, Euler characteristics $\chi=2-2g < 0$.
If we tile a surface with Euler characteristics $\chi$ with $a$ pentagons, $b$ hexagons and $c$ heptagons in such a way that three polygons meet in every vertex,
the following relationship will hold: $6\chi=a-c$. Thus, a sphere can be tiled with 12 pentagons (dodecahedron), a torus can be tiled with only hexagons, and 
hyperbolic quotient spaces can be tiled with only hexagons and $-6\chi$ heptagons. \shortonly{See the full version \cite{somarxiv} for the details on our manifolds.}

\longonly{The smallest hyperbolic quotient space is a non-orientable surface with $\chi=-1$ (six heptagons), which we call the minimal quotient space.
Hurwitz surfaces are hyperbolic quotient spaces that are highly symmetric: a Hurwitz surface of genus $g$ will have precisely $84(g-1) = -42\chi$ automorphisms, which is the
highest possible number \cite{hurwitz}, and corresponds to all the rotations of each of the $-6\chi$ heptagons.
A Hurwitz surface of genus $g=3$ is called the Klein quartic; Hurwitz surfaces also exist for larger genera, such as 7 or 14.}


\longonly{
\begin{figure}[h!]
\centering
\subfig{0.48\linewidth}{img/fundamental-torus2.pdf} \hskip -1mm
\subfig{0.48\linewidth}{img/fundamental-elliptic.pdf}  \hskip -1mm
\longonly{\\ \subfig{0.48\linewidth}{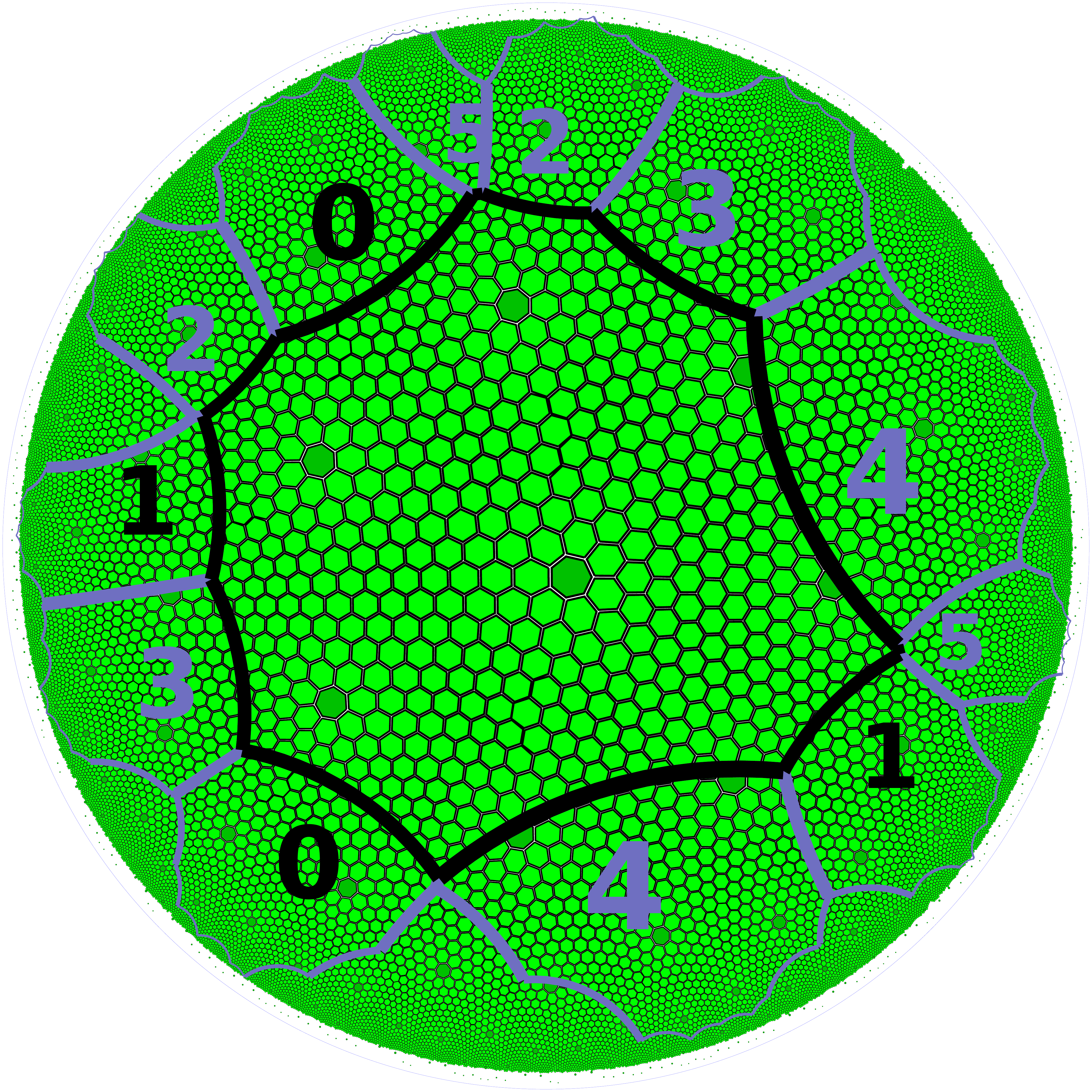}  \hskip -1mm
\subfig{0.48\linewidth}{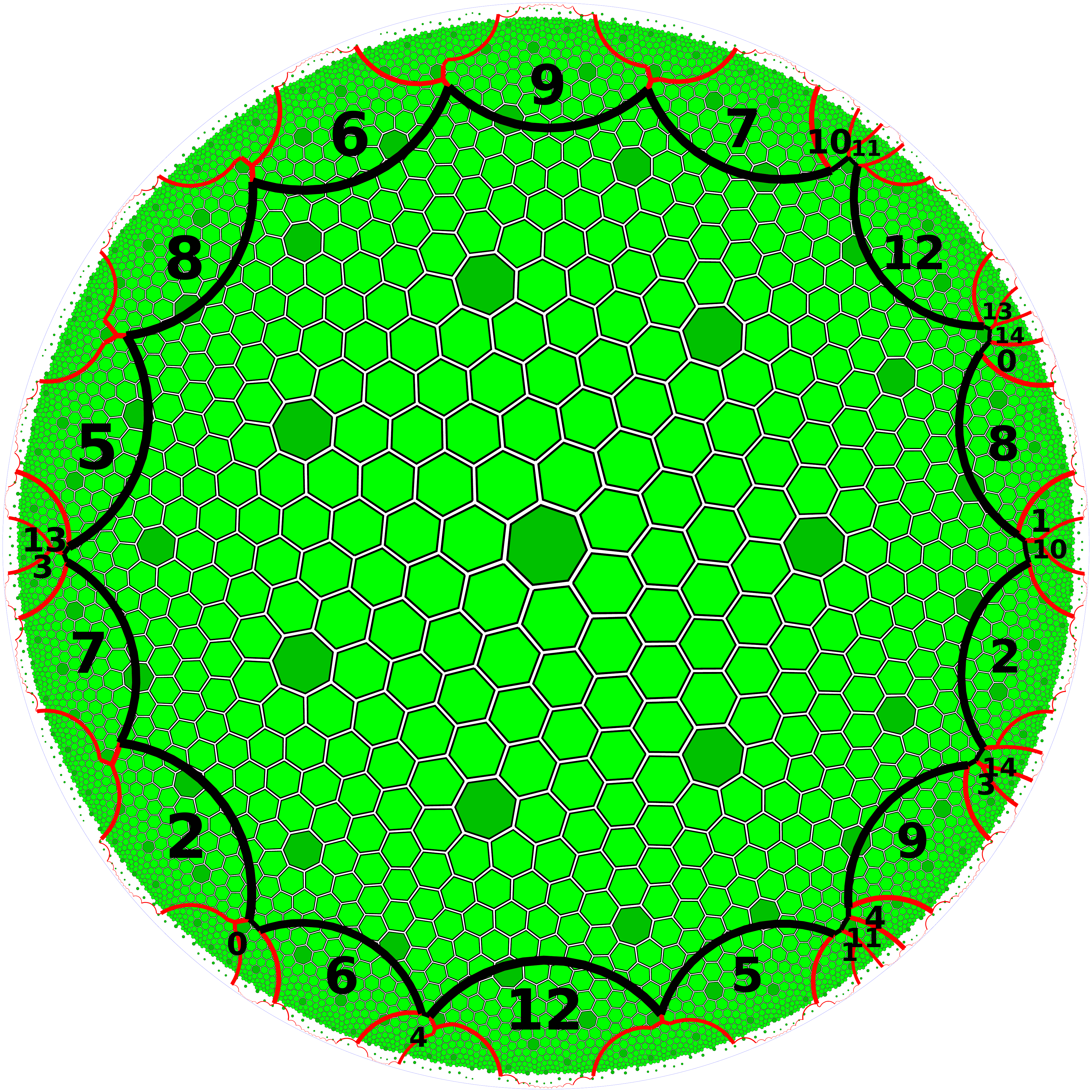}}
\caption{\label{quoths} Fundamental domains: a) torus; b) elliptic plane\longonly{; c) minimal quotient; d) Klein quartic}}
\end{figure}
}

\longonly{Figs. \ref{quoths} depict the fundamental domains for the mentioned quotient spaces. 
Heptagons and pentagons are colored with a darker shade of green. We use the Goldberg-Coxeter construction to add extra hexagons to our tessellation.
A fundamental domain is a subset of the covering space which contains one element of every set of identified points; 
intuitively, we obtain we quotient space by gluing the edges of the fundamental domain. The edges we should glue together are marked with the same numbers;
in Figure \ref{quoths} (c), gray numbers denote that the edge should be reversed first (like in the M\"obius band).}


\subsection{Dispersion}

The natural interpretation of the dispersion function mentioned in the Prerequisities section no longer works in non-Euclidean geometry.
In particular, the exponential nature of the hyperbolic plane makes the random walk
process behave very differently in larger time frames (see, e.g., \cite{heat_kernel} for a~study of heat conduction in the hyperbolic plane). For example, it is well known that the random walk on a~two-dimensional Euclidean grid returns to the starting point (and any other point) with
probability 1. In a~two-dimensional hyperbolic grid, this probability decreases exponentially with distance. Interestingly, Ontrup and Ritter \cite{ritter99,ontrup} who originally introduced non-Euclidean SOMs did not discuss this issue. In applications we may also use quotient spaces, which changes the situation even further -- the information particle could reach the same neuron $j$ in many different ways (e.g., by going left or by going right).

For that reason, we use a~different dispersion function, based on numerically simulating the random walk process on our manifold. We compute the probability $P_{i,j,t}$ that the information particle starting in neuron $i$ will end up in neuron $j$ after $t$ time steps.
This probability can be computed with a~dynamic programming algorithm: for $t=0$ we have $P_{i,j,t} = 1$ if and only if $i=1$ and 0 otherwise;
for $t+1$ we have $P_{i,j,t+1} = P_{i,j,t} + p \sum_k P_{i,k,t} - P_{i,j,t}$, where we sum over all the neighbors $k$ of the neuron $j$.
See Algorithm \ref{dispalg} for the pseudocode which computes $P_{i,j,t}$. In this pseudocode, $N(i)$ denotes the neighborhood of the neuron $i$.
This algorithm has time complexity $O(n^2T)$. Our application is based on highly symmetric surfaces, which lets us to 
reduce the time complexity by taking advantage of the symmetries. 
\longonly{For example, on the torus, we can reduce the time complexity of $O(nT)$,
since $P_{i,j,t} = P_{i',j',t}$ if the transition vector between the neuron $i$ and neuron $j$ is the same as the transition vector between $i'$ and $j'$.
A Hurwitz surface of Euler characteristic $\gamma$ has $42|\gamma|$ symmetries, allowing us to reduce the time complexity by this factor.}

\begin{algorithm}[tb]
\textbf{Parameter}: the set of all nodes of a network $V$; neighborhoods $N(i)$; the number of time steps $T$ and precision $p$\\
\textbf{Output}: the dispersion array $P_{i,j,t}$ for $t=0,\ldots,T$

\begin{algorithmic}[1]
\FOR{$i, j \in V$}
  \STATE $P_{i,j,t} := 0$
  \ENDFOR
\FOR{$i \in V$}
  \STATE $P_{i,i,t} := 1$
  \ENDFOR
\FOR{$t=0,\ldots,T-1$}
  \FOR{$i, j \in V$}
    \STATE $P_{i,j,t+1} := P_{i,j,t}$
    \ENDFOR
  \FOR{$k \in N(i)$}
    \STATE $P_{i,j,t+1} := P_{i,j,t+1} + p \cdot (P_{i,k,t} - P_{i,j,t})$
    \ENDFOR
  \ENDFOR
\end{algorithmic}

\caption{Dispersion algorithm \label{dispalg}.}
\end{algorithm}

In iteration $t$, the weights are updated for every neuron $j$ according to the formula:
$
w_j := w_j + \frac{\eta P_{i,j,f(t)}}{\max_t P_{i,j,t}}(x_t - w_j),
$
where $i$ is the winning neuron, and $f(t) = T (1-\frac{t}{t_{max}})^\longonly{s}\shortonly{2}$.
\longonly{We take $s=2$ to make the dispersion radius scale linearly with time, similar to the Gaussian formula.}


\section{Example visualizations of our results}
To visualize the result of the proposed algorithm we will use the classic iris flower dataset by Fisher \cite{iris} and the palmerpenguins dataset \cite{palmerpenguins}. 
Figure \ref{kohonen} depicts the visual example result of the SOM clustering. Coloring of the tiles allows for the examination of the clusters. We utilize a~standard tool: inverted U-matrix (unified distance matrix) where the
Euclidean distance between the representative vectors of neurons in the neighborhood is depicted in grayscale. The darker the shade, the less dissimilarity there is within
the neighborhood. We may see the smooth and gradual changes.

\longonly{We compare various setups for SOMs (keeping similar number of tiles). 
Figures \ref{kohonen}(a,b) depict the results for a standard SOM setup (Euclidean plane tiled with squares) as a benchmark. Note that the results are
rather unsatisfactory -- SOM with Euclidean setup does not combat the known issue of mixed observations in two of the groups, moreover, for penguins dataset
it also suggests that there are more than three groups (see the boundary within violet observations). On the contrary, SOMs based on our setup perfom visually better
Redefining neighborhood by introducing hexagons and heptagons helps in minimizing the intermixing. Moreover, one can see that setups with closed manifolds (Figures c-f)
lead to a better visual distinction of the edges among the groups. Especially in the case of Klein quartic, due to the exponential growth, we can fit similar objects closer to each other than it would be possible in the Euclidean plane.}


\longonly{
\begin{figure}[!h]
\centering 
\subfig{0.45\linewidth}{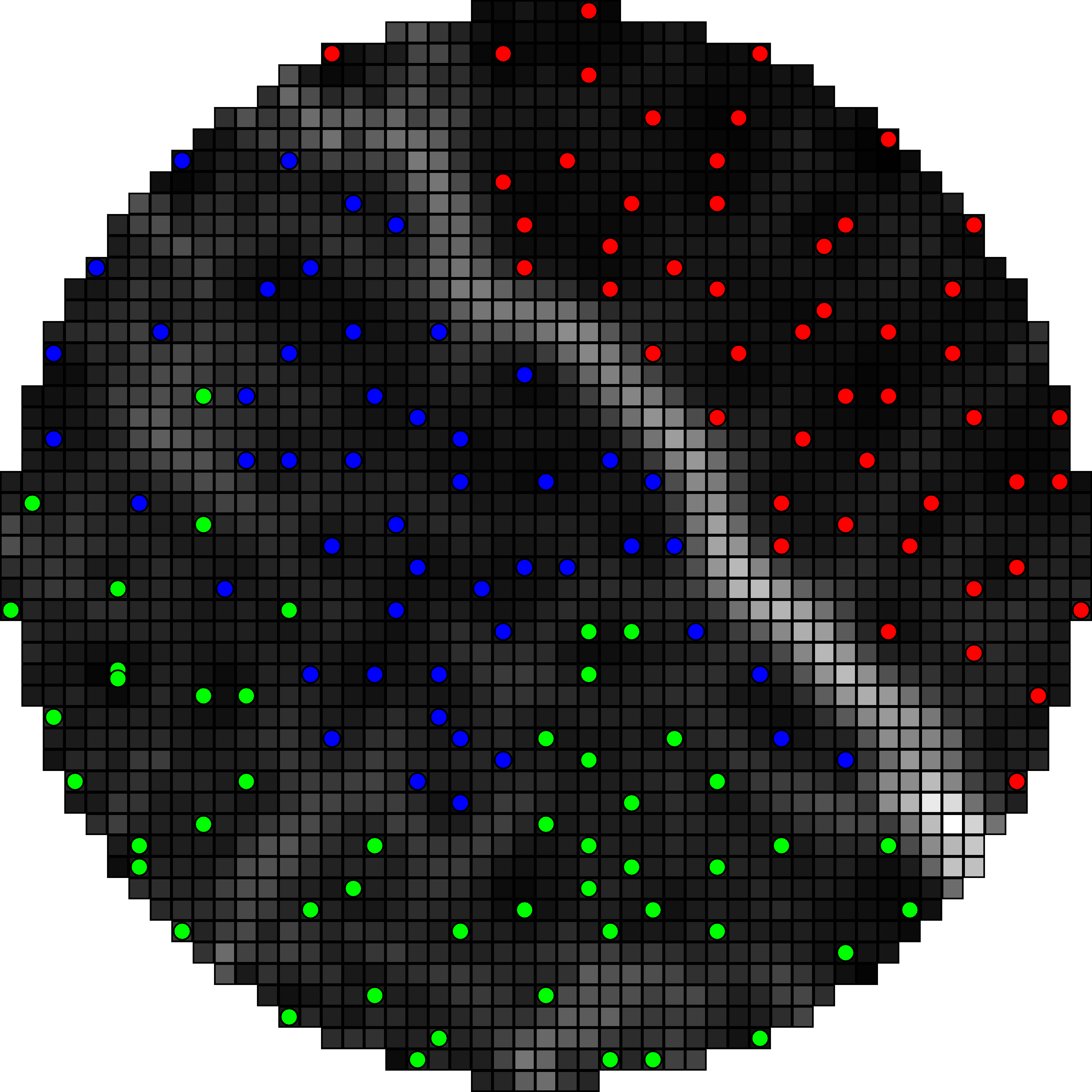}
\subfig{0.45\linewidth}{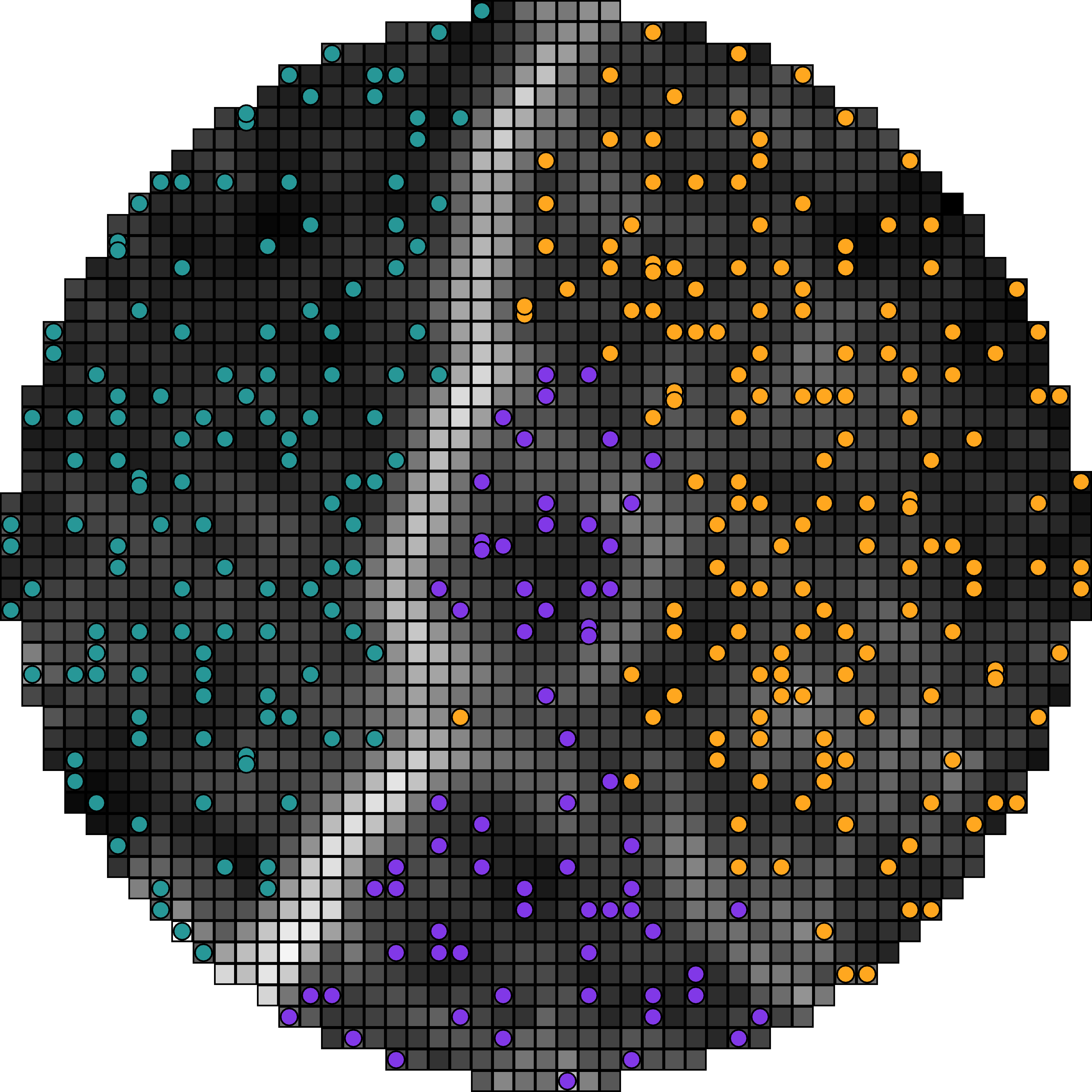} \\
\subfig{0.45\linewidth}{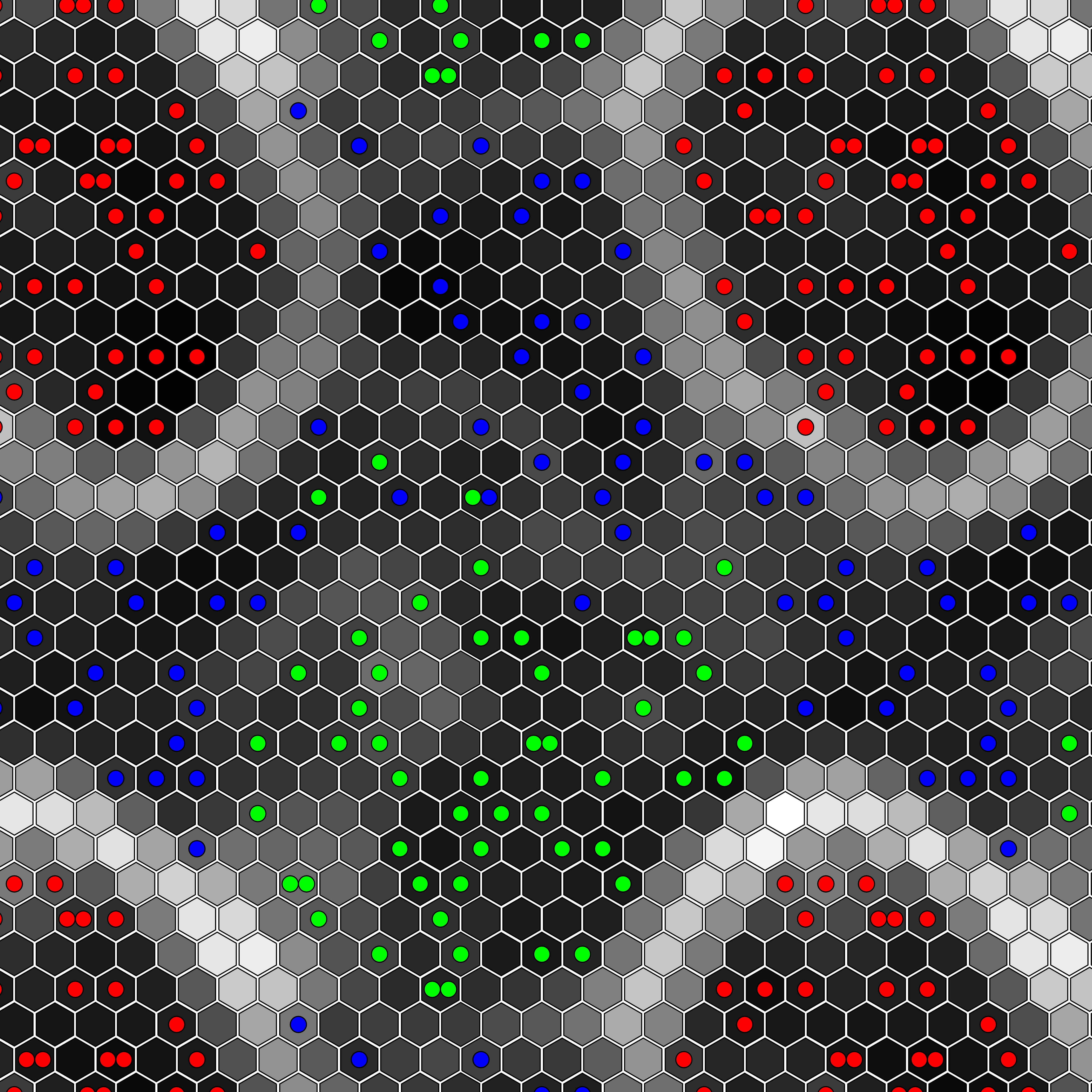}
\subfig{0.45\linewidth}{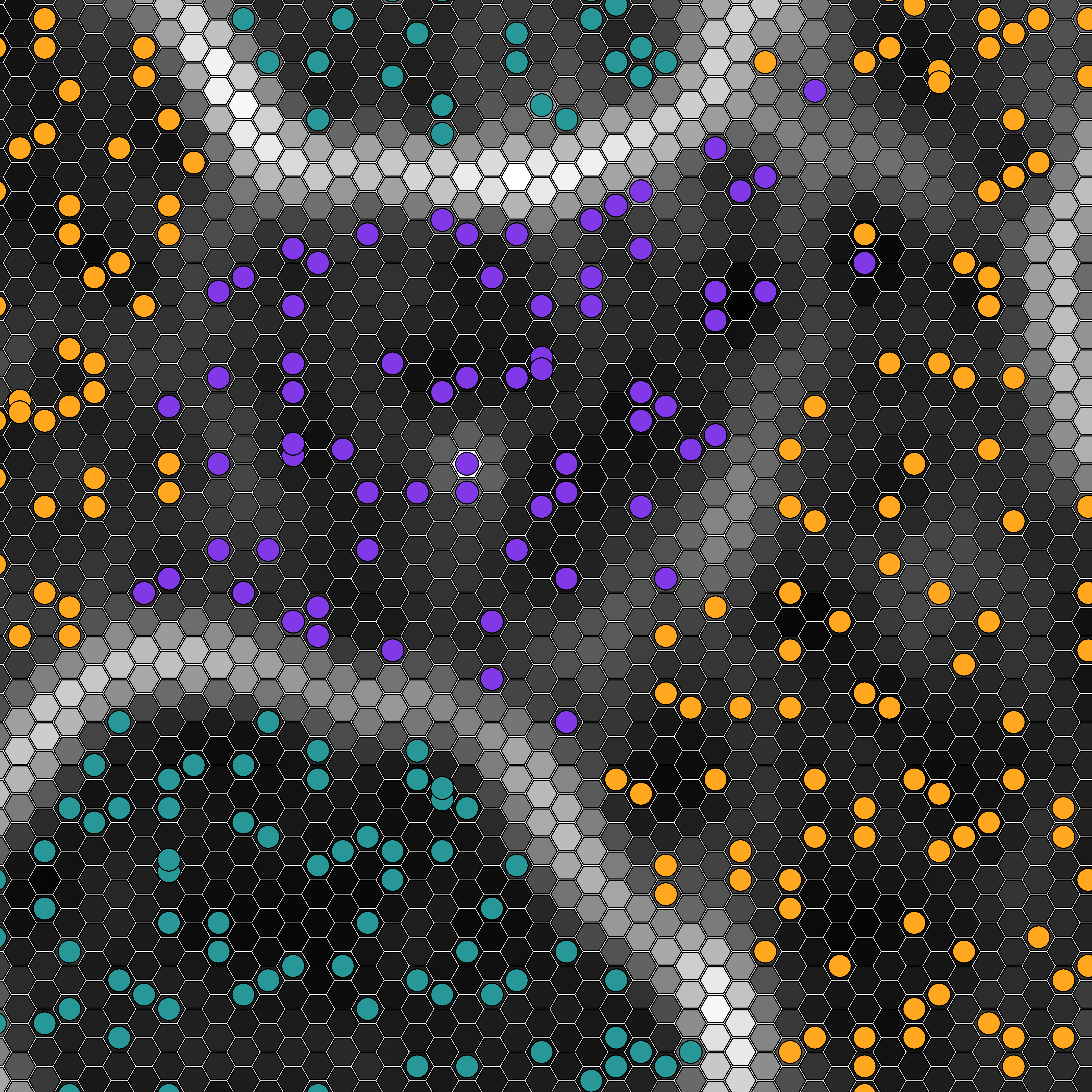} \\
\subfig{0.45\linewidth}{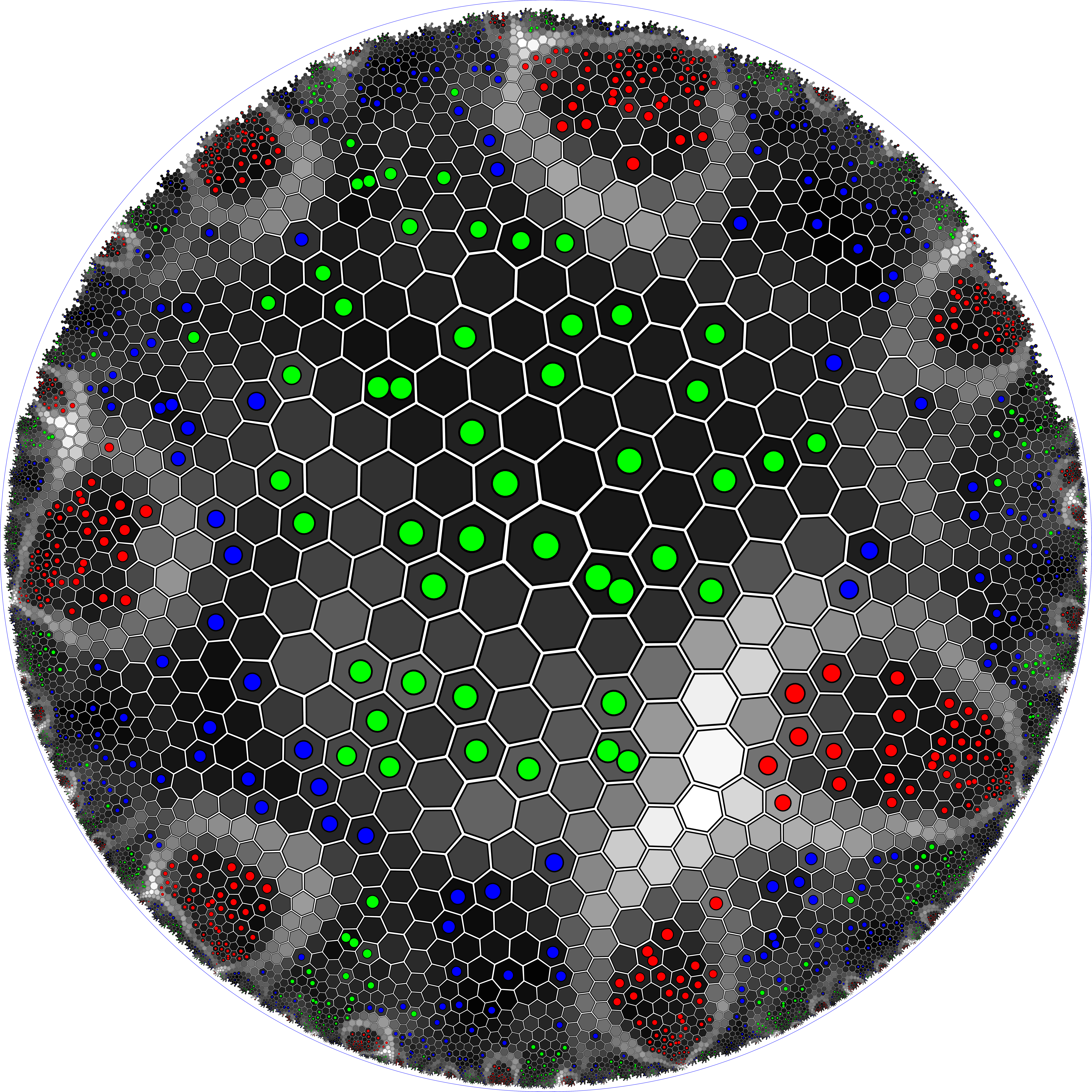}
\subfig{0.45\linewidth}{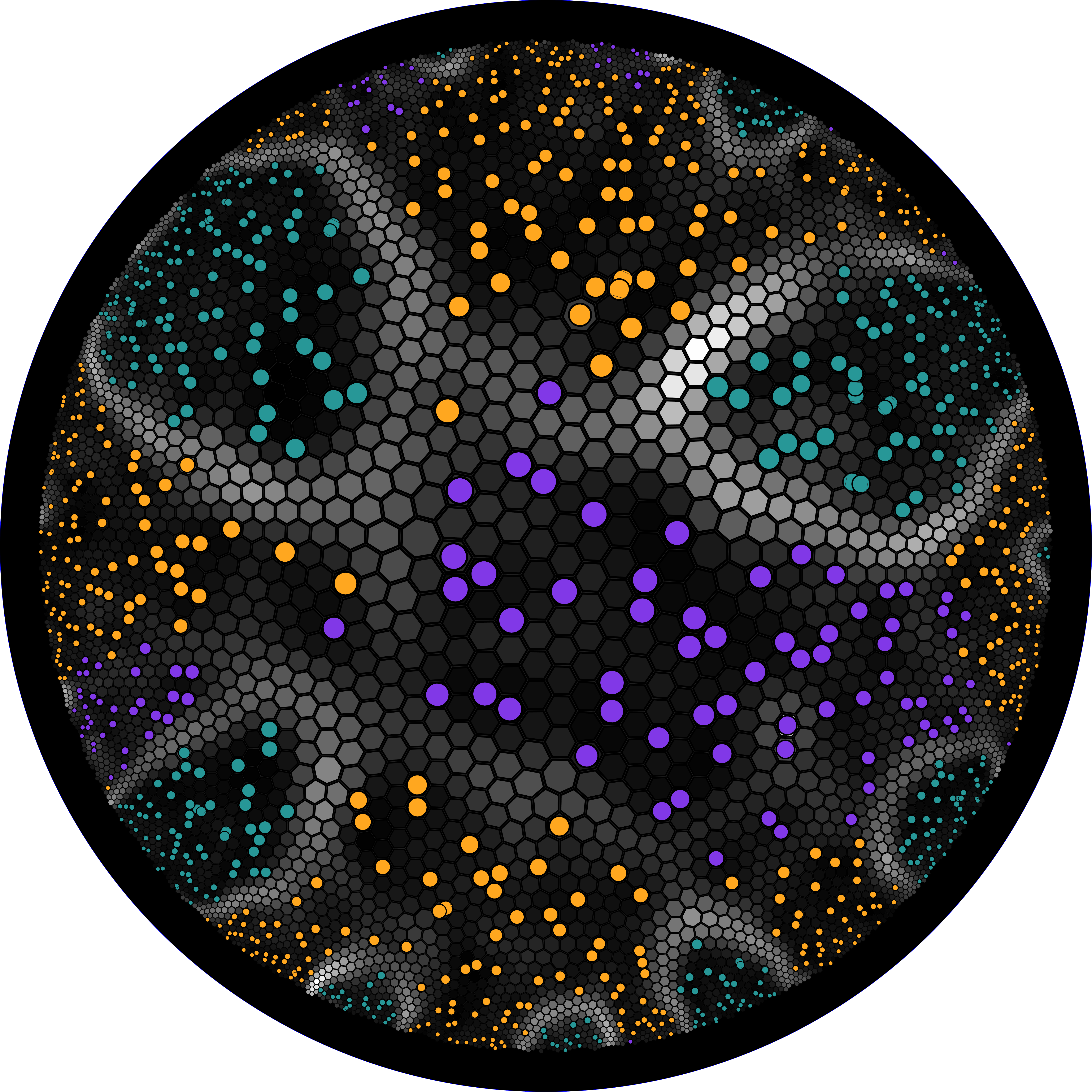} \\
\subfig{0.45\linewidth}{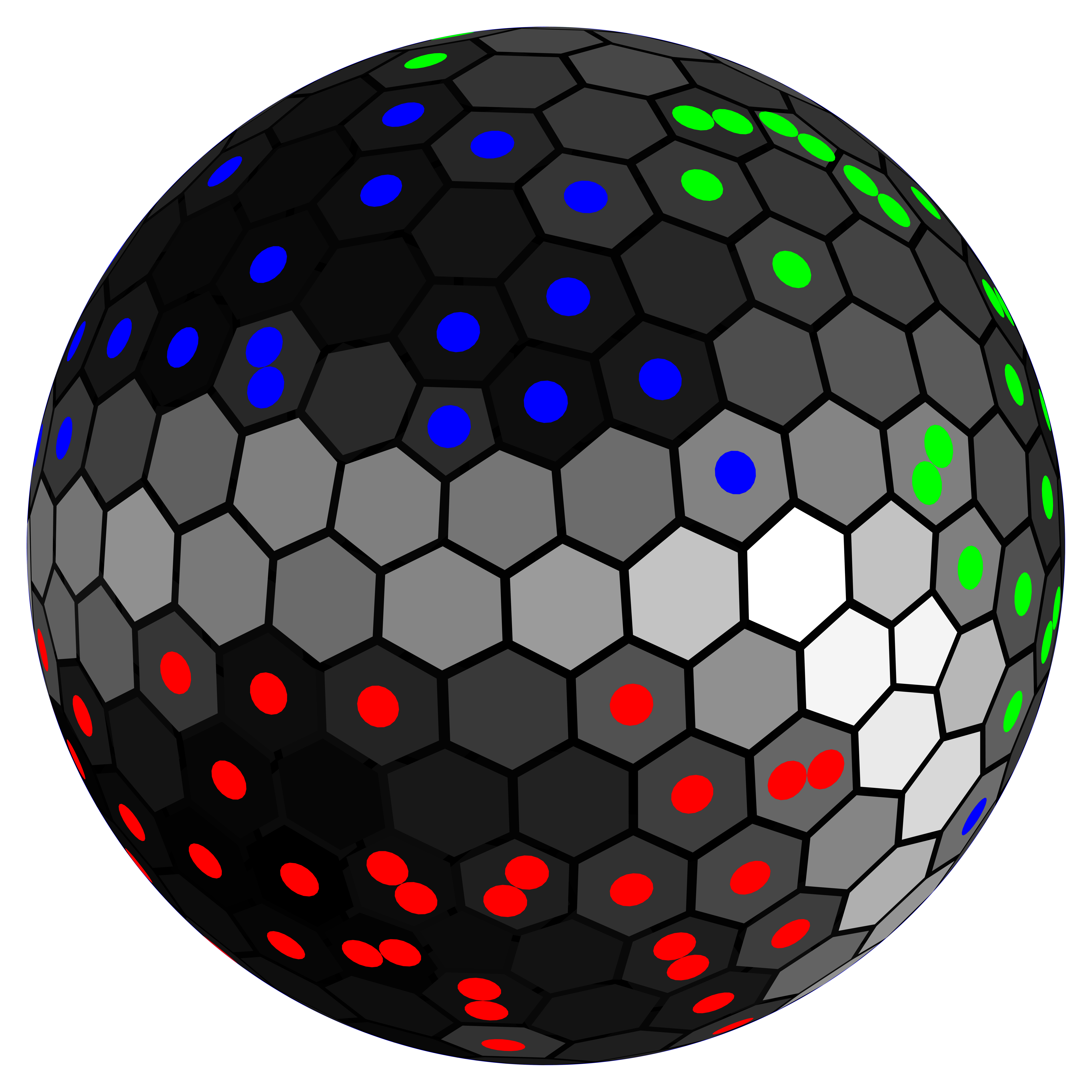} 
\subfig{0.45\linewidth}{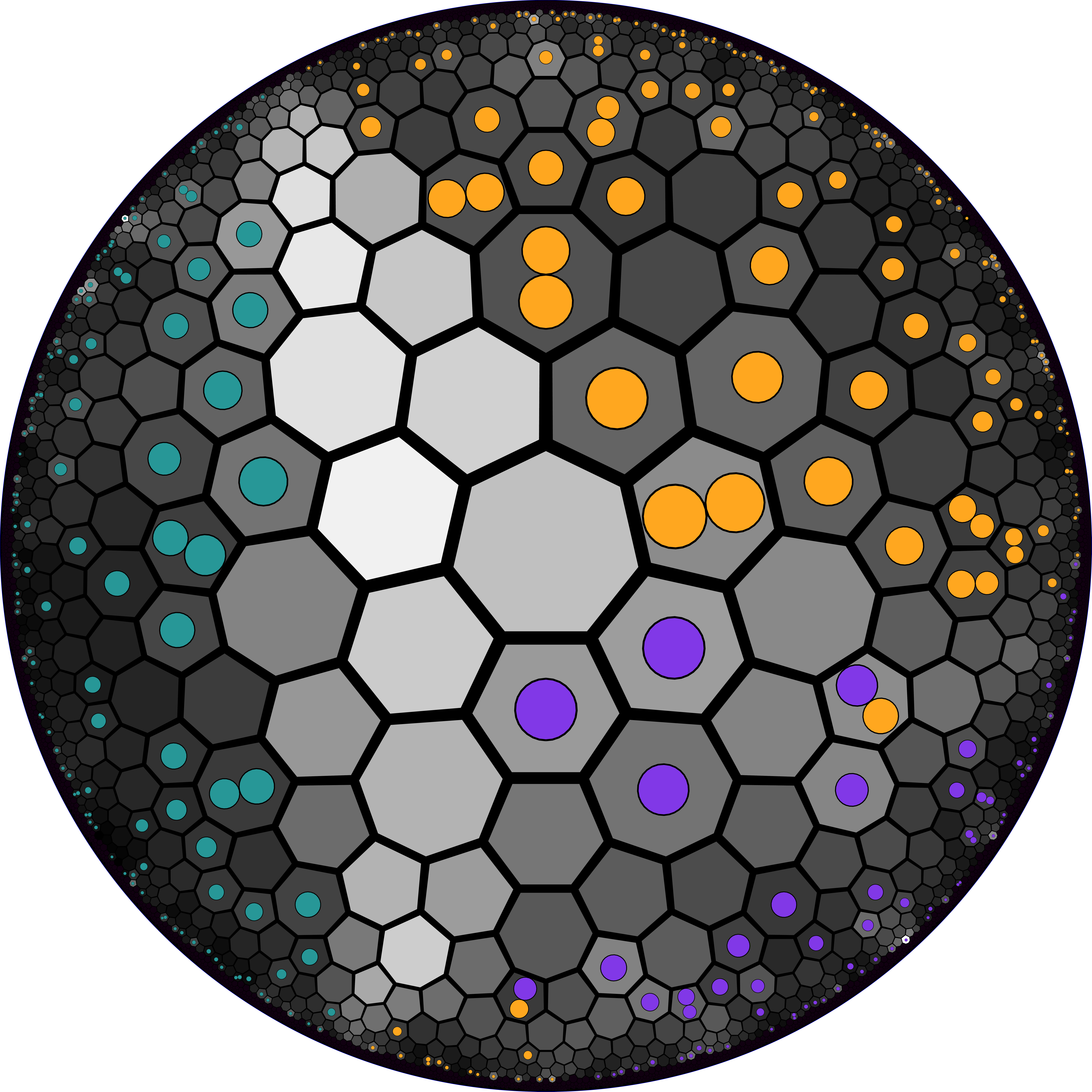} 
\caption{Example results of SOM on the iris flower (aceg) and palmerpenguins (bdfh) dataset.
(ab) a disk on the Euclidean square grid, (cd) a torus with the hex grid, (ef) Klein quartic, (g) sphere in orthographic projection, (h) a hyperbolic disk.
 \label{kohonen}}
\end{figure}
}
\shortonly{
\begin{figure}[!h]
\centering 
\subfig{0.31\linewidth}{img/penguins-torus.pdf}
\subfig{0.31\linewidth}{img/iris-kq.pdf}
\subfig{0.31\linewidth}{img/penguins-hdisk.pdf} 
\caption{Example results of SOM on the iris flower (b) and palmerpenguins (ac) dataset.
(a) a torus\longonly{ with the hex grid}, (b) Klein quartic, (c) a disk in $\bbH^2$.
\label{kohonen}}
\end{figure}
}
While static visualizations work for Euclidean geometry, 
hyperbolic geometry is useful for visualization of hierarchical data, where we can focus by centering the visualization on any node
in the hierarchy \cite{lampingrao,munzner}. Our visualization engine lets the viewer to
smoothly change the central point of the projection to any point in the manifold, and thus clearly see the spatial relationships of every cluster.

The locations and the neighborhoods returned by SOMs have interpretation. Given that competition and adaptation stages force the neighborhood to attract similar objects,
 the distance between the neurons becomes a~measure for similarity: the further, the less similar objects are. We may use the resulting classification and mapping in further analyses.

\section{Experiments}

Our general experimental setup is as follows. 

\begin{itemize}
\item We construct the original manifold $O$. Let $T_O$ be the set of tiles and $E_O$
be the set of edges between the tiles.
\item We map all the tiles into the Euclidean space $m: T_O \ra \bbR^d$, where $d$ is
the number of dimensions.
\item We construct the target embedding manifold $E$. Let $T_E$ be the set of tiles
and $E_E$ be the set of edges between the tiles.
\item We apply our algorithm to the data given by $m$, This effectively yields an
embedding $e: T_O \ra E_O$.
\item We measure the quality of the embedding.
\end{itemize}

To limit the effects of randomness (random initial weight of neurons, random ordering of data) 
we apply this process independently 100 times for every pair of manifolds\longonly{ $E$ and $O$}.

\shortonly{
\begin{figure}
\centering
\subfig{0.24\linewidth}{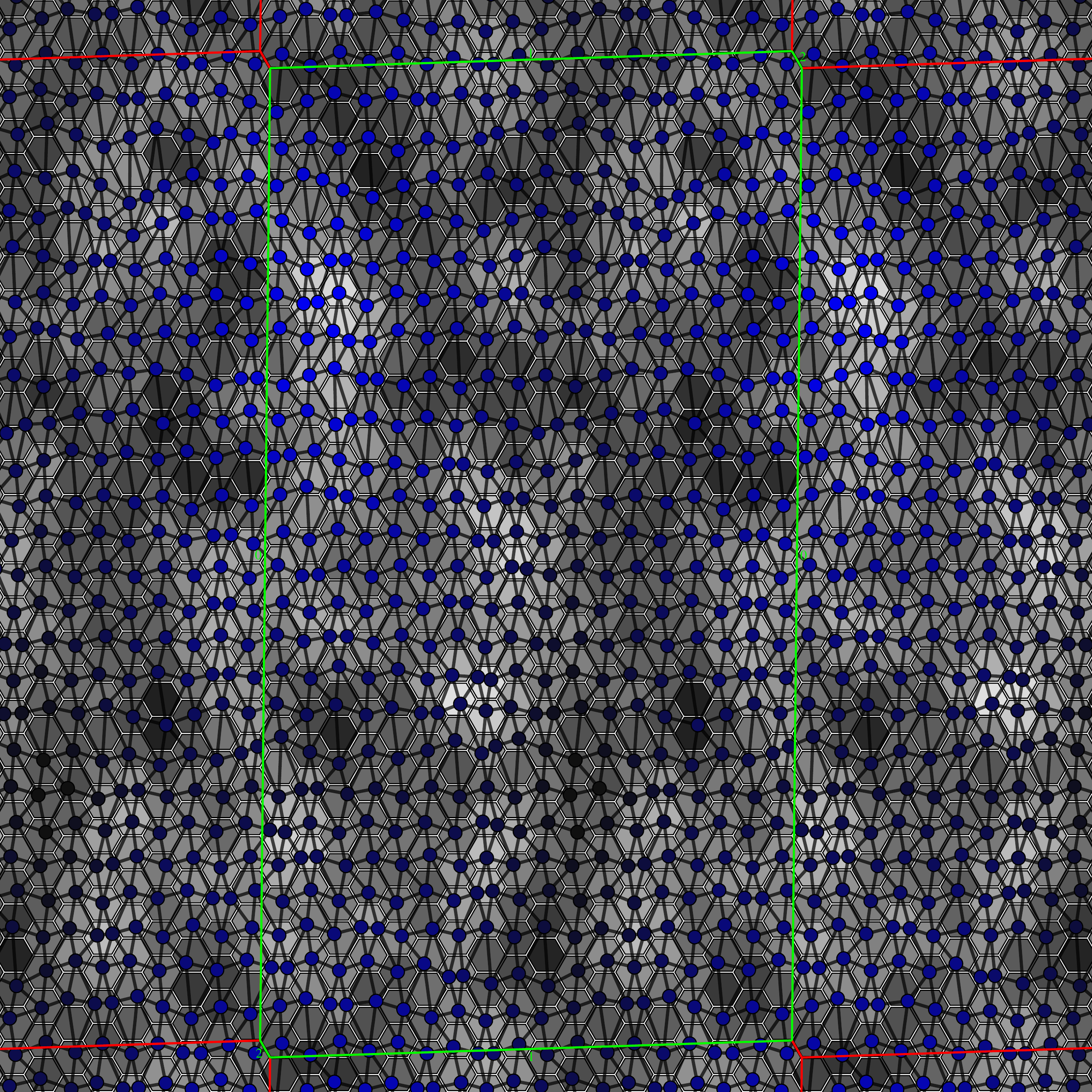} \hskip -1mm
\subfig{0.24\linewidth}{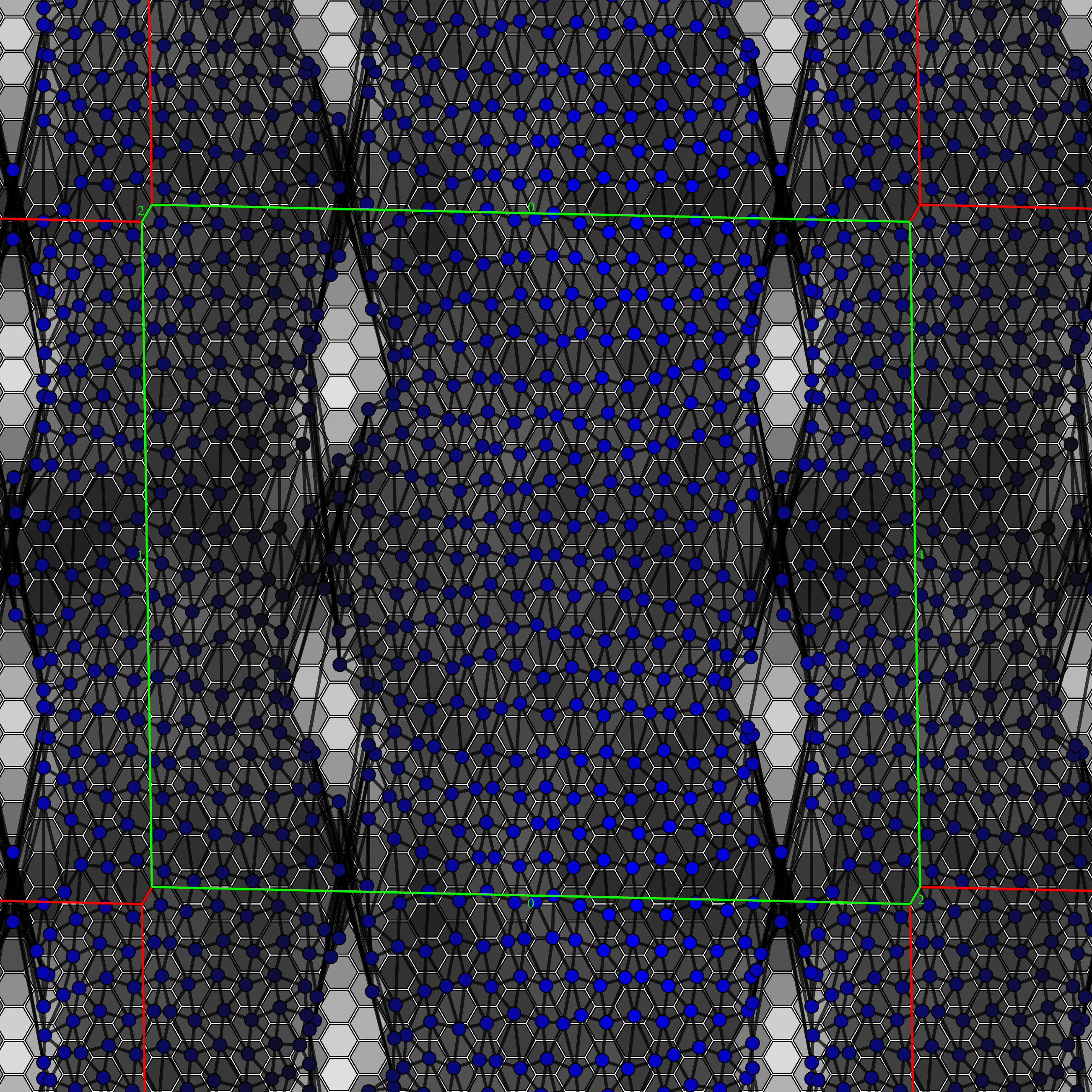} \hskip -1mm
\subfig{0.24\linewidth}{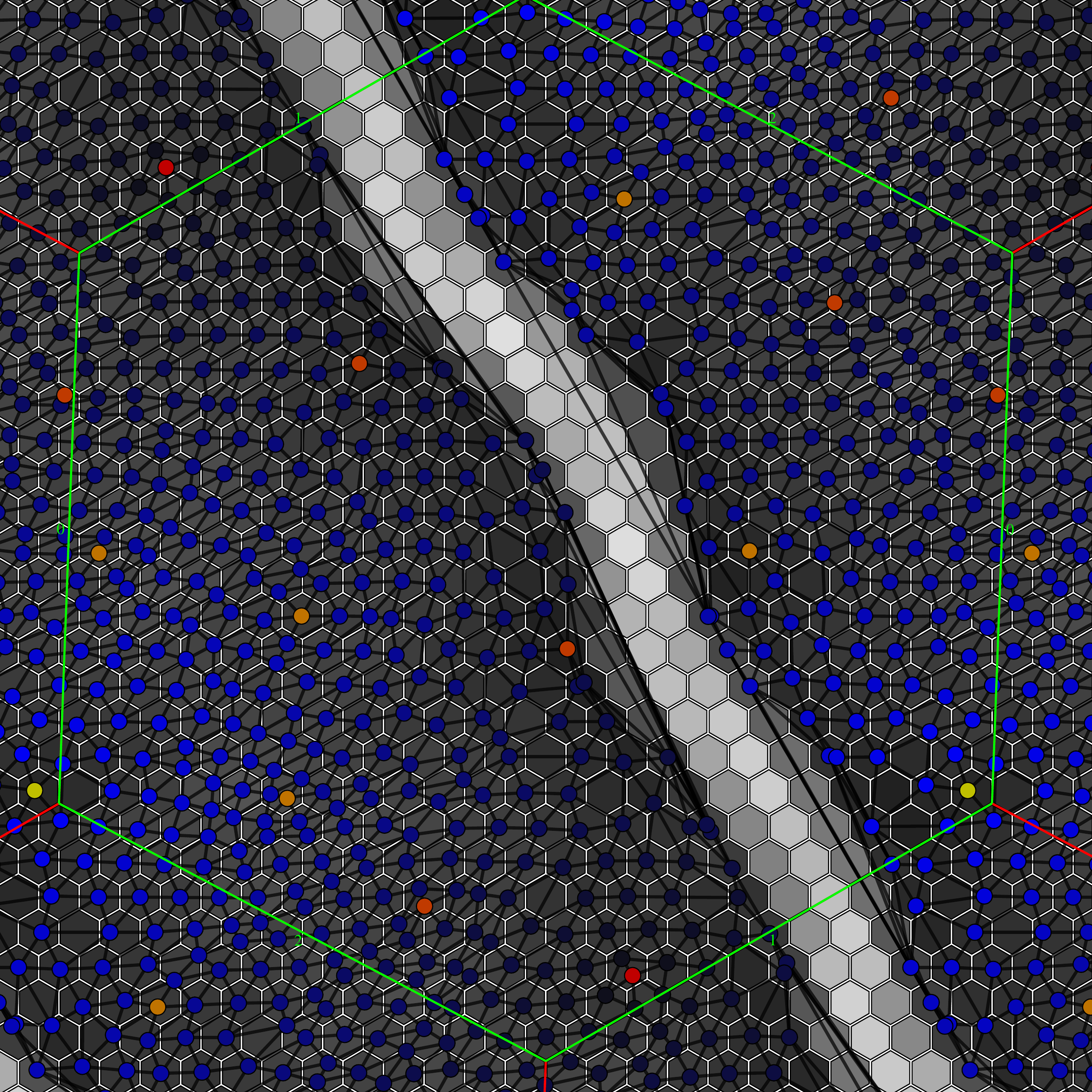} \hskip -1mm
\subfig{0.24\linewidth}{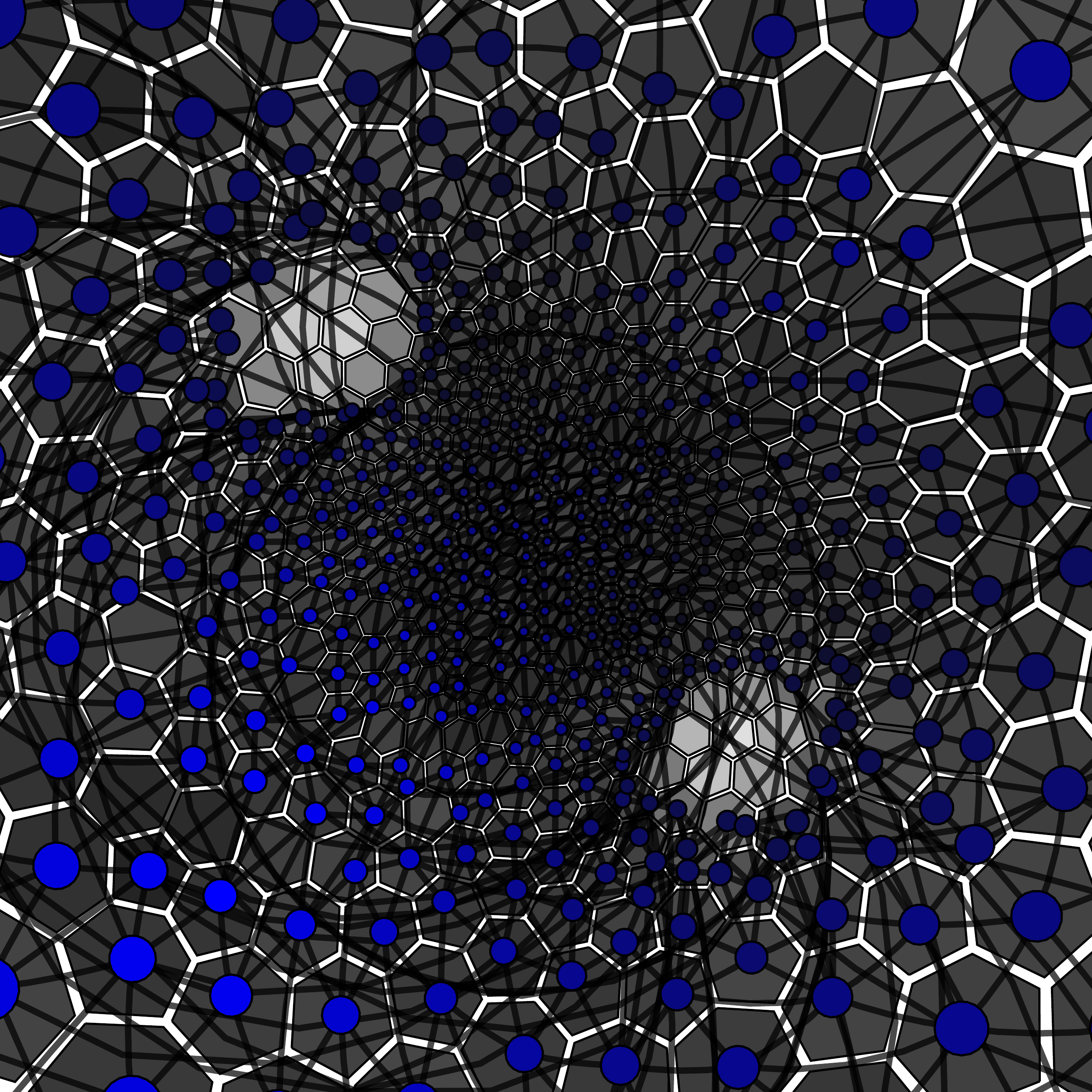}
\caption{\label{retrieval}
(a)~square torus to rectangular torus; 
(b)~rectangular torus to square torus;
(c)~sphere \longonly{mapped }to hex torus; (d)~hex torus \longonly{mapped }to sphere}
\end{figure}
}
\longonly{
\begin{figure*}
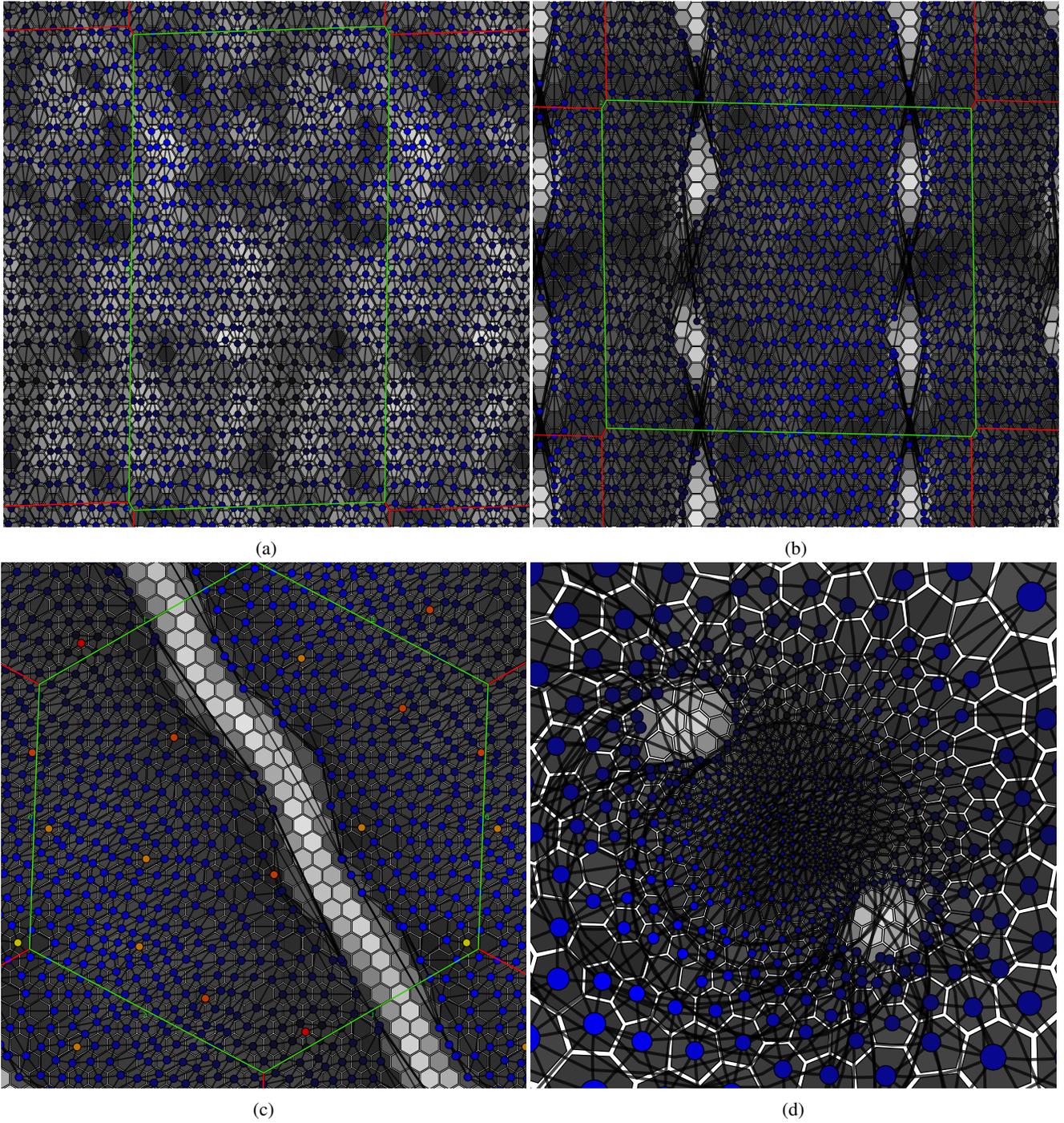

\centering
\subfig{0.49\linewidth}{img/well-mapped-torus.pdf} \hskip -1mm
\subfig{0.49\linewidth}{img/badly-mapped-torus.pdf} \hskip -1mm
\subfig{0.49\linewidth}{img/sphere-to-torus.pdf} \hskip -1mm
\subfig{0.49\linewidth}{img/torus-to-sphere.pdf}
\caption{\label{retrieval}
(a)~square torus to rectangular torus; 
(b)~rectangular torus to square torus;
(c)~sphere mapped to hex torus; (d)~hex torus mapped to sphere}
\end{figure*}
}

In Figure \ref{retrieval} the effects of four runs are shown. Small gray polygons
(hexagons) represent the tiles of $E$. The green and red polygons depict the fundamental
domain. Every circle represents a tile from $T_O$ that has been mapped to the
manifold $E$, to the tile shown in the visualization. Edges between circles correspond
to the orignal edges $E_O$ between them.

In Figures \ref{retrieval}a and \ref{retrieval}b, both $E$ and $O$ are tori of
different shapes. In the run shown in Figure \ref{retrieval}a we obtain what we consider
a successful map: the topology of the data is recovered correctly (despite the different
shape of the two tori). Figure \ref{retrieval}b
shows an unsuccessful recovery of topology. In this case, the original torus $O$ has been cut into
two cylinders $O_1$ and $O_2$, which are respectively mapped to cylinders $E_1$ and $E_2$
in $E$; however, the two maps $O_1 \ra E_1$ and $O_2 \ra E_2$ are mirrored. 
This issue is clearly visible in our visualization: parts of the boundary areas between
$E_1$ and $E_2$ contain no tiles, and the edges show singularities.

Figures \ref{retrieval}c and \ref{retrieval}d show a pair of mappings where $E$ and $O$ have
different topologies (a sphere and a torus). Since the topologies are different, there
is no way to map $T_E$ to $T_O$ without singularities. In Figure \ref{retrieval}c our algorithm
has stretched the sphere $O$ on the poles, obtaining a cylinder; that cylinder is then mapped
to a cylinder obtained by cutting the torus. The torus has been cut along the white area.
Most edges are mapped nicely, to pairs of close cells, but some edges close to the poles 
will have to go around the cylinder. Figure \ref{retrieval}d is the inverse case. The torus
is obtained by removing two disks at the poles, and gluing the boundaries of the removed
disks. Edges which connect the boundaries of the two disks go across the whole sphere,
while the remaining edges have the correct topological structure.


\longonly{
\begin{figure}
\centering
\subfig{0.99\linewidth}{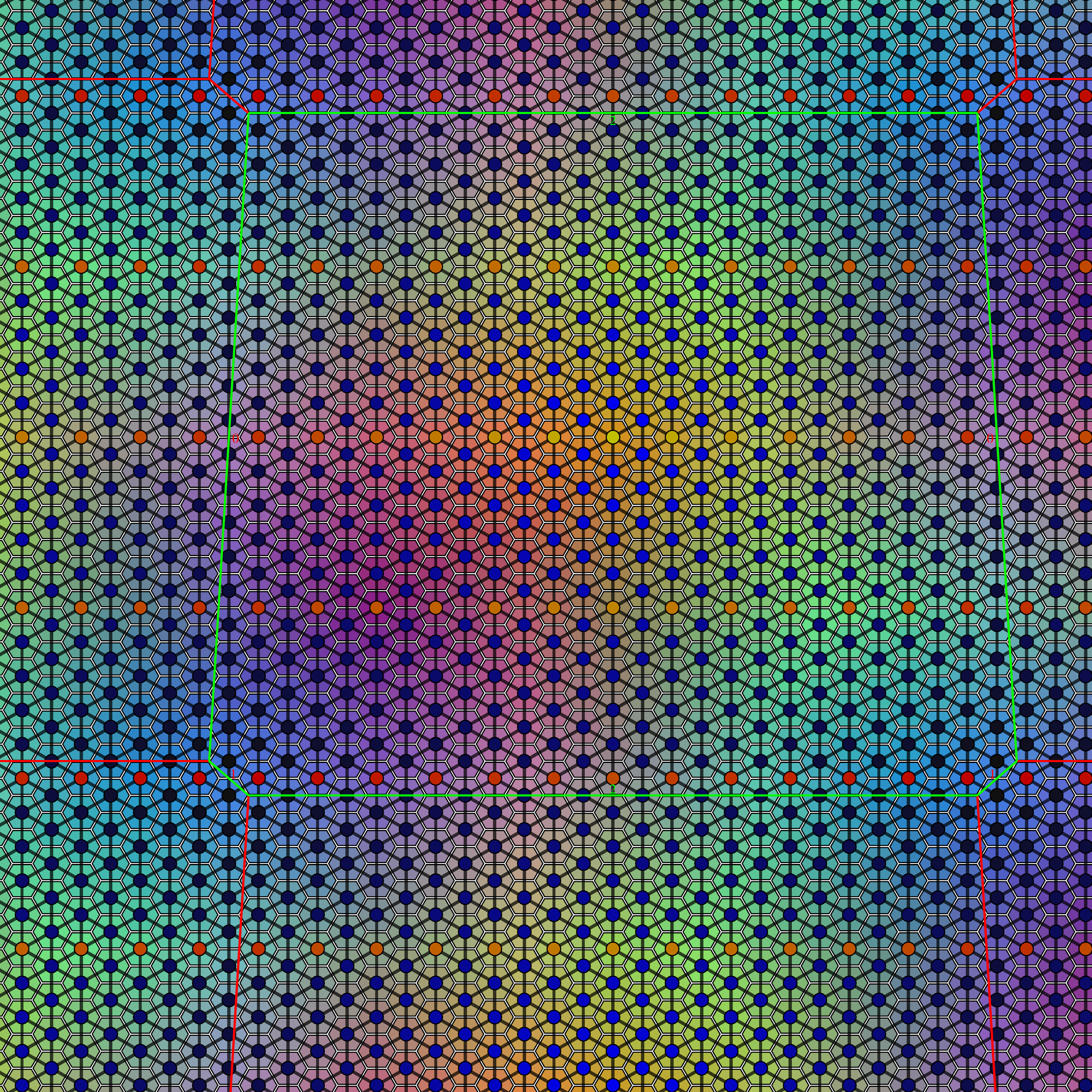} \hskip -1mm
\subfig{0.49\linewidth}{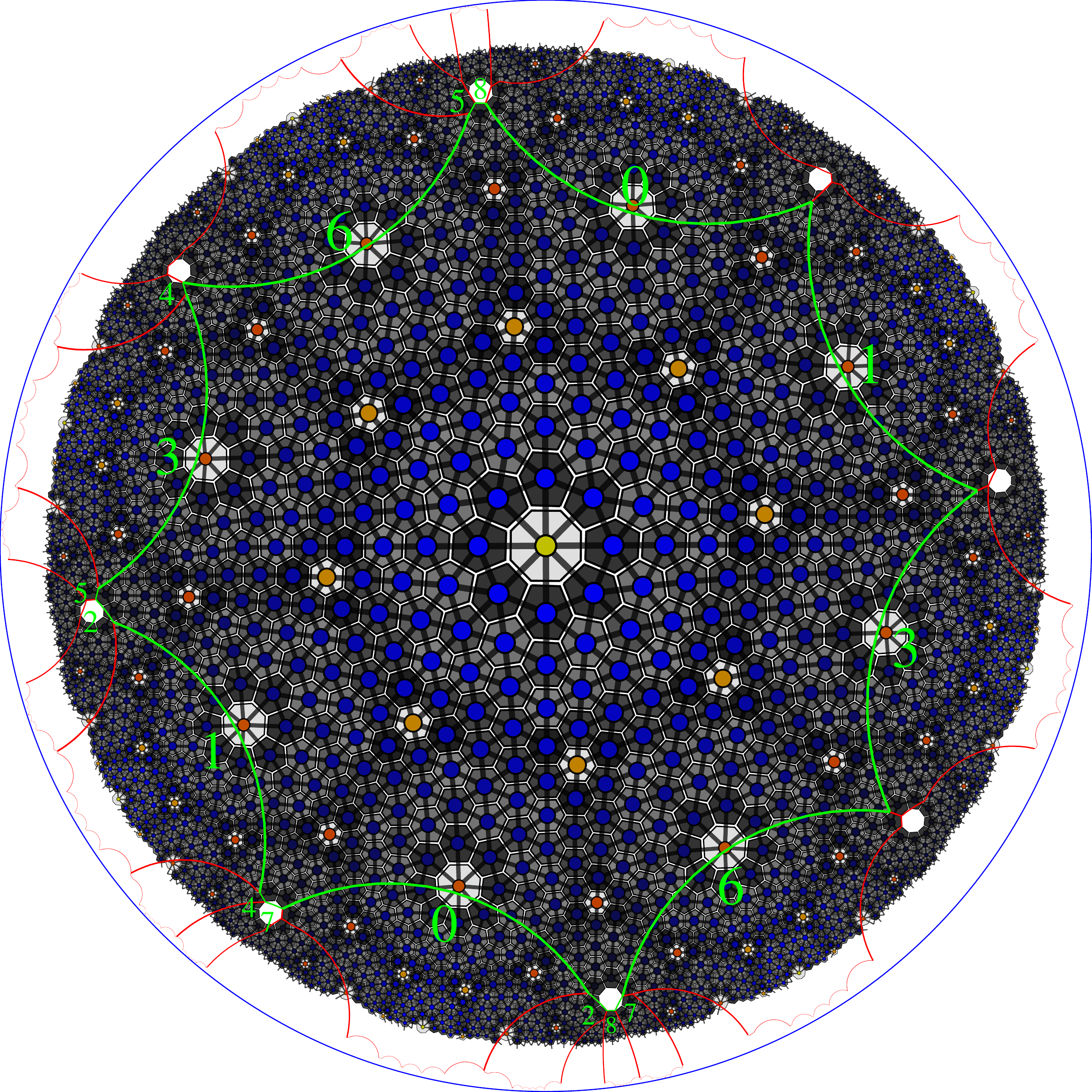} \hskip -1mm
\subfig{0.49\linewidth}{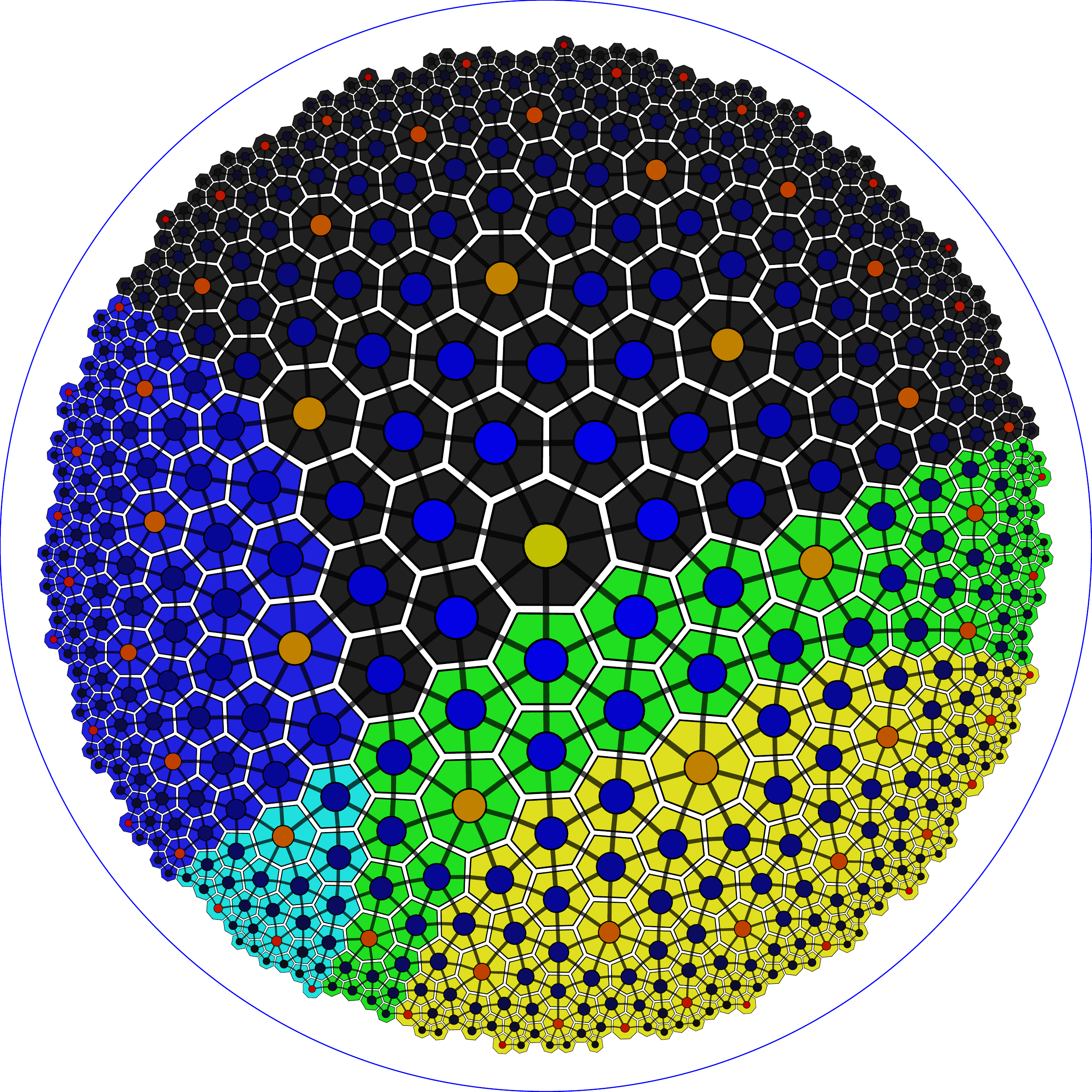} \hskip -1mm
\caption{\label{embs}
(a)~signpost on Klein bottle; 
(b)~signpost on Bolza surface;
(c)~landscape on the disk}
\end{figure}
}

\longonly{
\subsection{List of manifolds}
\begin{table*}[h]
\resizebox{\textwidth}{!}{
\begin{tabular}{l|rrrrrrrrrrrrrrrrrrrrr}
name&$n$&edges&col&embtype&$a$&$b$&o&s&c&q&d&q&$p$&curvature&$\chi$&$A$&g&m&avgdist&avgdist2&kmax\\ \hline
disk10&520&1214&60&landscape&1&0&1&0&0&0&2&3&7&-1&---&1&h&10&7.37242&58.5336&0.847069\\
disk11&520&1326&60&landscape&1&1&1&0&0&0&2&3&7&-0.313462&---&3&h&14&9.21862&94.4243&0.904728\\
disk20&520&1340&60&landscape&2&0&1&0&0&0&2&3&7&-0.217308&---&4&h&16&9.58927&103.118&0.914314\\
disk21&520&1381&60&landscape&2&1&1&0&0&0&2&3&7&-0.136538&---&7&h&18&10.3977&123.463&0.928886\\
disk40&520&1409&60&landscape&4&0&1&0&0&0&2&3&7&-0.0557692&---&16&h&22&11.0977&143.849&0.940335\\
disk43&520&1445&60&landscape&4&3&1&0&0&0&2&3&7&-0.0153846&---&37&h&24&11.5947&160.277&0.947179\\
disk-euclid&520&1479&60&landscape&1&0&1&0&0&0&2&3&6&0&0&1&e&26&11.9668&174.804&0.951944\\
elliptic&541&1620&6&signpost&6&6&0&1&1&1&2&3&5&0.0110906&1&108&s&15&9.43387&101.41&0.918638\\
sphere&522&1560&3&natural&6&2&1&1&1&0&2&3&5&0.0229885&2&52&s&20&10.1551&121.663&0.937623\\
sphere4&510&1524&3&natural&7&6&1&1&1&0&2&3&4&0.0235294&2&127&s&20&9.94615&116.516&0.93592\\
torus-hex&529&1587&6&natural&1&0&1&1&1&1&2&3&6&0&0&1&e&15&8.95455&90.7727&0.91323\\
torus-sq&520&1560&4&natural&1&0&1&1&1&1&2&3&6&0&0&1&e&16&8.96724&91.5915&0.916843\\
torus-rec&522&1566&4&natural&1&0&1&0&1&1&2&3&6&0&0&1&e&19&9.74856&112.363&0.934869\\
klein-sq&520&1560&52&signpost*&1&0&0&0&1&1&2&3&6&0&0&1&e&16&8.93757&90.7337&0.915332\\
Bolza&502&1512&22&signpost*&6&3&1&1&1&1&2&3&8&-0.0239044&-2&63&h&15&8.0759&73.1651&0.900291\\
Bolza2&492&1488&12&signpost&5&1&1&1&1&1&2&3&8&-0.0487805&-4&31&h&12&7.42308&60.7784&0.87572\\
minimal&524&1575&6&signpost&5&5&0&0&1&1&2&3&7&-0.0114504&-1&75&h&17&8.7837&87.8727&0.915569\\
Zebra&516&1554&12&signpost&4&3&1&0&1&1&2&3&7&-0.0232558&-2&37&h&16&8.76821&88.2358&0.919384\\
KQ&528&1596&24&signpost&3&2&1&1&1&1&2&3&7&-0.0454545&-4&19&h&13&7.73279&65.8849&0.877095\\
Macbeath&576&1764&72&signpost&2&1&1&1&1&1&2&3&7&-0.125&-12&7&h&13&7.12609&55.667&0.862014\\
triplet1&520&1638&156&signpost&1&1&1&1&1&1&2&3&7&-0.3&-26&3&h&9&5.89461&37.4094&0.816052\\
triplet2&520&1638&156&signpost&1&1&1&1&1&1&2&3&7&-0.3&-26&3&h&11&5.97726&38.8343&0.832687\\
triplet3&520&1638&156&signpost&1&1&1&1&1&1&2&3&7&-0.3&-26&3&h&10&5.88092&37.3156&0.818189\\
\end{tabular}
}
\caption{The list of manifolds in the experiment. $n$=neurons/samples, col=columns, o=orientable (1=TRUE), s=symmetric (1=TRUE), q=quotient (1=TRUE), c=closed (1=TRUE), d=dimension, 
$a,b$ -- Goldberg-Coxeter parameters, $p,q$ -- Schl\"afli symbol, g=geometry (hyperbolic/euclidean/spherical), m=max tile distance,
$A$=area, $\chi$=Euler characteristics, kmax -- maximum Kendall coefficient.}\label{manilist}
\end{table*}
Table \ref{manilist} lists all manifolds in our experiment.
\begin{itemize}
\item KQ (Klein Quartic), Macbeath, and triplet (first Hurwitz triplet) are Hurwitz manifolds (closed hyperbolic manifolds with underlying \sch{7}{3} tessellation exhibiting very high symmetry.
\item Zebra and minimal are less symmetric manifolds, also with underlying \sch{7}{3} tessellation.
\item Bolza surface has underlying \sch{8}{3} tessellation, and Bolza2 is its double cover. They are also highly symmetric.
\item sphere and sphere4 are spheres with different underlying tilings (\sch{5}{3} and \sch{4}{3}). Elliptic is the elliptic plane (\sch{5}{3}).
\item torus-hex, torus-sq (square torus) and torus-rec (rectangular torus) are tori with different shapes (\sch{6}{3}). 
klein-sq is the Klein bottle.
\item Disks are disks with different Goldberg-Coxeter subdivisions of \sch{7}{3}. Each of them consists of 520 cells closest to the origin.
\end{itemize}
A closed manifold with Euler characteristics $\chi \neq 0$ and underlying $\{p,3\}$ tessellation will have $u=6\chi/(6-p)$ underlying tiles. These tiles will form
$t=pu/3$ triangles. Goldberg-Coxeter construction $\gp{a}{b}$ will replace each of these triangles with 
$\frac A2=\frac{(2a+b)^2 + 3b^2}{4}$ tiles. Thus, the total number of tiles in the manifold equals $n=u + \frac{A-1}2 t$.
We can control $n$ by changing the Goldberg parameters $a$ and $b$. 
However, for the first Hurwitz triplet we have $u=156$, so we do not have much control. We get
$n=520$ for $\gp{1}{1}$, and we adjust $a$ and $b$ for all the other manifolds to have as close $n$ as possible. For non-orientable manifolds only $b=0$ or $b=a$ are legal.
Curvature is defined as $2q/(q-2)-v$, where $v$ is the average number of neighbor tiles (counting tiles outside of the sample in the case of disks).
We consider two manifolds to have same geometry if both are hyperbolic, both are Euclidean or both are spherical.
We consider two closed manifolds to have same topology if they have the same Euler characteristics and orientability.
This happens in the following cases: all disks; all tori; all triplet manifolds; sphere vs sphere4; Bolza vs Zebra; and Bolza2 vs KQ.
}

\longonly{\paragraph{Double density manifolds}
Taking both $E$ and $O$ from the same list of manifolds could potentially cause overfitting. To combat this issue, we also
consider double density manifolds, which are obtained by doubling both Goldberg parameters $a$ and $b$. This increases the number of samples roughly four times
(exactly four times in the cases of disks and tori).}

\subsection{Embedding into $\bbR^d$}\label{sub:embed}

We need to embed the manifold $O$ into $\bbR^d$ in such a way that both its topology and
geometry are preserved, that is, distances in $\bbR^d$ are a good representation of
actual geodesic distances in $O$. We use the following methods.

\paragraph{Natural} Well-known embeddings are known for the following cases:
\begin{itemize}
\longonly{
\item The Euclidean disk $\bbD^2$ has a well-known embedding to $\bbR^2$ (as
explained later, we do not use this embedding for consistency).}
\item Sphere $\bbS^2$ has a well-known embedding to $\bbR^3$.
\item The square torus has an embedding to $\bbR^4 = \bbR^2 \times \bbR^2$,
obtained by representing the torus as $\bbT^2 = \bbS^1 \times \bbS^1$  and mapping every
circle $\bbS^1$ to $\bbR^2$.
\longonly{\item} For the rectangular torus, we use two circles of sizes corresponding to
the length of edges.
\longonly{\item} For the hexagonal torus, we use three circles, corresponding to the
three axes.
\end{itemize}

\paragraph{Signpost} For closed hyperbolic manifolds we use the following
method. We choose the subset of tiles ${t_1, \ldots, t_d}$ as \emph{signposts}.
Then, for every tile t, we set $m(t) = (\dist(t,t_1), \ldots, \dist(t,t_d))$.
In most cases we choose the signposts to be the tiles of degree other than $6$.
\longonly{
We use other methods in the case of Klein bottle (where we use 13x4 signposts arranged regularly
in the 13x20 tessellation) and Bolza surface (where we also add the vertices of the original
tiles before the Goldberg-Coxeter construction, since the distances from 6 Bolza tiles are
not enough to identify the tile). Figure \ref{embs}ab shows perfect mappings ($E=O$);
signpost tiles are marked with red.}

\paragraph{Landscape} For hyperbolic and Euclidean disks, we use the following method.
We find all the zig-zag lines in the tessellation. These zig-zag lines go along the edges
and split the manifold into two parts. They are obtained by turning alternately to left and
right at every vertex\longonly{ (we assume that all vertices are of valence 3 here)}. 
\longonly{
In Figure \ref{embs}\longonly{c}, we have three zig-zag lines in the $GC(2,1)$ disk, splitting
the disk into 5 regions; as seen, zig-zag lines approximate straight lines. }Let $L$ be the
set of all lines. We assign a random Gaussian vector $v_l$ to each straight line $l\in L$.
The central tile $t_0$ has all coordinates equal to 0. For any other tiles $t$, we find
the set $L_t$ of all the straight lines separating $t_0$ and $t$, and set $m(t) = \sum_{l \in L_t} v_l$.

We call this \emph{landscape method} because it is inspired by the method used to generate
natural-looking landscapes in HyperRogue \cite{hyperrogue}. It represents the reason why
hyperbolic geometry may appear in real-life data such as social network analysis: every line
$l \in L$ represents a part of the social network becoming polarized or specialized\longonly{ and thus
changing their values}.

\subsection{Measuring the quality of final embedding}

\longonly{We are interested in measuring the quality of the final embeddings. 
The following two methods are natural.
{\bf Energy} is given as $\frac{1}{|E_O|} (\sum_{(t,t') \in E_O} \dist(e(t), e(t'))^2-1)$. As we have seen
in our visualization, topologically incorrect edges become stretched, and thus the energies of embeddings
include them are high. The {\bf Kendall coefficient} $k$ measures the correlation of $d_O=\dist(t_1,t_2)$ and
$d_E=\dist(e(t_1), e(t_2))$; every pair of pairs of distances $((d_O,d_E), (d'_O,d'_E))$ contributes 
1 if $d_O > d'_O$ and $d_E > d'_E$ or $d_O < d'_O$ and $d_E < d'_E$, and -1 if $d_O > d'_O$ and $d_E < d'_E$ or
$d_O < d'_O$ and $d_E > d'_E$. We normalize by dividing by the total number of pairs where
$d_O \neq d'_O$. The {\bf Kendall unfitness} is then $100\cdot(1-k)$.}

\longonly{Unfortunately, both of these measures return relatively bad values for embeddings which are actually
correct, such as the embedding from Figure \ref{retrieval}a, where the map
is stretched in one direction (say, horizontal) and compressed in the other direction (say, vertical). The
cases where $(t_1,t_2)$ is a horizontal pair and $(t'_1,t'_2)$ is a horizontal pair worsen the Kendall coefficient. A similar issue happens
for energy.}

\longonly{Therefore we need a topological measure of quality.} 
\shortonly{Our final embeddings should correctly preserve the topology.}
Our primary method of measuring topology preservation is based on the ideas of Villmann et al
\cite{villmann}. This measure is based on the neuron weights $w_t$ for every tile $t \in T_E$.
For every tile $t \in T_E$, let $p_t$ be the point in the manifold $O$ which is the closest
to $w_t$. Define the Voronoi cell $V_t$ as the set of points in the manifold $O$ which are closer to $p_t$
than to any other $p_{t'}$, i.e., $V_t = \{x \in O: \forall t' \in T_E |x-p_t| \leq |x-p_{t'}|\}$. Two Voronoi
cells $V_t$ and $V_{t'}$ are adjacent if $V_t \cap V_{t'} \neq \emptyset$. This way, we obtain two graphs
on the set of tiles $T_E$: the graph $G_E = (T_E, E_E)$ where two tiles are adjacent iff there is an edge in $E$
between them, and the graph $G_V = (T_E, E_V)$ where two tiles $t$, $t'$ are adjacent iff their Voronoi cells
are adjacent.

For an embedding that ideally preserves the topology and also preserves the geometry well enough, we have
$E_V = E_E$. In general, let $d_E(t_1,t_2)$ and $d_V(t_1,t_2)$ be the length of the shortest paths between
tiles $t_1$ and $t_2$ in the graphs $G_V$ and $G_E$. We define the \emph{Villmann measure} of the embedding as
$v = \max_{(t_1,t_2) \in E_E} d_V(t_1,t_2) + \max_{(t_1,t_2) \in E_V} d_E(t_1,t_2)-2$. 
Ideally, we get $v=0$; embeddings which stretch the manifold or have local folds yield small values of $v$ ($2\leq v \leq 4$),
and embeddings which do not preserve the topology yield larger values. An embedding does not preserve the topology
if one of two cases hold: the induced map from $E$ to $O$ is not continuous (making the first component of $v$ large)
or the induced map from $O$ to $E$ is not continuous (making the second component of $v$ large) \cite{villmann}.
\shortonly{See the full version for other methods of measuring embedding quality.}

\longonly{This measures the largest discontinuity. We might also want to measure the number of discontinuities. One natural formula is
$\sum_{(t_1,t_2) \in E_E} (d_V(t_1,t_2)-1)^2 + \sum_{(t_1,t_2) \in E_V} (d_E(t_1,t_2)-1)^2$. Our experiments indicate that such a formula
is very sensitive to local folds, which are in turn very sensitive to the parameters of the SOM algorithm (for sphere-to-sphere mappings,
dispersion scaling power $s=1$ yields significantly smaller values than $s=2$, both for simulated and Gaussian dispersion), making it
difficult to compare various algorithms.}

\longonly{
Another measure of topology preservation is the tears measure, which ignores stretches and local folds. In all the bad cases in Figure \ref{retrieval},
topological errors are exhibited by areas where no tiles from $T_O$ are mapped. However, not all empty
tiles are bad -- some tiles will remain empty simply because $T_O$ is not dense enough. Therefore,
tears($r$) is measured as follows: we count the number of tiles $t \in T_E$ which are empty but
there are tiles $t_1$ and $t_2$ such that $\dist(t,t_1), \dist(t,t_2) \leq r$ and $\dist(t,t_1)+\dist(t_t2) = \dist(t_1,t_2)$.
For $r=1$, this method prevents us from counting empty cells visible in Figure \ref{retrieval}a.
We use $r=1$. This measure is suitable only when both manifolds $E$ and $O$ are closed.}


\newcommand*{\SuperScriptSameStyle}[1]{%
  {\small{$^{#1}$}}%
  }%

\newcommand*{\impc}{\SuperScriptSameStyle{*}}
\newcommand*{\impb}{\SuperScriptSameStyle{**}}
\newcommand*{\imp}{\SuperScriptSameStyle{\dagger}}

\def\closeword{clos}
\def\lcloseword{closed}
\def\diskword{disk}

\longonly{
\begin{table*}[bt]
\resizebox{\linewidth}{!}{
\begin{tabular}{l|c|c|c|c|c|c|c|c|c|c|c}
  & \multicolumn{4}{c|}{energy} & \multicolumn{4}{c|}{Kendall unfitness} & \multicolumn{3}{c}{tears(1)} \\
  $O$ & \closeword & \diskword & \closeword & \diskword & \closeword & \diskword & \closeword & \diskword & \multicolumn{3}{|c}{\closeword} \\
  $E$ & \closeword & \diskword & \diskword & \closeword & \closeword & \diskword & \diskword & \closeword & \multicolumn{3}{|c}{\closeword} \\ \hline
  Effect type  & \multicolumn{8}{c|}{${\Delta E(y)}/{\Delta x_k}$} & $E(y|x)$ & $E(y|x, y>0)$ & $P(y>0|x)$  \\ \hline
(Intercept) &     5.649\imp &     4.613\imp &    21.947\imp &     3.142\imp &    37.975\imp &    22.481\imp &    57.984\imp &    55.725\imp & --- & --- & ---\\
$O$ hyperbolic &     2.534\imp &    -1.735\imp &     3.407\imp &     0.392\imp &     8.579\imp &     2.092\imp &     4.430\imp &    21.516\imp &    21.419\imp &     0.001\imp & 0.003\imp\\
$O$ spherical &    -0.397\imp & --- &    -6.561\imp & --- &    -2.573\imp & --- &    -9.595\imp &     5.652\imp &     5.640\imp & 0.000169\imp & -4.2e-04\imp\\
same\_manifold &    -2.953\imp &    -0.379\imp & --- & --- &   -38.203\imp &     0.681\impb & --- &   -74.210\imp &   -65.403\imp &    -0.488\imp & -0.341\imp\\
same\_geometry &    -3.129\imp &    -0.981\imp &     0.276\imp &    -0.414\imp &    -2.224\imp &   -14.200\imp &     2.080\imp &    -4.608\imp &    -4.594\imp & -0.000192\imp & -3.6e-04\imp\\
both\_orientable &     0.146\imp & --- &    -3.828\imp &     0.444\imp &    -0.637\imp & --- &    -6.813\imp &     3.189\imp &     3.179\imp & 0.000136\imp & 2.3e-04\imp\\
both\_symmetric &    -0.356\imp & --- &     8.238\imp &     0.545\imp &     1.482\imp & --- &    16.318\imp &     0.223 &     0.222 & 8.6e-06 & -1.7e-05\\
diff\_curvature\_pos &    49.145\imp &     0.976\imp &    67.177\imp &     0.394\imp &    89.692\imp &    22.575\imp &    27.560\imp &   124.302\imp &   123.956\imp &     0.005\imp & 0.009\imp\\
diff\_curvature\_neg &    -6.583\imp &     5.365\imp &   -10.691\imp &    -1.340\imp &    39.961\imp &    31.779\imp &     7.286\imp &   422.827\imp &   421.652\imp &     0.016\imp & 0.035\imp\\
diff\_samples &    -0.010\imp & --- &    -0.008\imp &     0.005\imp &    -0.047\imp & --- &    -0.048\imp &     0.198\imp &     0.197\imp & 7.58e-06\imp & 1.4e-05\imp\\
same\_topology &     0.564\imp & --- & --- & --- &     2.010\imp & --- & --- &     3.603\imp &     3.594\imp & 0.000114\imp & -8.3e-04\imp\\
\hline
  N                        & 25600       & 4900        & 11200        & 11200       & 25900        & 4900         & 11200       & 11200       & \multicolumn{3}{c}{25600} \\
  (pseudo)$R^2$            & 0.8496      & 0.7195      & 0.9077       & 0.2471      & 0.7519       & 0.7432       & 0.7736      & 0.6361      & \multicolumn{3}{c}{0.1703}  \\
  $R^2_{adj}$               & 0.8495      & 0.7193      & 0.9076       & 0.2466      & 0.7518       & 0.7429       & 0.7735      & 0.6358      & \multicolumn{3}{c}{---}  \\
\end{tabular}  
}
\caption{Factors affecting quality of SOM embedding. Partial effects for OLS and marginal effects for tobit. \imp, \impb, \impc denote significance at 1\%, 5\%, 10\% level, accordingly.
In all regressions, p-values for joint significance tests equaled 0.000.
}\label{regs}
\end{table*}
}

\longonly{
\begin{table*}[bt]
\resizebox{\linewidth}{!}{
\begin{tabular}{l|c|c|c|c|c|c|c|c|c|c|c}
  & \multicolumn{4}{c|}{energy} & \multicolumn{4}{c|}{Kendall unfitness} & \multicolumn{3}{c}{tears(1)} \\
  $O$ & \closeword & \diskword & \closeword & \diskword & \closeword & \diskword & \closeword & \diskword & \multicolumn{3}{|c}{\closeword} \\
  $E$ & \closeword & \diskword & \diskword & \closeword & \closeword & \diskword & \diskword & \closeword & \multicolumn{3}{|c}{\closeword} \\ \hline
  Effect type  & \multicolumn{8}{c|}{\scalebox{.9}{${\Delta E(y)}/{\Delta x_k}$}} & \scalebox{.9}{$E(y|x)$} & \scalebox{0.9}{$E(y|x,\!y\!>\!0)$} & \scalebox{.9}{$P(y>0|x)$}  \\ \hline
(Intercept) & 5.483\imp &     2.212\imp &    11.598\imp &     1.591\imp & 36.971\imp &    23.578\imp &    59.334\imp &    55.909\imp & --- & --- & ---\\ 
$O$ hyperbolic & 2.908\imp &    -0.931\imp &     1.642\imp &     0.173\imp & 10.171\imp &     2.465\imp &     4.432\imp &    18.884\imp &    18.769\imp &     0.002\imp & 0.003\imp\\ 
$O$ spherical & -0.072 & --- &    -3.395\imp & --- & -2.337\imp & --- &    -9.476\imp &     3.992\imp &     3.979\imp & 0.000205\imp & -4.2e-04\imp\\ 
same\_manifold & -3.257\imp &    -0.166\imp & --- & --- & -34.519\imp &     2.025\imp & --- &   -69.686\imp &   -60.862\imp &    -0.556\imp & -0.341\imp\\ 
same\_geometry & -3.606\imp &    -0.559\imp &     0.116\impb &    -0.208\imp & -2.676\imp &   -14.219\imp &     1.806\imp &    -3.716\imp &    -3.700\imp & -0.000245\imp & -3.6e-04\imp\\ 
both\_orientable & 0.485\imp & --- &    -1.976\imp &     0.188\imp & -0.952\imp & --- &    -6.548\imp &     2.976\imp &     2.963\imp & 0.000204\imp & 2.3e-04\imp\\ 
both\_symmetric & -0.232\imp & --- &     4.330\imp &     0.257\imp & 1.821\imp & --- &    15.724\imp &     0.911\imp &     0.907\imp & 5.74e-05\imp & -1.7e-05\\ 
diff\_curvature\_pos & 51.397\imp &     1.446\imp &   149.335\imp &     0.097\impb & 91.479\imp &   100.401\imp &   101.167\imp &   616.885\imp &   614.368\imp &     0.038\imp & 0.009\imp\\ 
diff\_curvature\_neg & -7.692\imp &    11.848\imp &   -22.978\imp &    -1.158\imp & 49.689\imp &   132.153\imp &    29.844\imp &  1667.528\imp &  1660.723\imp &     0.103\imp & 0.035\imp\\ 
diff\_samples & -0.005\imp & --- &    -0.002\imp & 0.0002\imp & -0.056\imp & --- &    -0.018\imp &     0.040\imp &     0.040\imp & 2.46e-06\imp & 1.4e-05\imp\\ 
same\_topology & 0.885\imp & --- & --- & --- & 1.351\imp & --- & --- &     7.564\imp &     7.543\imp & 0.000316\imp & -8.3e-04\imp\\ 
  N                       & 25600       & 4900       & 11200        & 11200       & 25600        & 4900         & 11200       & 11200       & \multicolumn{3}{c}{25600} \\
  (pseudo)$R^2$           & 0.8513      & 0.7384     & 0.9091       & 0.1522      & 0.7535       & 0.7602       & 0.7964      & 0.6636      & \multicolumn{3}{c}{0.1728}  \\
  $R^2_{adj}$              & 0.8513      & 0.7382      & 0.9090       & 0.1517       & 0.7535       & 0.7600       & 0.7963      & 0.6634      & \multicolumn{3}{c}{---}  \\
\end{tabular}  
}
\caption{Factors affecting quality of SOM embedding (double density of original manifold). Partial effects for OLS and marginal effects for tobit. \imp, \impb, \impc denote significance at 1\%, 5\%, 10\% level, accordingly.
In all regressions, p-values for joint significance tests equaled 0.000.
}\label{regs_d2}
\end{table*}
}

\longonly{
\begin{table*}[bt]
\resizebox{\linewidth}{!}{
\begin{tabular}{l|c|c|c|c|c|c|c|c}
  $O$ & \multicolumn{3}{c|}{\closeword} & \multicolumn{3}{c|}{\diskword} & \closeword & \diskword \\
  $E$ & \multicolumn{3}{c|}{\closeword} & \multicolumn{3}{c|}{\diskword} & \diskword & \closeword \\ \hline
  Effect type  & $E(y|x)$ & $E(y|x, y>0)$ & $P(y>0|x)$  & $E(y|x)$ & $E(y|x, y>0)$ & $P(y>0|x)$ &  \multicolumn{2}{|c}{${\Delta E(y)}/{\Delta x_k}$} \\ \hline
(Intercept) & --- & --- & --- & --- & --- & --- & 29.700\imp & 32.425\imp\\ 
$O$ hyperbolic &     1.789\imp &     1.789\imp & 3.53e-07\imp &    -0.223 &    -0.180 &    -0.024 &     0.470\imp &    -2.568\imp\\
$O$ spherical &     1.942\imp &     1.942\imp & 1.49e-07\imp & --- & --- & --- &     0.807\imp & ---\\
same\_manifold &   -15.153\imp &   -14.697\imp &    -0.076\imp &    -5.900\imp &    -7.235\imp &    -0.653\imp & --- & ---\\
same\_geometry &    -3.088\imp &    -3.088\imp & -1.1e-06\imp &    -6.068\imp &    -4.524\imp &    -0.487\imp &    -3.165\imp &    -2.696\imp\\
both\_orientable &     0.322\imp &     0.322\imp & 4.78e-08\imp & --- & --- & --- &     1.196\imp &    -0.043\\
both\_symmetric &    -1.344\imp &    -1.344\imp & -1.65e-07\imp & --- & --- & --- &    -1.830\imp &     0.354\imp\\
diff\_curvature\_pos &    -9.534\imp &    -9.534\imp & 0 &     4.837\imp &     3.928\imp &     0.532\imp &     3.344\imp &    -5.788\imp\\
diff\_curvature\_neg &   -10.300\imp &   -10.300\imp & 0 &     6.730\imp &     5.466\imp &     0.741\imp &   -10.195\imp &   -12.806\imp\\
diff\_samples &     0.017\imp &     0.017\imp & 0 & --- & --- & --- &     0.025\imp &     0.013\imp\\
same\_topology &    -9.021\imp &    -9.011\imp & -0.000822\imp & --- & --- & --- & --- & ---\\
\hline
N              &\multicolumn{3}{c|}{25600}   & \multicolumn{3}{c|}{4900}   & 11200  & 11200 \\
(pseudo)$R^2$  &\multicolumn{3}{c|}{0.2099}  & \multicolumn{3}{c|}{0.1355} & 0.5882 & 0.5488 \\
$R^2_{adj}$     &\multicolumn{3}{c|}{---}     & \multicolumn{3}{c|}{---}    & 0.5879 & 0.5485 \\
\end{tabular}  
}
\caption{Factors affecting quality of SOM embedding -- Villmann measure. Partial effects for OLS and marginal effects for tobit. \imp, \impb, \impc denote significance at 1\%, 5\%, 10\% level, accordingly. In all regressions, p-values for joint significance tests equaled 0.000.
}\label{regs_vil}
\end{table*}
}

\longonly{
\begin{table*}[bt]
\resizebox{\linewidth}{!}{
\begin{tabular}{l|c|c|c|c|c|c|c|c}
  $O$ & \multicolumn{3}{c|}{\closeword} & \multicolumn{3}{c|}{\diskword} & \closeword & \diskword \\
  $E$ & \multicolumn{3}{c|}{\closeword} & \multicolumn{3}{c|}{\diskword} & \diskword & \closeword \\ \hline
  Effect type  & $E(y|x)$ & $E(y|x, y>0)$ & $P(y>0|x)$  & $E(y|x)$ & $E(y|x, y>0)$ & $P(y>0|x)$ &  \multicolumn{2}{|c}{${\Delta E(y)}/{\Delta x_k}$} \\ \hline
(Intercept) & --- & --- & --- & --- & --- & --- &    29.789\imp &    33.035\imp\\ 
$O$ hyperbolic &     1.806\imp &     1.806\imp & 3.52e-07\imp &    -0.025 &    -0.018 &    -0.002 &     0.701\imp &    -2.591\imp\\ 
$O$ spherical &     2.078\imp &     2.078\imp & 1.55e-07\imp & --- & --- & --- &     0.717\imp & ---\\ 
same\_manifold &   -15.228\imp &   -14.771\imp &    -0.075\imp &    -5.163\imp &    -4.719\imp &    -0.561\imp & --- & ---\\ 
same\_geometry &    -3.078\imp &    -3.078\imp & -1.07e-06\imp &    -8.714\imp &    -6.325\imp &    -0.493\imp &    -3.120\imp &    -2.788\imp\\ 
both\_orientable &     0.386\imp &     0.386\imp & 5.76e-08\imp & --- & --- & --- &     0.965\imp &    -0.318\imp\\ 
both\_symmetric &    -1.398\imp &    -1.398\imp & -1.69e-07\imp & --- & --- & --- &    -1.473\imp &     0.283\imp\\ 
diff\_curvature\_pos &   -37.127\imp &   -37.127\imp & 0 &    27.320\imp &    19.796\imp &     2.243\imp &    21.645\imp &   -24.700\imp\\ 
diff\_curvature\_neg &   -47.384\imp &   -47.384\imp & 0 &    42.883\imp &    31.073\imp &     3.521\imp &   -41.980\imp &   -53.582\imp\\ 
diff\_samples &     0.004\imp &     0.004\imp & 0 & --- & --- & --- &     0.012\imp &    -0.002\imp\\ 
same\_topology &    -9.053\imp &    -9.043\imp & -0.00081\imp & --- & --- & --- & --- & ---\\ 
\hline
N              &\multicolumn{3}{c|}{25600}   & \multicolumn{3}{c|}{4900}   & 11200  & 11200 \\
(pseudo)$R^2$  &\multicolumn{3}{c|}{0.2092}  & \multicolumn{3}{c|}{0.1402} & 0.6078 & 0.5598 \\
$R^2_{adj}$     &\multicolumn{3}{c|}{---}     & \multicolumn{3}{c|}{---}    & 0.6075 & 0.5595 \\
\end{tabular}  
}
\caption{Factors affecting quality of SOM embedding -- Villmann measure. Partial effects for OLS and marginal effects for tobit. \imp, \impb, \impc denote significance at 1\%, 5\%, 10\% level, accordingly. In all regressions, p-values for joint significance tests equaled 0.000.
}\label{regs_vil_d2}
\end{table*}
}

\longonly{
\begin{table*}[bt]
\resizebox{\linewidth}{!}{
\begin{tabular}{l|c|c|c|c|c|c|c|c}
  $O$ & \multicolumn{3}{c|}{\lcloseword} & \multicolumn{3}{c|}{\diskword} & \lcloseword & \diskword \\
  $E$ & \multicolumn{3}{c|}{\lcloseword} & \multicolumn{3}{c|}{\diskword} & \diskword & \lcloseword \\ \hline
  Effect type  & $E(y|x)$ & $E(y|x, y>0)$ & $P(y>0|x)$  & $E(y|x)$ & $E(y|x, y>0)$ & $P(y>0|x)$ &  \multicolumn{2}{|c}{${\Delta E(y)}/{\Delta x_k}$} \\ \hline
(Intercept) & --- & --- & --- & --- & --- & --- &    29.79\imp &    33.04\imp\\ 
$O$ hyperbolic &     1.81\imp &     1.81\imp & 3.5e-07\imp & -0.025 & -0.018 & -0.0021 &     0.70\imp &    -2.59\imp\\ 
$O$ spherical &     2.08\imp &     2.08\imp & 1.5e-07\imp & --- & --- & --- &     0.72\imp & ---\\ 
same\_manifold &   -15.23\imp &   -14.77\imp &    -0.08\imp &    -5.16\imp &    -4.72\imp &    -0.56\imp & --- & ---\\ 
same\_geometry &    -3.08\imp &    -3.08\imp & -1.1e-06\imp &    -8.71\imp &    -6.32\imp &    -0.49\imp &    -3.12\imp &    -2.79\imp\\ 
both\_orientable &     0.39\imp &     0.39\imp & 5.8e-08\imp & --- & --- & --- &     0.96\imp &    -0.32\imp\\ 
both\_symmetric &    -1.40\imp &    -1.40\imp & -1.7e-07\imp & --- & --- & --- &    -1.47\imp &     0.28\imp\\ 
diff\_curvature\_pos &   -37.13\imp &   -37.13\imp & 0 &    27.32\imp &    19.80\imp &     2.24\imp &    21.64\imp &   -24.70\imp\\ 
diff\_curvature\_neg &   -47.38\imp &   -47.38\imp & 0 &    42.88\imp &    31.07\imp &     3.52\imp &   -41.98\imp &   -53.58\imp\\ 
diff\_samples &    0.004\imp &    0.004\imp & 0 & --- & --- & --- &    0.012\imp & -0.0018\imp\\ 
same\_topology &    -9.05\imp &    -9.04\imp & -0.00081\imp & --- & --- & --- & --- & ---\\ 
\hline
N              &\multicolumn{3}{c|}{25600}   & \multicolumn{3}{c|}{4900}   & 11200  & 11200 \\
(pseudo)$R^2$  &\multicolumn{3}{c|}{0.2092}  & \multicolumn{3}{c|}{0.1402} & 0.6078 & 0.5598 \\
$R^2_{adj}$     &\multicolumn{3}{c|}{---}     & \multicolumn{3}{c|}{---}    & 0.6075 & 0.5595 \\
\end{tabular}  
}
\caption{Factors affecting quality of SOM embedding -- Villmann measure. Partial effects for OLS and marginal effects for tobit. \imp, \impb, \impc denote significance at 1\%, 5\%, 10\% level, accordingly. In all regressions, p-values for joint significance tests equaled 0.000.
}\label{regs_vil_d2}
\end{table*}
}

\shortonly{
\begin{table*}[bt]
\begin{tabular}{l|c|c|c|c|c|c|c|c}
\toprule
  $O$ & \multicolumn{3}{c|}{\lcloseword} & \multicolumn{3}{c|}{\diskword} & \lcloseword & \diskword \\
  $E$ & \multicolumn{3}{c|}{\lcloseword} & \multicolumn{3}{c|}{\diskword} & \diskword & \lcloseword \\ \hline
  Effect type  & $E(y|x)$ & $E(y|x, y>0)$ & $P(y>0|x)$  & $E(y|x)$ & $E(y|x, y>0)$ & $P(y>0|x)$ &  \multicolumn{2}{|c}{${\Delta E(y)}/{\Delta x_k}$} \\ \midrule
(Intercept) & --- & --- & --- & --- & --- & --- & 29.70\imp & 32.42\imp\\ 
$O$ hyperbolic &     1.79\imp &     1.79\imp & 3.5e-07\imp &    -0.22 &    -0.18 & -0.024 &     0.47\imp &    -2.57\imp\\
$O$ spherical &     1.94\imp &     1.94\imp & 1.5e-07\imp & --- & --- & --- &     0.81\imp & ---\\
same\_manifold &   -15.15\imp &   -14.70\imp &    -0.08\imp &    -5.90\imp &    -7.23\imp &    -0.65\imp & --- & ---\\
same\_geometry &    -3.09\imp &    -3.09\imp & -1.1e-06\imp &    -6.07\imp &    -4.52\imp &    -0.49\imp &    -3.16\imp &    -2.70\imp\\
both\_orientable &     0.32\imp &     0.32\imp & 4.8e-08\imp & --- & --- & --- &     1.20\imp & -0.043\\
both\_symmetric &    -1.34\imp &    -1.34\imp & -1.7e-07\imp & --- & --- & --- &    -1.83\imp &     0.35\imp\\
diff\_curv\_pos &    -9.53\imp &    -9.53\imp & 0 &     4.84\imp &     3.93\imp &     0.53\imp &     3.34\imp &    -5.79\imp\\
diff\_curv\_neg &   -10.30\imp &   -10.30\imp & 0 &     6.73\imp &     5.47\imp &     0.74\imp &   -10.20\imp &   -12.81\imp\\
diff\_samples &    0.017\imp &    0.017\imp & 0 & --- & --- & --- &    0.025\imp &    0.013\imp\\
same\_topology &    -9.02\imp &    -9.01\imp & -0.00082\imp & --- & --- & --- & --- & ---\\
\hline
N              &\multicolumn{3}{c|}{25600}   & \multicolumn{3}{c|}{4900}   & 11200  & 11200 \\
(pseudo)$R^2$  &\multicolumn{3}{c|}{0.2099}  & \multicolumn{3}{c|}{0.1355} & 0.5882 & 0.5488 \\
$R^2_{adj}$     &\multicolumn{3}{c|}{---}     & \multicolumn{3}{c|}{---}    & 0.5879 & 0.5485 \\ \bottomrule
\end{tabular}  
\caption{Factors affecting quality of SOM embedding -- Villmann measure. Partial effects for OLS and marginal effects for tobit. \imp, \impb, \impc denote significance at 1\%, 5\%, 10\% level, accordingly. In all regressions, p-values for joint significance tests equaled 0.000.
}\label{regs_vil}
\end{table*}
}

\subsection{Quantitative Results}
We use the following parameters: $t_{max}=30000$ iterations, learning coefficient $\eta=0.1$,
dispersion precision $p=10^{-4}$, $T$ is the number of dispersion steps until the max value/min value $\leq$ 1.6, 60 landscape dimensions,
manifolds with about 520 tiles.
Computing 57600 embeddings takes 4 hours on 8-core Intel(R) Core(TM) i7-9700K CPU @ 3.60GHz.
Our implementation is included in the RogueViz non-Euclidean geometry engine\longonly{ \cite{rogueviz2021}}.
The results of our experiments, code and visualizations are at 
\url{https://figshare.com/articles/software/Non-Euclidean_Self-Organizing_Maps_code_and_data_/16624393}.

\paragraph{Comparison of simulated and Gaussian dispersion}

We use the aforementioned measures of quality to check if simulated dispersion improves
the quality of the embedding in comparison to Gaussian. In the Gaussian dispersion function, 
we take the discrete distance between tiles as the distance between two neurons.
We take advantage of the paired nature of the data
and compute the differences between the values of the quality measure obtained with Gaussian and the simulated
dispersion. We use Wilcoxon test to check if the difference is statistically insiginificant,
against the alternative hypothesis that the results from simulated dispersion are better. We use 1\% significance level.

\longonly{
\begin{table}[h]
\scalebox{0.9}{
\begin{tabular}{l|c|c|c|c|c|c|c}
          & \multirow{3}{*}{all} & \multicolumn{2}{|c|}{$E=O$} & \multicolumn{4}{c}{$E\neq O$} \\ 
  $O$     &         & \closeword & \diskword             & \closeword & \diskword & \closeword & \diskword \\
  $E$     &         & \closeword & \diskword             & \closeword & \diskword & \diskword & \closeword \\ \hline
  energy  & 0.00   & 0.00  & 0.02$^\ddagger$             & 0.00       & 0.00      & 0.00          & 0.92$^\ddagger$ \\ \hline
  K. unfit. & 0.00   & 0.00  & 0.97$^\ddagger$             & 0.00       & 0.00      & 1.00          & 0.00          \\ \hline
  tears(1) & 1.00   & 0.00  & ---                        & 1.00       & ---       & ---           & ---          \\ \hline
  Villmann & 0.00   & 0.00  & 0.00                       & 0.00       & 0.00      & 0.00          & 0.00   \\ 
\end{tabular}
}
\caption{
P-values for Wilcoxon tests on differences between quality measures from
SOMs with Gaussian against simulated dispersion. $H_1$ indicates better results from simulated dispersion. $\ddagger$ denotes statistically insignificant difference.}\label{wilcoxon}
\end{table}
}

\longonly{
\begin{table}[h]
\scalebox{0.9}{
\begin{tabular}{l|c|c|c|c|c|c|c}
          & \multirow{3}{*}{all} & \multicolumn{2}{|c|}{$E=O$} & \multicolumn{4}{c}{$E\neq O$} \\ 
  $O$     &         & \closeword & \diskword             & \closeword & \diskword & \closeword & \diskword \\
  $E$     &         & \closeword & \diskword             & \closeword & \diskword & \diskword & \closeword \\ \hline
  energy  & 0.00   & 0.91$^\ddagger$  & 0.77$^\ddagger$  & 1.00       & 0.00      & 0.00          & 1.00 \\ \hline
  K. unfit. & 0.00   & 0.00  & 0.99$^\ddagger$ & 0.00       & 0.00      & 1.00          & 0.00          \\ \hline
  tears(1)   & 1.00  & 0.00  & ---          & 1.00       & ---        & ---            & ---          \\ \hline
  Villmann  & 0.00   & 0.00  & 0.00        &  0.00  & 0.00      & 0.00     & 0.00 \\
\end{tabular}
}
\caption{
P-values for Wilcoxon tests on differences between quality measures from
SOMs with Gaussian against simulated dispersion (double density of original manifold). $H_1$ indicates better results from simulated dispersion. $\ddagger$ denotes statistically insignificant difference.}\label{wilcoxon_d2}
\end{table}
}

\longonly{
Our results (Table~\ref{wilcoxon}) show that the embeddings obtained with simulated dispersion have usually lower energy than the
embeddings obtained with Gaussian dispersion. Two expections are the scenario when we embed disk data on a closed manifold and the scenario when we embed correctly disks.
However, those results are statistically significant (p-value for two-sided Wilcoxon test 0.153 for the first case and 0.044 for the second one). Embeddings obtained with
simulated dispersion also yield results better preserving the original relationships in data with respect to Kendall coefficient.
Two exceptions are: the insignificant difference for correctly used discs (p-value for two sided Wilcoxon test 0.061) and a significant difference
in favor of Gaussian in the case of wrongly embedded closed manifold to disk data. For completeness, we also show the results for topological errors.
Here, we find some evidence of advantage of Gaussian if we embed to wrong manifold. However, embedding into a
wrong manifold naturally comes with creating tears in embeddings. 
In the case of correctly embedded closed manifolds, simulated dispersion yields lower number of tears. 
Our results are stable even for double density of the original manifold (Table~ \ref{wilcoxon_d2}).}

\shortonly{Our results show that the embeddings obtained with simulated dispersion have usually lower Villmann's measure than the embeddings obtained with Gaussian dispersion. 
This remains true whether we use single or double density of the original manifold, or when we restrict only to closed/disk $O$, or closed/disk $E$. The p-value is 0.00
in all cases. The results are slightly different for other measures of embedding quality; see the full version for more details.}

\longonly{
\begin{figure*}
\centering
\subfig{0.24\linewidth}{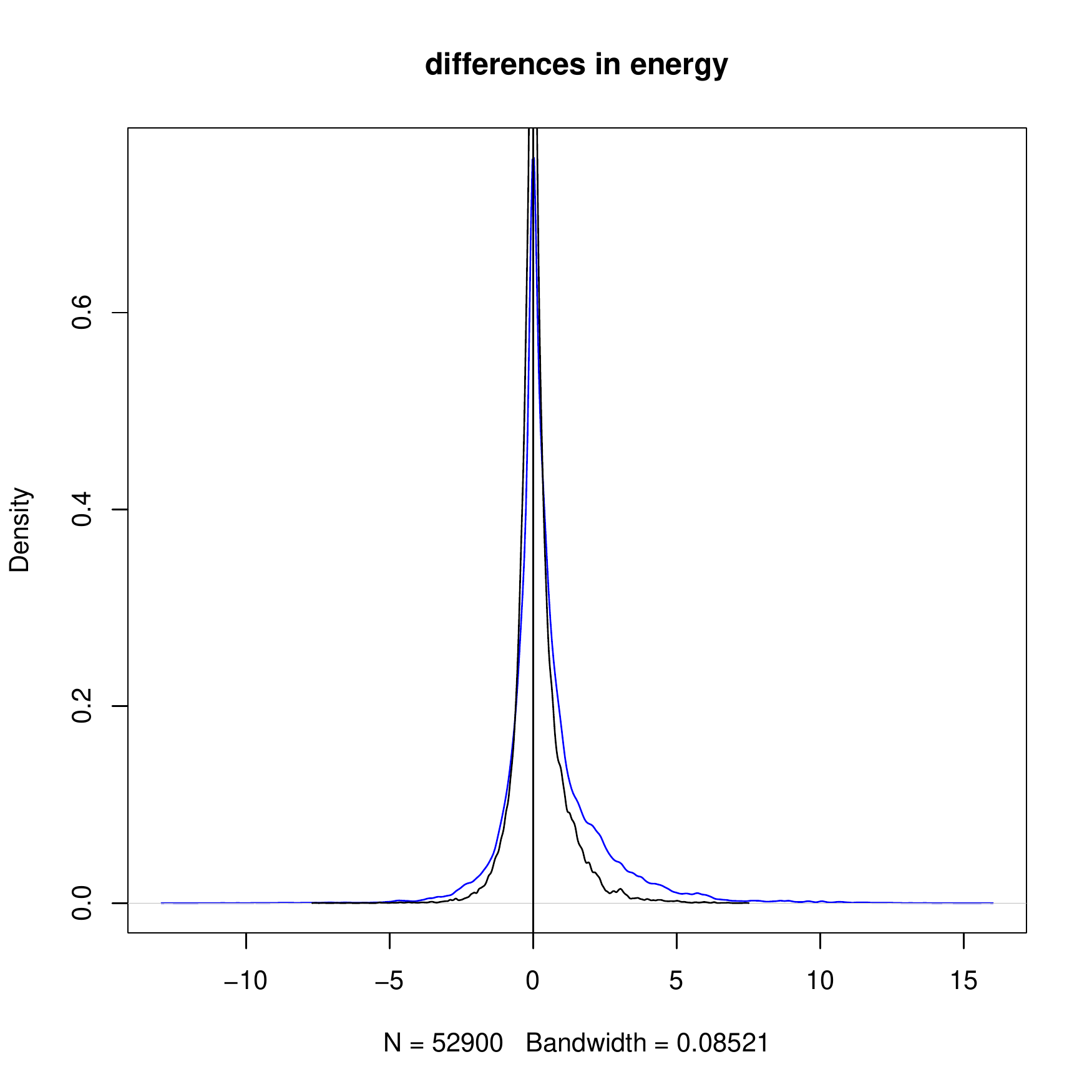} \hskip -1mm
\subfig{0.24\linewidth}{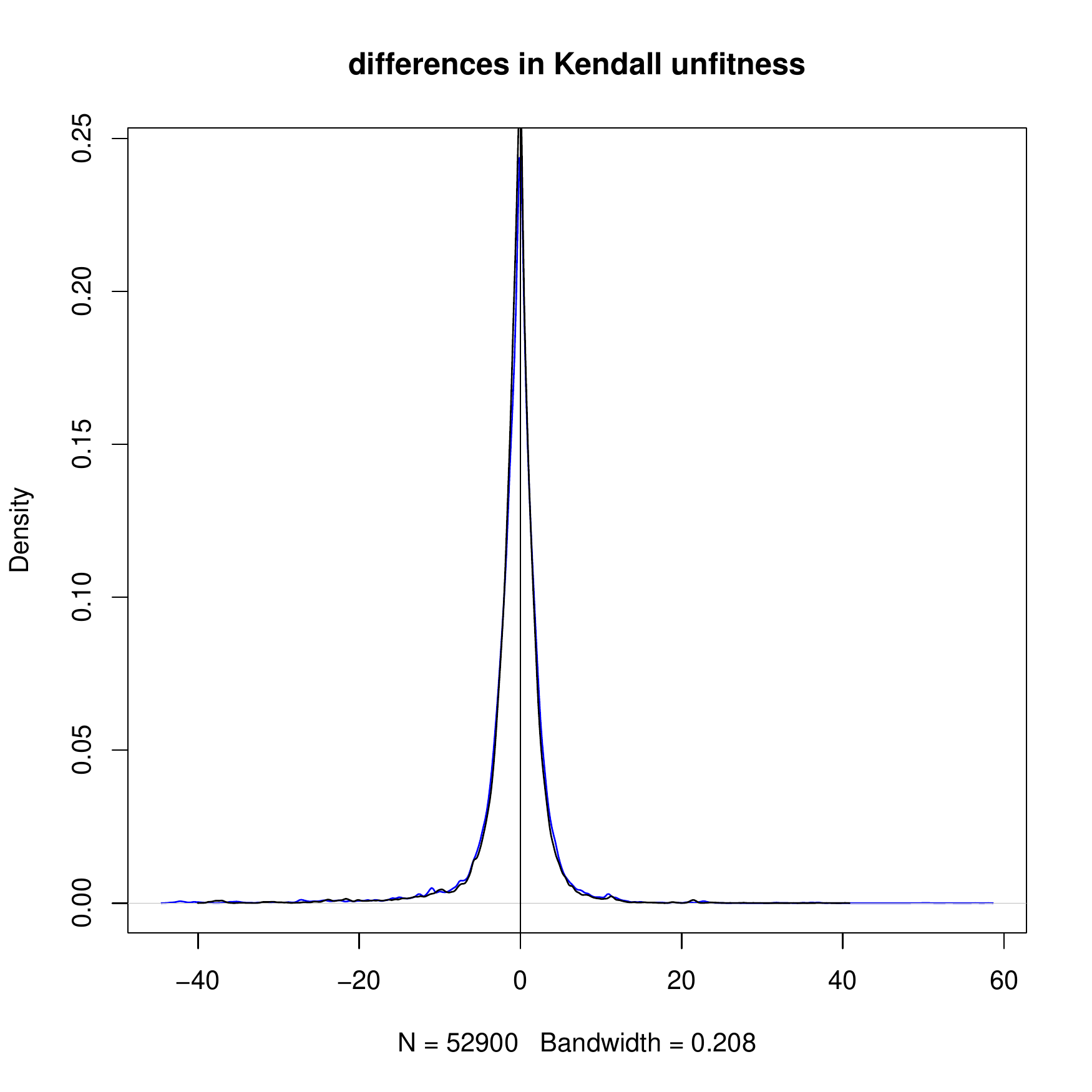} \hskip -1mm
\subfig{0.24\linewidth}{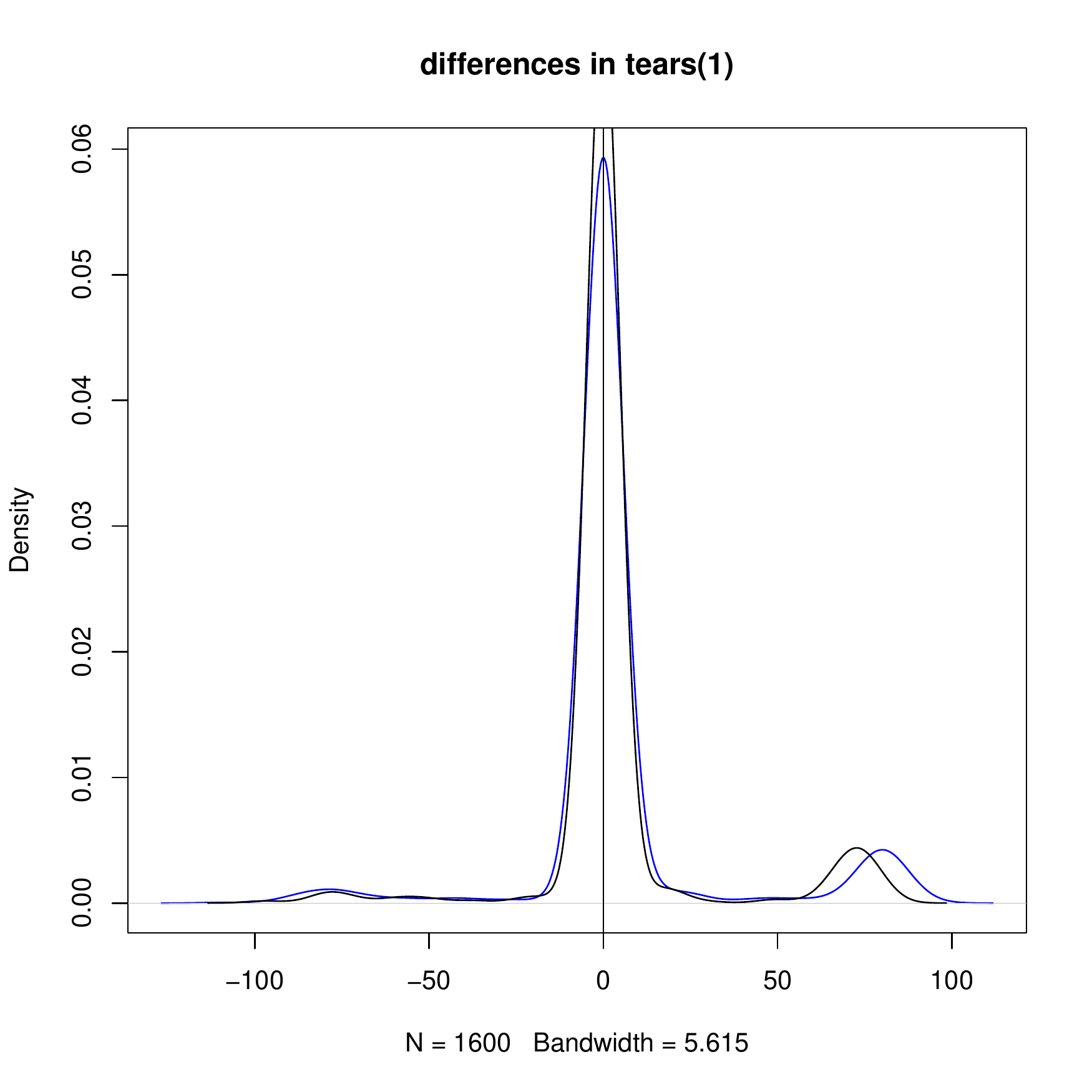}
\subfig{0.24\linewidth}{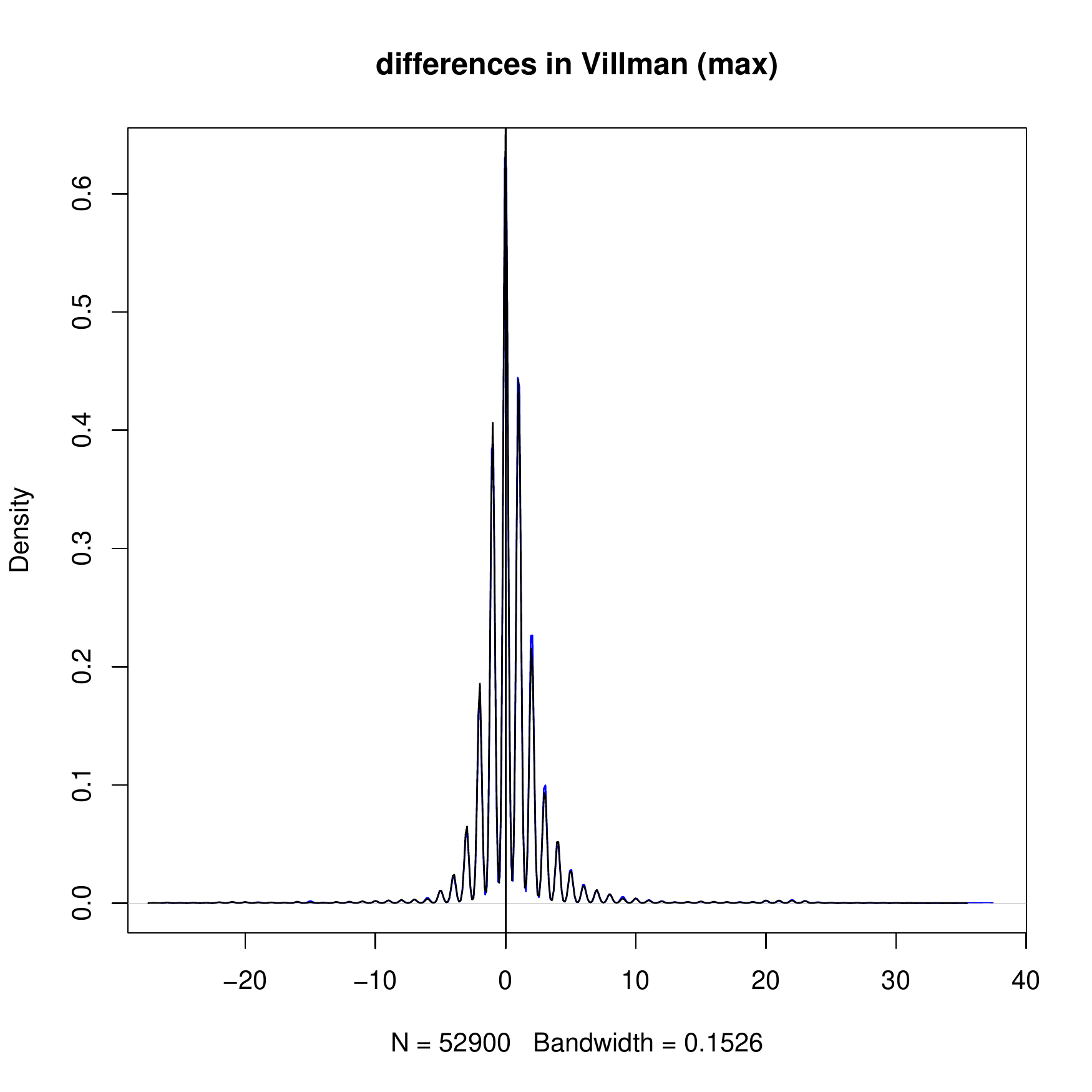}
\caption{\label{fig:diffd2}
Distribution of differences (Gaussian$-$simulated) for (a) energy, (b) Kendall unfitness, (c) tears(1) for the same closed manifold, (d) Villmann measure. Black graph is double density, blue graph is single density.}
\end{figure*}
See Figure \ref{fig:diffd2} for the Distribution of differences (Gaussian$-$simulated) for our mesasures.}

\paragraph{Quality of shape recovery}

We have already shown that using simulated dispersion improves the quality of the embedding. However, SOM on even correctly
chosen manifold may be prone to errors. Here, we will analyze the factors that affect the errors in embeddings. \longonly{To this end, for errors in energy and Kendall
unfitness we will use OLS regression.}

\longonly{According to data in Table~\ref{regs}, if original data comes from hyperbolic geometry, SOMs err more in comparison to original data from Euclidean geometry.
On the contrary, SOMs err less on data from spherical geometry than from Euclidean geometry in terms of energy and Kendall coefficient. Correct choice of
embedding manifold reduces errors (with exception for Kendall coefficient for correctly chosen disks). Same geometry has ambiguous effect. With correctly chosen
embedding manifolds, SOMs err less, otherwise, same geometry may not help. Wrong choices of curvature usually come with greater numbers of errors: using embedding manifold with lower curvature than the original
(positive difference in curvature, {\it diff\_curv\_pos})
always worsens the fit; higher curvature than the original usually improves the energy.
Surprisingly, same topology worsens Kendall coefficient and increases energy errors. However, these measures are not well fit for this purpose.}

\longonly{Tears in visualizations of data embeddings from SOMs should indicate boundaries of clusters in data.
In most cases, the users of SOM do not know the real shape of data. In our setup, we did not
create clusters, so tears are errors that could mislead the user of SOM. Therefore, understanding what factors affect
the topological errors is crucial for the users. However, OLS fails to account for the qualitative difference between zero (lack of errors) observations and non-zero observations when we analyze tears.
Therefore for topological errors, we will use censored regression model (tobit model) \cite{tobit,greene}.}

\longonly{Out of 25600 embeddings on closed manifolds we got 2081 embeddings without topological errors (8.1\%). For correctly chosen closed manifolds this percentage was
significantly higher (94.3\%). The probability of SOMs making topological errors ($P(y\!>\!0|x)$) decreases if we correctly choose manifold, topology or at least geometry.
Differences in curvature vastly increase both the probability of SOM yielding tears and the number of those errors (for whole sample $E(y|x)$ and conditionally
if SOM erred $(E(y|x,\!y{>}0)$). Again, original hyperbolic manifolds are harder to recover in comparison to
Euclidean manifolds; also, orientability increases the difficulty of the task. Symmetry is insignificant when it comes to topological errors. The results are stable for the double density of the original manifold \longonly{(Table~\ref{regs_d2})}\shortonly{(full version)}\vshortonly{\cite{somarxiv}}.
}

OLS regression fails to account for the qualitative difference between low values of Villmann measure (only local geometric errors) and high values (discontinuities). 
Therefore for topological errors where correct embeddings are possible, we will use censored regression model (tobit model) \cite{tobit,greene}. To this end, we left censor values of Villmann measures lower than 8 to 0.

In the case of closed manifolds embedded on close manifolds, out of 25600 embeddings we got 2082 embeddings without topological errors (8.1\%). If disks were embedded to disks, out of 4900 embeddings 2499 were free of topological errors (51\%). For correctly chosen manifolds this percentage was
significantly higher (94.3\% for closed ones and 100\% for disks). The probability of SOMs making topological errors ($P(y\!>\!0|x)$) decreases if we correctly choose manifold, topology or at least geometry. Differences in curvature vastly increase both the probability of SOM yielding tears and the number of those errors (for whole sample $E(y|x)$ and conditionally
if SOM erred $(E(y|x,\!y{>}0)$) in the case of disks. Again, original hyperbolic closed manifolds are harder to recover in comparison to
Euclidean manifolds; also, orientability increases the difficulty of the task. 
The results are stable for the double density of the original manifold\longonly{ (Table~\ref{regs_d2})}\shortonly{ (full version)}\vshortonly{(\cite{somarxiv})}.

\section{Discussion}

\paragraph{Choosing the manifold.} One of the major concerns regarding using non-Euclidean SOMs is the choice of the underlying manifold. Depending on what is the core interest of the researcher, the choice of the underlying manifold may vary. It is typical for multidimensional analysis techniques that the eventual choice of the setup can be subjective. To make sure the results are robust, one may conduct the simulations
on distinct spaces. \longonly{Our proposition allows for easy comparison of the results. }The Goldberg-Coxeter construction lets us use
similar number of neurons for different manifolds, controling for the number of possible groups. Later diagnostics may include comparison of information criteria.

\longonly{
\paragraph{Distances in Gaussian} 
The Gaussian dispersion \cite{ritter99} was based on geometric distance, while in our benchmark we take the discrete distance between tiles.
\longonly{Table~\ref{wilcoxon_gauss} contains p-values for Wilcoxon tests with alternative hypotheses that SOMs with Gaussian
dispersion based on geometric distances perform worse than those with Gaussian dispersion based on discrete distances.
Geometric distance between $a$ and $b$ is the length of the (shortest) geodesic from $a$ to $b$ on sphere, Euclidean plane or hyperbolic plane.
Original proposition \cite{ritter99} did not take into account quotient spaces.
In the case of quotient spaces, this notion is less natural, since there may be multiple geodesics from $a$ to $b$; therefore, 
we performed our comparison only for disks and spheres.
The results obtained with discrete distances were significantly better than the results obtained with geometric ones in terms of all metrics but Villmann's measure. According to Villmann's measure, geometric distances were a better fit when the embedding manifold is not the same as the original one.}
}

\longonly{
\begin{table}[h]
\centering
\begin{tabular}{l|c|c|c}
            & all      & $E=O$ & $E\neq O$ \\ \hline
  energy    & 0.00     & 0.00  & 0.00 \\ \hline
  K. unfit. & 0.00     & 0.00  & 0.00 \\ \hline
  Villmann  & 1.00     & 0.00  & 1.00 \\
\end{tabular}
\caption{
P-values for Wilcoxon tests on differences between SOMs with Gaussian dispersion based on geometric distance and SOMs with Gaussian dispersion based on geometric distance. $H_1$ indicates better results from discrete distances.}\label{wilcoxon_gauss}
\end{table}
}

\longonly{
\paragraph{Landscape dimension}
\begin{figure}
\centering
\subfig{0.49\linewidth}{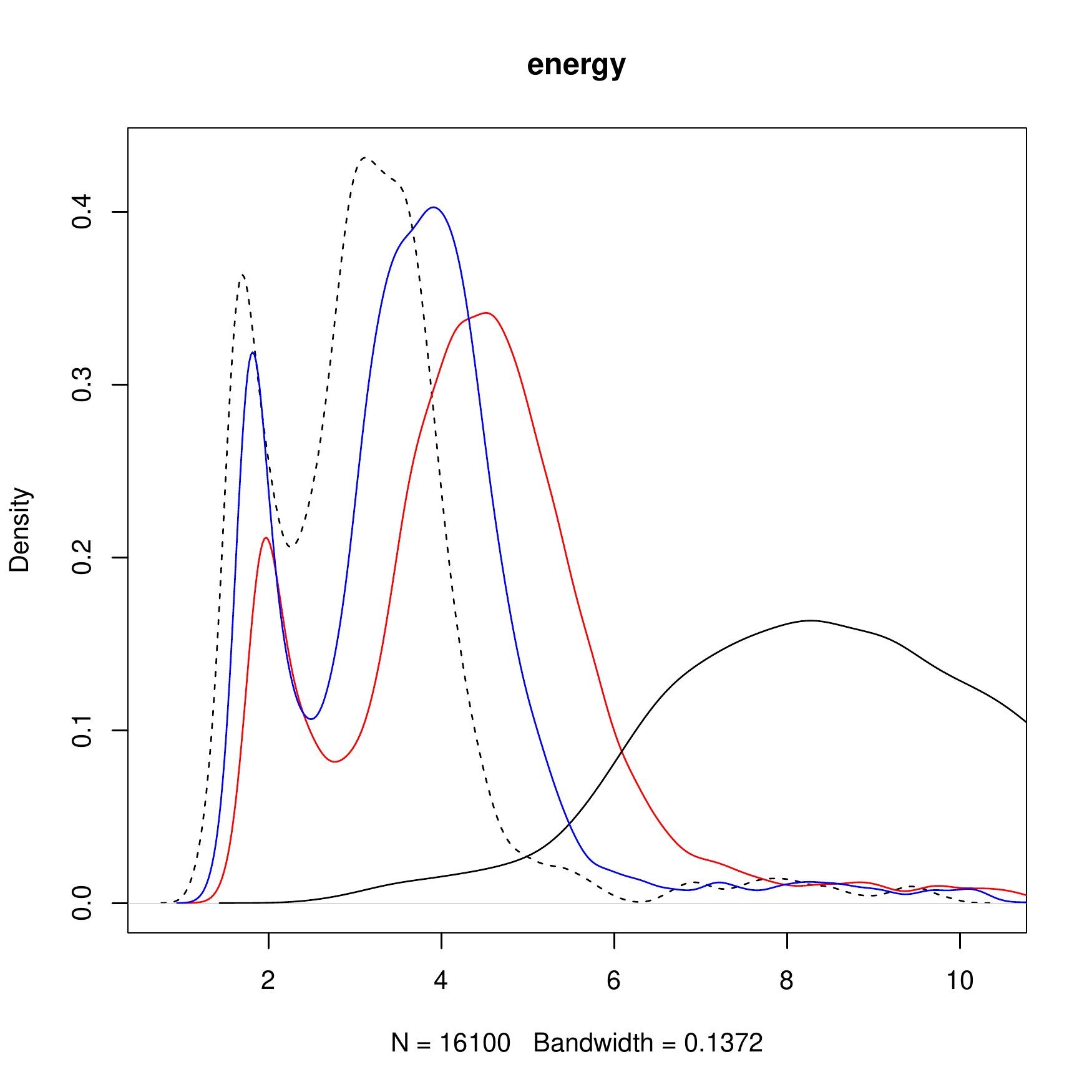} \hskip -1mm
\subfig{0.49\linewidth}{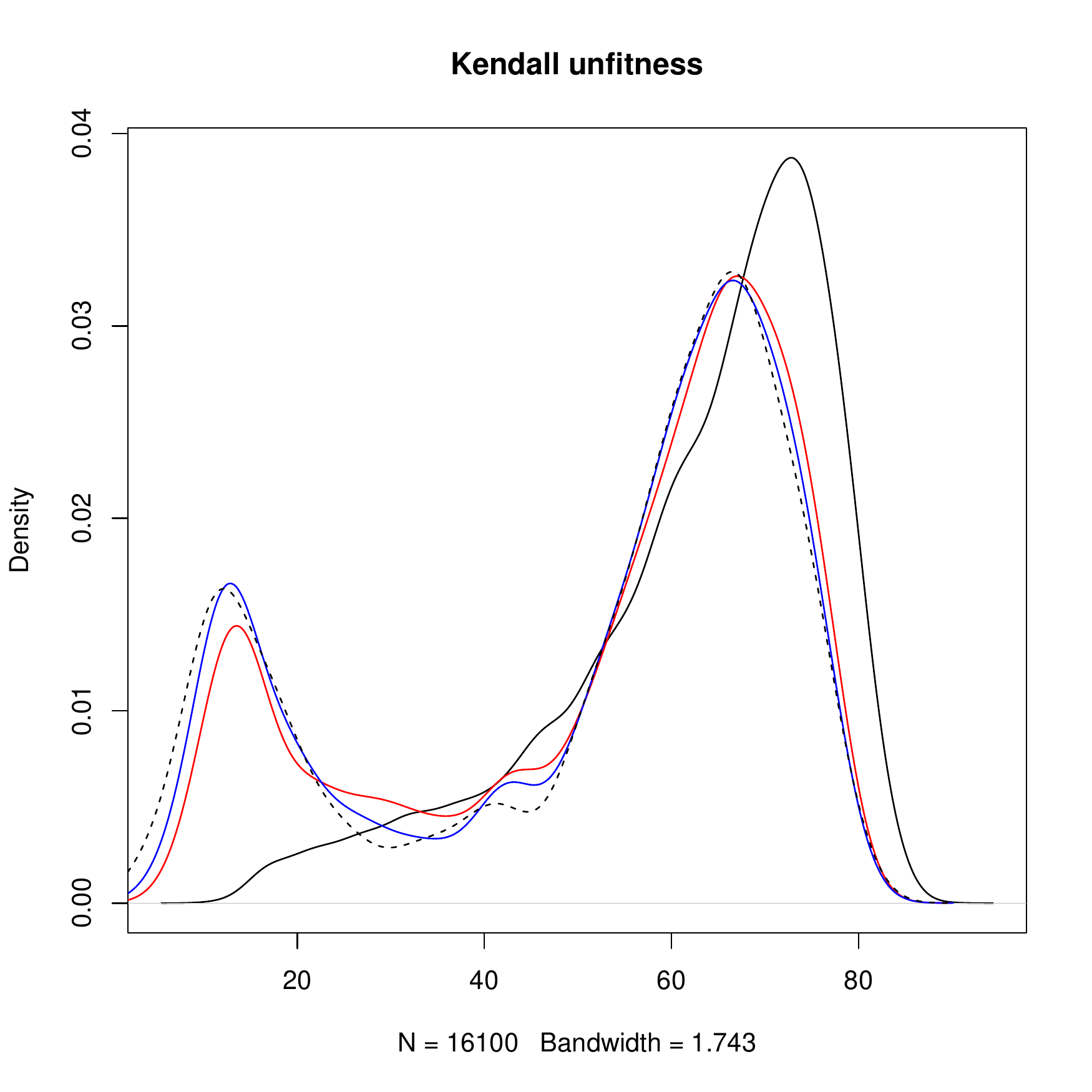}
\subfig{0.49\linewidth}{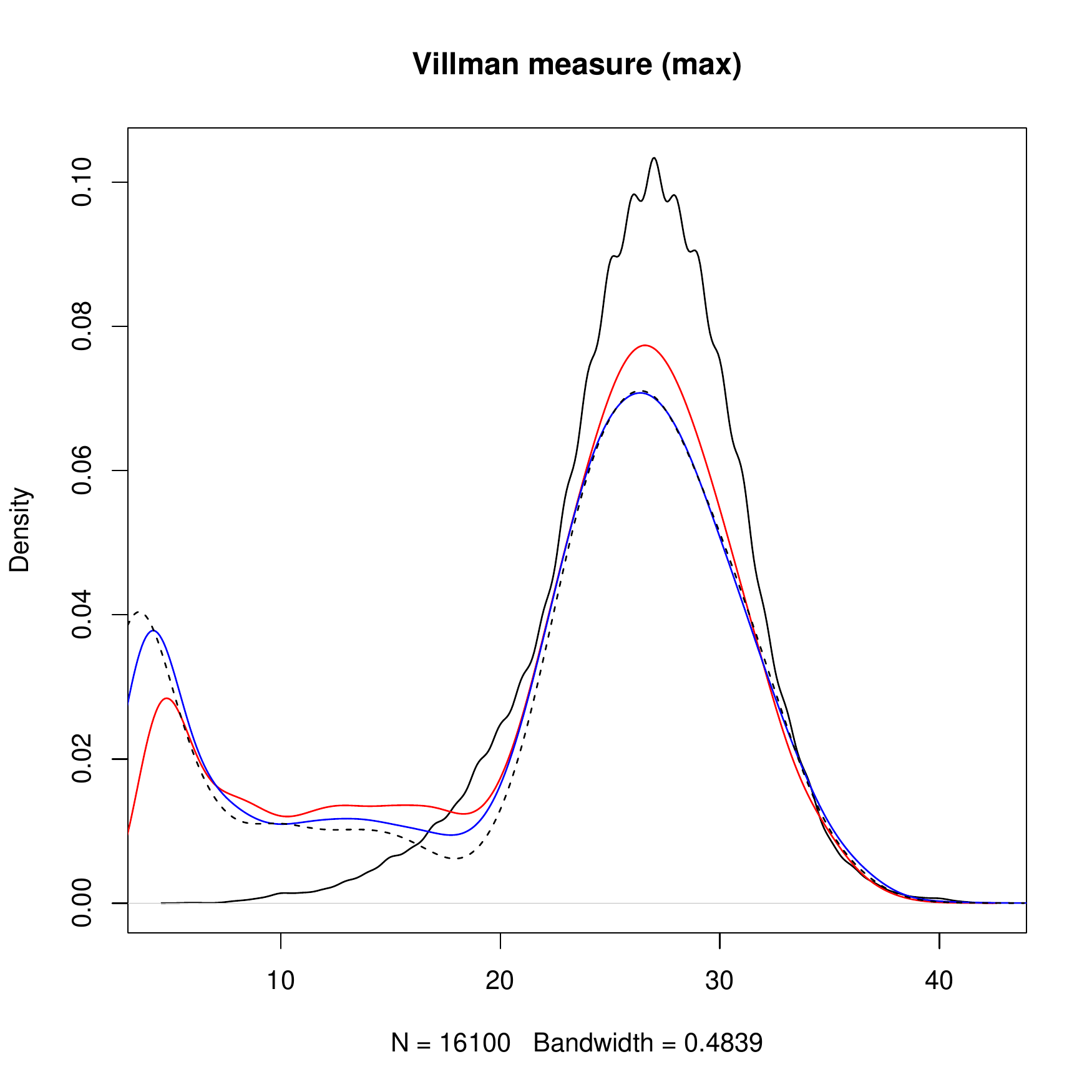}
\subfig{0.49\linewidth}{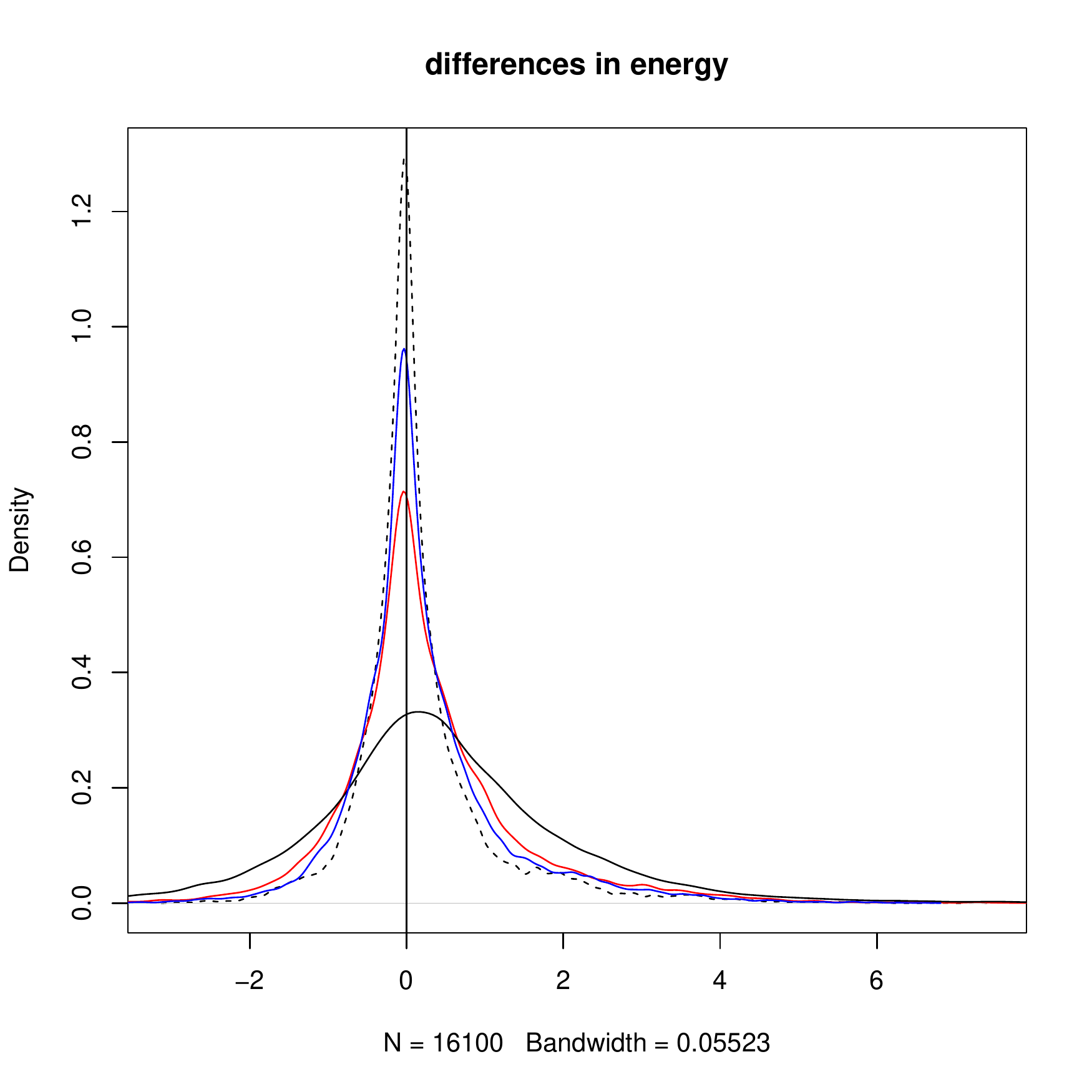} \hskip -1mm
\subfig{0.49\linewidth}{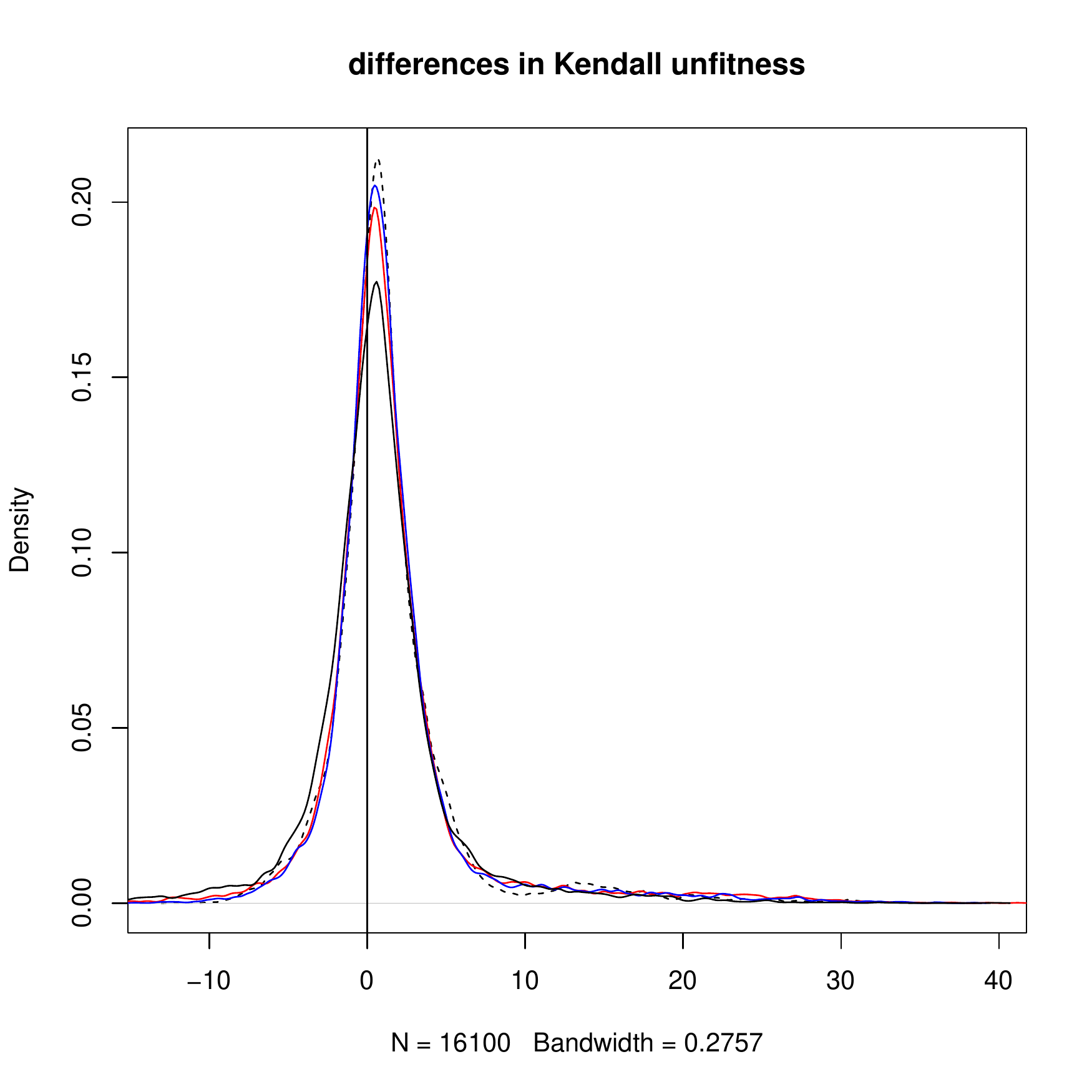}
\subfig{0.49\linewidth}{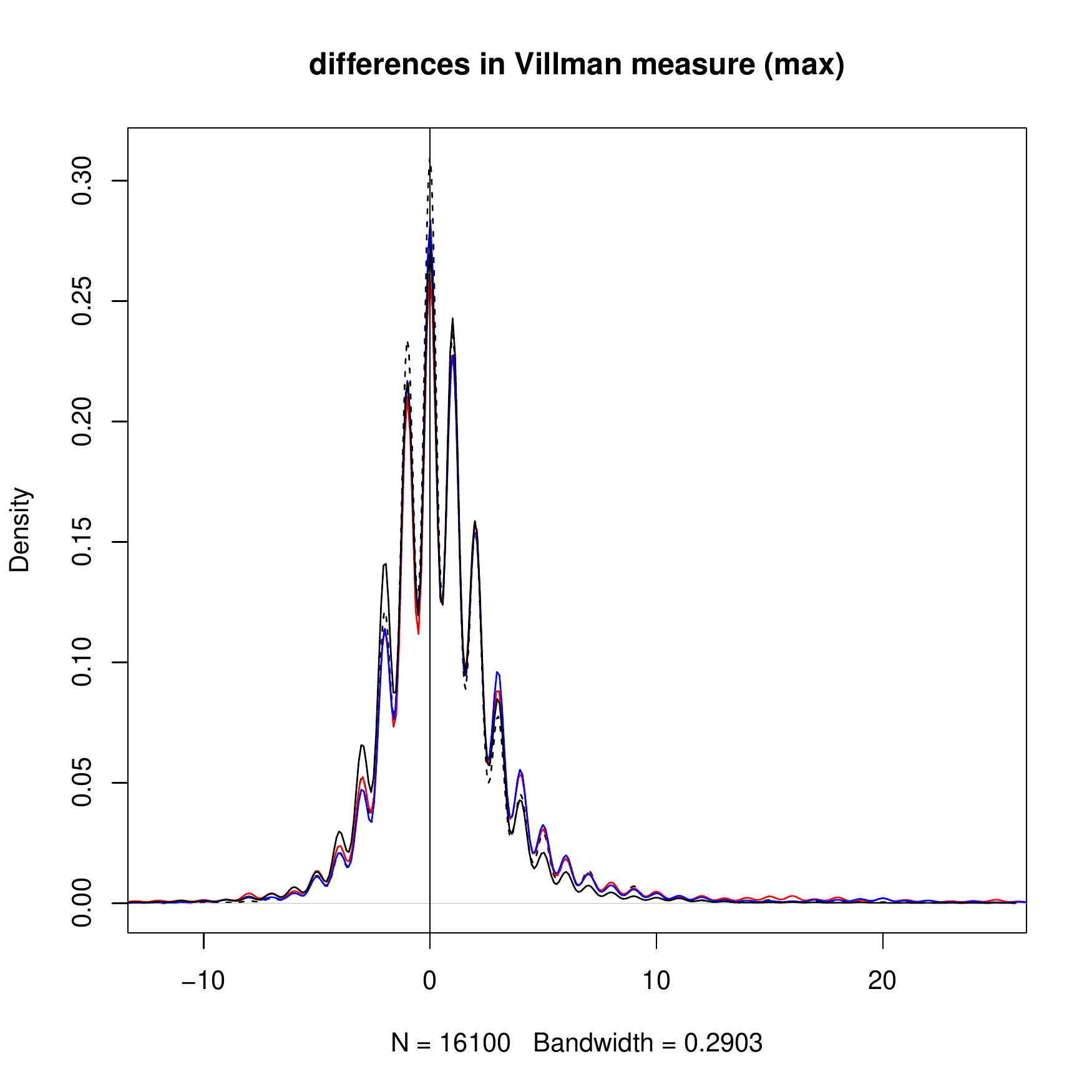}
\caption{\label{fig:ldim}
Changing the landscape dimension. (black) $d=10$, (red) $d=30$, (blue) $d=60$, (dashed) deterministic.
(a)~absolute energy (simulated dispersion);
(b)~absolute Kendall unfitness (simulated dispersion);
(c)~absolute Villmann measure (simulated dispersion);
(d)~difference in energy (Gaussian$-$simulated);
(e)~difference in Kendall unfitness (Gaussian$-$simulated).}
(f)~difference in Villmann measure (Gaussian$-$simulated);
\end{figure}
\begin{figure}
\centering
\subfig{0.49\linewidth}{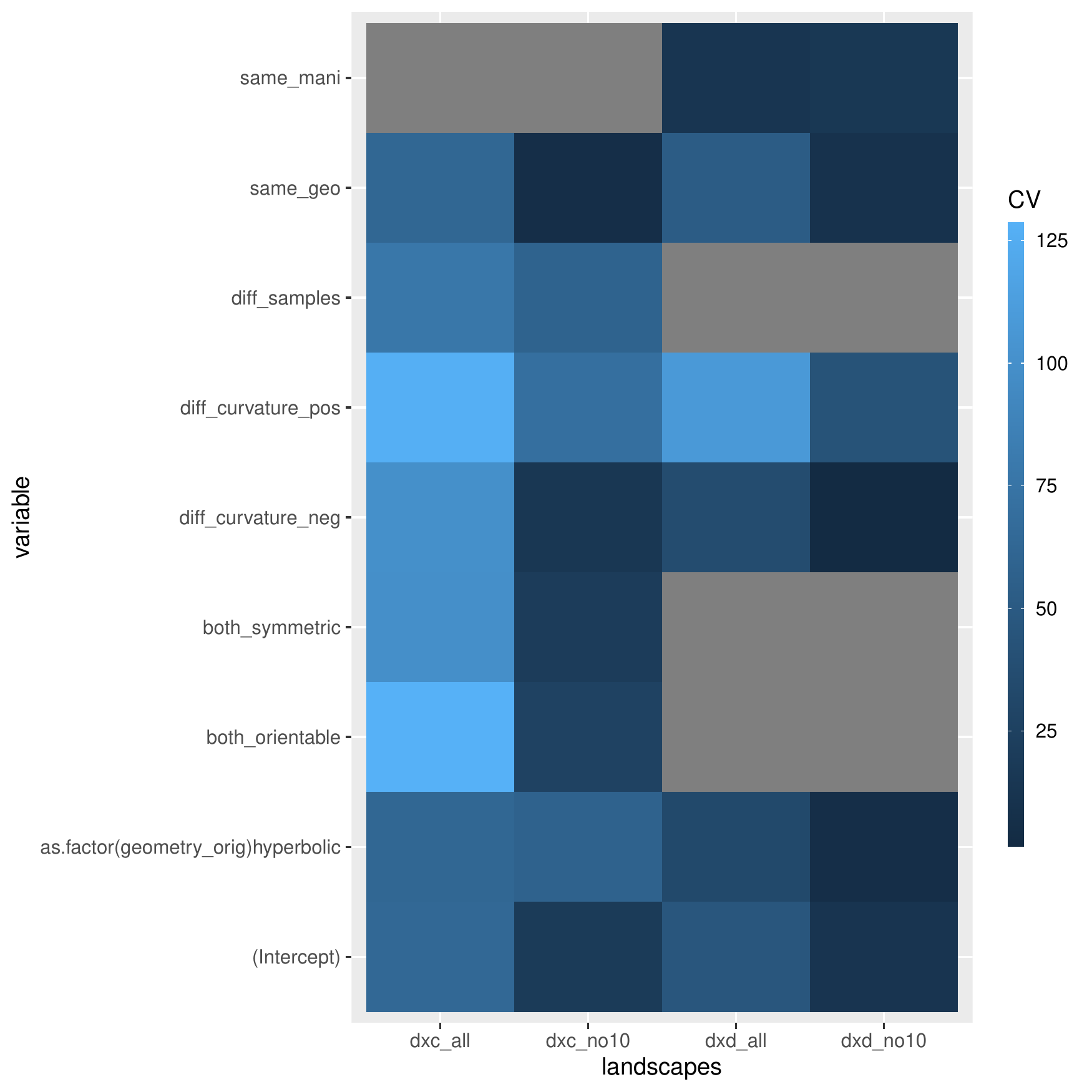} \hskip -1mm
\subfig{0.49\linewidth}{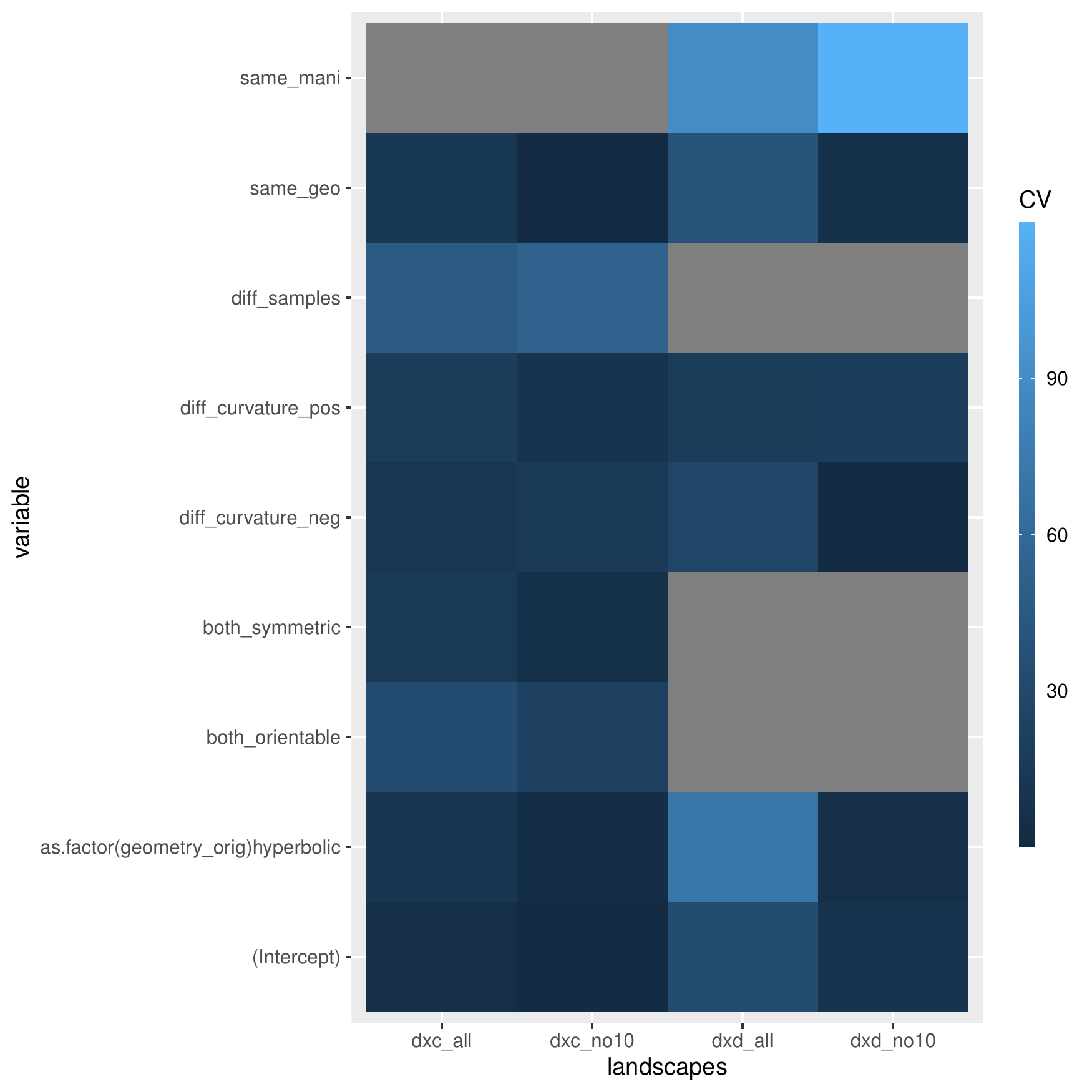}
\caption{\label{fig:lreg}
Changing the landscape dimension. Coefficient of variation (CV) for OLS coefficients.
(a)~absolute energy (simulated dispersion);
(b)~absolute Kendall unfitness (simulated dispersion).}
\end{figure}
}
\longonly{
\begin{table}[h]
\centering
\begin{tabular}{l|c|c|c|c}
                           & landscape  & all      & $E=O$ & $E\neq O$ \\ \hline
\multirow{3}{*}{energy}    & $d=10$        & 0.00     & 0.00  & 0.00 \\ 
                           & $d=30$        & 0.00     & 0.00  & 0.00 \\ 
                           & $d=60$        & 0.00     & 0.02$\ddagger$  & 0.00 \\
                           & deterministic & 0.00     & 0.56$\ddagger$ & 0.00 \\ \hline
\multirow{3}{*}{K. unfit.} & $d=10$        & 1.00     & 0.99$\ddagger$ & 1.00  \\ 
                           & $d=30$        & 1.00     & 0.96$\ddagger$ & 1.00 \\ 
                           & $d=60$        & 1.00     & 0.03$\ddagger$ & 1.00 \\
                           & deterministic & 1.00     & 0.00           & 1.00  \\ \hline
\multirow{3}{*}{Villmann}  & $d=10$        & 0.00     & 0.01$\ddagger$ & 0.00 \\ 
                           & $d=30$        & 0.00     & 0.00 & 0.00  \\ 
                           & $d=60$        & 0.00     & 0.00 & 0.00 \\
                           & deterministic & 0.00     & 0.00 & 0.00 \\ \hline
\end{tabular}
\caption{
P-values for Wilcoxon tests on differences between quality measures from SOMs with Gaussian against simulated dispersion (computed with various landscapes). $H_1$ indicates better results from simulated dispersion. $\ddagger$ denotes statistically insignificant difference.}\label{wilcoxon_landscapes}
\end{table}
As explained in Subsection \ref{sub:embed}, we are using the landscape dimension of $d=60$ for our experiments.
With a large enough value of $d$, random Gaussian vectors $v_l\in \bbR^d$ (agreeing with the interpretations above)
should produce an embedding with good properties. For simulations, we can also use the deterministic variant of the landscape
method, where we take $d=|L|$ and pick every $v_l$ to be a different unit vector.}

\longonly{
Figure \ref{fig:ldim}(ab) presents the distribution of energy and Kendall unfitness. We only consider disks as original manifolds.
Dimension 10 is clearly not sufficient in our case. With higher dimension, the distances between vertex coordinates are a better approximation
of their distances in the manifold. While the deterministic variant achieves the best scores, its high number of dimensions significantly
slows down our algorithm, and a non-determinisitc variant with a lower number of dimensions is more relevant for applications. On the other hand,
dimension of landscape does not impact significantly our qualitative results -- the insights driven from the analysis of Wilcoxon tests (Table~\ref{wilcoxon_landscapes})
and density plots (Figure~\ref{fig:ldim}) are similar and stable. Simulated dispersion scores better than the Gaussian dispersion in terms of energy.
}

\longonly{
We also checked if the choice of landscape dimension impacts the insights from OLS regressions. To this end, we computed coefficients of variation.
The coefficient of variation (CV) is the ratio of the sample standard deviation to the sample mean. From Gauss-Markov theorem, we know that if the errors in the linear regression model are uncorrelated, have equal variances and expectation value of zero (valid in our case), OLS estimator is the best, linear, unbiased estimator.
Moreover, OLS coefficients are normally distributed, so the average taken
on those coefficients obtained with different landscapes is also normally distributed. If the landscape has no significant impact on the coefficients, we should obtain relatively low CVs. Figures~\ref{fig:lreg} depict heatmaps of the obtained CVs. Similarly to the insights from Figure~\ref{fig:ldim}
we notice that the variation is higher if we include dimension 10; the coefficients obtained from higher number of dimensions are comparable. As our sample is very small,
we find the CVs low enough to conclude that the choice of the landscape had no siginificant effect on the qualitative insights from regressions.
}

\shortonly{\paragraph{Other parameters} 
The full version contains results of changing other parameters, such as the landscape dimension and the density of original manifolds.}

\section{Conclusions}
In this paper, we provide the general setup for non-Euclidean SOMs. 
We utilize symmetric quotient spaces to make our maps uniform, the Goldberg-Coxeter construction to remove the limitations related to the number and size of available grids, and suggest using a dispersion function different than Gaussian
to match the underlying geometry.

It is surprising to us that the idea of using non-Euclidean templates seems to have been neglected after
the initial papers \cite{ritter99,ontrup}. There is research on extending the SOM algorithm to the
cases where the data $D$ is no longer considered a subset of $\bbR^k$ with Euclidean distances, but rather
based the distances are based on dissimilarity matrices or Mercer kernels \cite{somrossi,kerneltrick}.
While such data representations are sometimes referred to as non-Euclidean, they are not directly related to
non-Euclidean geometry. 
Such approaches can be seen as orthogonal to ours: they run SOM on Euclidean
lattices but change the representation of $D$, while in our approach, the data manifold in still embedded
into $\bbR^k$, but we change the template. One possible direction of further research is to combine both
approaches. However, contrary to our approach, the non-geometrical nature of these settings makes
them less usable for visualization.

\longonly{
In this paper, we restricted ourselves to two-dimensional geometries. An exciting future direction is using three-dimensional visualizations.
While two-dimensional non-Euclidean geometries only differ in curvature, which can be negative, zero or positive, 
in three dimensions we have eight Thurston geometries \cite{thurston1982}. In addition to the three isotropic
geometries, we have the product spaces
$\bbH^2\times\bbR$ and $\bbS^2\times\bbR$ (obtained by adding an extra dimension to hyperbolic plane or the sphere
in the Euclidean way), twisted geometries Nil and $PSL(2,\bbR)$, and Solv, which has a different nature than
the hyperbolic space while still featuring exponential growth.
Recent advances in Virtual Reality and the visualization and tessellation of Thurston geometries \cite{rtviz,segmarching} make 
us believe that our approach can be adapted to such geometries, yielding insightful visualizations.
}
\shortonly{
In this paper, we restricted ourselves to two-dimensional geometries. 
An exciting future direction is using three-dimensional visualizations, applying the
recent advances in Virtual Reality and the visualization of Thurston geometries \cite{segmarching,rtviz}.
We believe that our approach can be adapted to such geometries, yielding insightful visualizations.
}

We are grateful to the referees, whose constructive comments on the earlier versions of this work helped us to improve the
quality of the paper. This work has been supported by the National Science Centre, Poland, grant UMO-2019//35/B/ST6/04456.

\bibliographystyle{named}
\bibliography{bibexport}

\end{document}